\documentclass[]{seed_version/bytedance_seed}
\microtypesetup{expansion=false}
\usepackage{amsmath,amssymb,needspace,url,xurl,fvextra}
\RecustomVerbatimEnvironment{verbatim}{Verbatim}{breaklines=true,breakanywhere=true,samepage=true}
\ifdefined\ACLSubmission
\newenvironment{aabtable}[1][tbp]{\begin{table*}[#1]}{\end{table*}}
\newenvironment{aabfigure}[1][tbp]{\begin{figure*}[#1]}{\end{figure*}}
\else
\newenvironment{aabtable}[1][htbp]{\begin{table}[#1]}{\end{table}}
\newenvironment{aabfigure}[1][htbp]{\begin{figure}[#1]}{\end{figure}}
\fi

\input{revision/arxiv_layout}
\input{revision/numbers}
\input{revision/draft_notes}
\graphicspath{{figs/}}
\title{BenchMint: From Real-World Tasks\\to Executable Benchmarks}
\author[1,2]{Zining~Huang}
\author[2]{Zuo~Wang}
\author[3]{Mengyao~Zhang}
\author[3]{Yilan~Piao}
\author[3]{Zhe~Wang}
\author[3]{Shuohan~Wang}
\author[3]{Ruiying~Wang}
\author[3]{Huicong~Yi}
\author[3]{Qianyu~Yang}
\author[3]{Zilong~Pan}
\author[3]{Wenlong~Wu}
\author[2]{Jiajun~Shi}
\author[2,\dagger]{Shen~Yan}
\author[3]{Xingyu~Pan}
\author[3]{Houxuan~Chen}
\author[4,3]{Yuan~Xu}
\author[1]{Jinzhao~Li}
\affiliation[1]{Tsinghua University}
\affiliation[2]{Seed, ByteDance}
\affiliation[3]{ByteDance}
\affiliation[4]{Peking University}
\contribution[\dagger]{Corresponding author}
\correspondence{Shen Yan (\email{sheny@bytedance.com})}
\hypersetup{pdfauthor={Zining Huang, Zuo Wang, Mengyao Zhang, Yilan Piao, Zhe Wang, Shuohan Wang, Ruiying Wang, Huicong Yi, Qianyu Yang, Zilong Pan, Wenlong Wu, Jiajun Shi, Shen Yan, Xingyu Pan, Houxuan Chen, Yuan Xu, Jinzhao Li},pdftitle={BenchMint: From Real-World Tasks to Executable Benchmarks},pdfsubject={Benchmark construction and evaluation methodology for foundation models in tool-using task settings},pdfkeywords={benchmark construction, foundation model evaluation, tool use, executable tasks, rubric scoring}}
\abstract{Large language models (LLMs) need evaluations of everyday reasoning and tool use, such as planning itineraries under user constraints and grounding numerical answers in evidence. We introduce BenchMint, an automated benchmark-construction workflow, and Daily-Life, its collection of 100 selected tasks across 11 real-world domains. BenchMint jointly constructs tasks and scoring criteria from the same source documents. Language-model judges assess possible answers against these criteria using tool traces and private evidence. We report 26 model/harness configurations, including GPT, Claude and Gemini families; 17 have complete domain profiles. These are descriptive configuration results, not a controlled model ranking. In one selected travel task, we corrected the reference total from 320 to 380 minutes while keeping both answers unchanged. The incorrect answer's arithmetic verdict switched from pass to fail; the correct answer's switched from fail to pass. With one ballot per cell, this exposes a reference error without establishing repeatability or collection-wide validity. Incomplete construction comparisons establish no general advantage of joint authoring or feedback-guided revision.
}
\begin{document}
\maketitle
\setlength{\parskip}{5pt plus 1pt}
\section{Introduction}
Open-ended tasks such as travel planning test how language models combine
evidence and use tools, while allowing different outputs to satisfy the same
requirements. A model using tools to plan a trip must turn booking documents,
weather information, and user preferences into an itinerary and budget.
Constructing a benchmark for this work requires specifying not only the request,
but also the evidence the model can obtain, the artifacts it must deliver, and
the criteria by which those artifacts will be judged. If a required fact is
private and inaccessible, or a reference uses the wrong reporting period, an
apparent solver failure may instead be a defect in the benchmark.

Automated construction can expand the supply of tool-using tasks. Existing
approaches generate datasets with declared properties, derive problems from
executable specifications, or compose and evolve tool-grounded
tasks~\citep{li2025autobencher,imani2025autobench,shi2025taskcraft,guo2025trace}.
Heterogeneous documents, private user facts, and artifact-level requirements
pose a further problem: a formal answer oracle is not available for every
obligation. Instructions, accessible evidence, reference answers, and rubric
criteria must remain consistent as the package is constructed and revised.

BenchMint addresses this construction problem by authoring tasks and weighted
rubrics from shared source materials. Each versioned package separates
solver-facing files and tools from private references, criteria, and controlled
user state. Execution traces and criterion-level feedback guide repair or
revision. We describe the intended consistency conditions as a
\emph{shared-evidence contract}: required facts must be obtainable, reference
targets derivable, and changes propagated to dependent components. The
implementation records these links in specifications and authoring context;
whether they hold is a property to assess on the final package, not a consequence
of joint authoring alone.

Daily-Life documents the workflow through 100 selected packages across 11
domains and records for 26 platform configurations. We first explain how a saved
travel task connects evidence, deliverables, and criteria, then distinguish the
collection, scoring endpoints, and evaluation cohorts. The results begin with
source-bound package defects and fixed-output controls, which directly test
what a grade depends on. Execution records and configuration profiles establish
what was run and reported under their respective conditions. Historical score
selection and operational reviews provide additional, observational evidence.

The contribution is an implemented benchmark-construction workflow, a documented
collection, and concrete diagnostics for evidence access, reference derivation,
and scoring coverage. These diagnostics reveal defects even in fully graded
tasks and criterion-specific responses to controlled evidence changes. They do
not establish collection-wide validity. Nor do the incomplete construction
trials establish that joint authoring or feedback improves selected final
packages; those distinct comparisons and their missing outcomes are addressed
in Section~\ref{sec:comparison-limits}.

\section{From Source Materials to Executable Tasks}
\label{sec:method}
BenchMint constructs an environment in which a foundation model can use tools
to produce scorable artifacts. Figure~\ref{fig:construction} follows the process
from captured sources to a public task, private scoring references, and
execution-based feedback.

\begin{aabfigure}[!t]
\centering
\includegraphics[width=\textwidth]{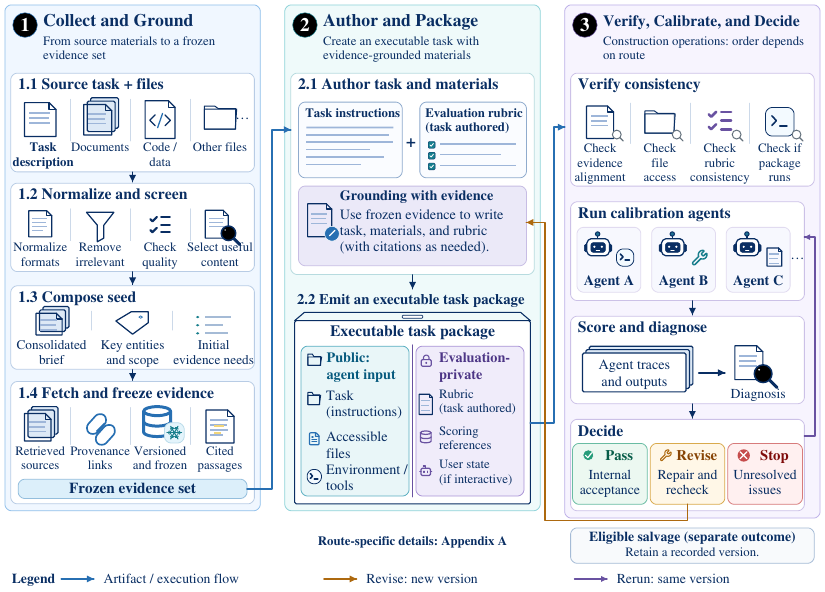}
\caption{\textbf{BenchMint's task-construction workflow.}
The three panels show evidence collection and capture, joint task/rubric
authoring, and verification with execution-based feedback. The middle panel
separates solver-visible inputs from private references and user state.
Orange feedback changes the package version; purple feedback remeasures the
same version. Check labels describe intended tests, not established properties
of every package. Acceptance and salvage are internal retention outcomes;
Appendix~\ref{app:implementation} gives the route-dependent checks and their limits.}
\label{fig:construction}
\end{aabfigure}

\subsection{Public inputs, private references, and deliverables}
\label{sec:saved-contract}
A versioned package $b_v=(u_v,m_v,e_v,o_v,r_v)$ contains instructions $u_v$,
accessible materials $m_v$, an execution environment $e_v$, private reference
targets $o_v$, and a weighted rubric $r_v$. The solver receives permitted
materials and tool access, then produces files and a final response. The judge
additionally receives reference targets and criteria. For interactive tasks,
private user state is a separate source of facts that the environment may
disclose under specified conditions; it is not unrestricted solver input.

The saved multi-city travel task illustrates the split. Booking materials,
weather references, restaurant pointers, and incomplete preferences support
two deliverables: an itinerary CSV and a Markdown report. Its private reference
and rubric require unresolved restaurant information to remain explicit:
\par\smallskip
\noindent\begin{minipage}{\linewidth}
\small\setlength{\tabcolsep}{3pt}
\begin{tabular}{@{}p{0.19\linewidth}p{0.75\linewidth}@{}}
\toprule
Accessible evidence & Merchant names, ratings, and per-person prices are unverified. \\
Required outputs & Pending entries in the itinerary CSV; dining-information gaps in the report. \\
Private checks & Y4: no invented facts; E7: CSV uncertainty; C5: report uncertainty. \\
\bottomrule
\end{tabular}
\end{minipage}\par\smallskip
\noindent Here the reference target is an information gap, not an invented
restaurant value. This is a saved specification, not a completed execution.
A changed reference version prevents attaching historical scores to it
(Appendix~\ref{app:example}).

\subsection{Authoring and checking a package}
The task author turns a seed and captured evidence into instructions,
constraints, materials, and an executable specification. The rubric author
receives the same grounded context and derives weighted criteria, check points,
failure examples, and reference targets. Conversion preserves supplied scoring
intent; regeneration can reorganize the task and rederive factual targets.

The shared-evidence contract has three conditions. \emph{Obtainability} requires
a route by which the solver can access each required fact. \emph{Derivability}
requires the reference target to follow from that evidence under the task's
assumptions. \emph{Version propagation} requires a change to reach dependent
instructions, disclosures, references, and criteria. In the travel example,
verified restaurant information would change both the expected outputs and
the criteria that currently require uncertainty. Such a change defines a new
package version.

The implementation carries these bindings in specifications and shared
authoring context, not a separately validated dependency graph. The verifier
checks structure, reachable materials, reference bindings, supported leakage
patterns, and evidence for declared reasoning challenges. Unsupported checks
remain warnings or unknowns. Check order and repair behavior depend on the
construction route; passing these checks does not establish semantic validity.
The final-version diagnostic in Appendix~\ref{app:release-contract} specifies
additional version and scoring-coverage requirements, but was not a release
gate applied to every archived package.

\paragraph{Evidence access and interaction.}
\label{sec:local-search}\label{sec:controlled-user}
Local search and fetch tools expose a frozen corpus. Task-specific local
evidence takes precedence over enabled live background search and model memory.
For interactive tasks, a controller authorizes private facts and a simulator
renders a response. Topic mentions or stalls can trigger disclosure, so a
release log establishes availability, not successful elicitation by the solver.
The interfaces, disclosure rules, and tested boundaries are specified in
Appendices~\ref{app:retrieval-control}--\ref{app:disclosure-control}.

\subsection{Execution feedback, revision, and selection}
\label{sec:hardening}\label{sec:revision-policy}
Execution traces and criterion feedback can direct material repair or revision
of a reasoning challenge, such as resolving conflicting sources or intersecting
constraints. Revision changes the package; remeasurement executes the same
version again; rescoring applies a judge to an existing trajectory. These
operations answer different questions and retain separate version records.

When enabled, calibration targets a configured score interval, tier ordering,
and endpoint separation. The Daily-Life posterior route selects the qualified
version with the lowest strongest-tier score, breaking ties by larger
separation. Ordinary acceptance and budget salvage are separate routes.
Some repair paths can retain pre-repair measurements, so internal retention
does not establish final-version calibration or valid difficulty.
Appendix~\ref{app:implementation} gives the exact policy and stopping rules.
We next define the scores and package endpoints used to interpret these records.

\section{Evaluation Setup}
\label{sec:evaluation-setup}
\subsection{Collection, selection, and reporting cohorts}
The archived Daily-Life collection contains \num{pre.n_items} selected packages
across \num{pre.n_domains} domains: \num{pre.n_scratch} authored from scratch,
\num{pre.n_regen} regenerated, and \num{pre.n_convert} converted from supplied
tasks. Human-supplied cases, use needs, and source materials inform scenario
selection; screening examines authenticity, substantive need, and authorability.
Selection also used observed model scores and domain quotas. The collection
therefore measures retained construction output, not source-to-output yield or
a representative sample of user requests. Questionnaire counts, recruitment,
expertise, and links to individual packages are unverified
(Appendix~\ref{app:example}).

The October 2026 platform snapshot reports 26 author-selected configurations.
The unit is a dated model configuration, including its harness when specified,
not a model in isolation. All 26 have aggregate scores; 17 have complete
11-domain vectors associated with the 20260710 exercise versions and 100-item
count vector. The other nine domain breakdowns are unavailable. These
associations do not establish a common final-package version, budget, or backend
identity across the platform records and package studies. The historical
21-configuration panel is a separate reporting cohort
(Appendix~\ref{app:current-dailylife-details}).

\subsection{Criterion scores and package-level endpoints}
\label{sec:agent-judge}
Agent-as-Judge inspects artifacts, tool results, traces, and dialogue against
private criteria, returning verdicts, reasons, and evidence
anchors~\citep{zhuge2025agentjudge}. Repeated judge invocations are ballots on
one output, not independent task samples.
For binary scoring, let $\widehat y_{a,c}\in\{0,1,\mathrm E,\mathrm M\}$ denote
fail, pass, evidence exclusion, or missing judgment for solver $a$ and criterion
$c$, with weight $w_c>0$. For judged set
$\mathcal K_a=\{c:\widehat y_{a,c}\in\{0,1\}\}$, the native score is
\begin{equation}
\begin{aligned}
q_a^{\mathrm{nat}}&=\operatorname{round}_{3}\!
\left(\frac{\sum_{c\in\mathcal K_a}w_c\widehat y_{a,c}}{D_a}\right),\\
D_a&=\sum_{c\in\mathcal K_a}w_c>0 .
\end{aligned}
\label{eq:criterion-score}
\end{equation}
Coverage is $D_a/W$, where $W=\sum_cw_c$ is authored weight. Excluded and
missing criteria remain explicit. At zero judged weight the ratio is undefined,
even if the stored sentinel is 0.0. Native completeness can coexist with
evidence exclusions or missing individual ballots. Appendix~\ref{app:judge-admission}
gives the aggregation and admission rules. Platform-adjusted aggregates are a
different endpoint: their transformation and per-item judgments are unavailable
in the current records, preventing reconstruction of their uncertainty.

\label{sec:state-glossary}
Package assessment separately records structure ($S$), execution ($E$),
evidence support ($B$), package quality ($Q$), credible scoring ($V$), and their
conjunction ($Y$). NULL means unestablished, UNKNOWN an unresolved check, and
NOT\_RUN an unexecuted observation. A recorded failure is distinct from all three.
The frozen study's $Q$ statuses are retained because the distributed record
does not specify the full cross-assessor aggregation rule. An accepted
full-scope comparison requires common frozen obligations, bound evidence and
versions, and complete coverage; a native score alone does not supply it.

\subsection{Evidence sources and comparison units}
A fixed, domain- and construction-mode-stratified sample of 20 packages supports
the mechanical audit. Study A assesses these frozen packages separately and
had consumed 14 of 20 opportunities at the reported freeze. The 11 recovered
historical executions establish earlier instruction-to-output links, not
11 successes in Study A. A separate imported-package experiment provides native
solver scores and a later common-reference package assessment using different
rubrics. Table~\ref{tab:evidence-index} maps each cohort to its version and endpoint.

The scorer controls hold outputs or selected output properties fixed while
changing relevant evidence or reference targets. Their repeated ballots test
within-case responses, not population accuracy. Operational human reviews are
observational and machine-visible. Historical repeat groups support score-selection
analyses, with incomplete full-version identity. Finally, Study B's joint-versus-separate
authoring comparison and the later feedback-versus-independent-reconstruction
trials target different construction effects. Their incomplete outcomes are
reported in Section~\ref{sec:comparison-limits}, not treated as successful ablations.

\section{Results}
\label{sec:retrospective}
We begin with direct evidence about package defects and scorer responses,
then examine source-bound execution, reported configuration scores, and
selection sensitivity. These evidence sources retain the distinct units
defined in Section~\ref{sec:evaluation-setup}.

\subsection{Full grading can conceal unsupported task requirements}
\label{sec:local-controls}
Two fully graded tasks contain evidence or reference mismatches
(Table~\ref{tab:study-overview}). P087 requires a 5-million-yuan operand present only in
the private seed, although its rubric attributes it to the public prompt.
P006's reference uses January--June, while the supplied source covers
February--July. Both intervals have six months, so a count check misses the
error. Both tasks have full native grading coverage but record package
quality $Q=0$.

\begin{table}[!htbp]
\centering\small
\caption{\textbf{Cohort coverage and decisive task evidence.}
The two cohort summaries have separate endpoints and observation windows
(Tables~\ref{tab:release-audit} and~\ref{tab:study-a-scope}).
Coverage is judged/authored weight, not an accepted score;
$Q$ is model-assessed package quality.
P017 is the archived exposure; other cases are final-package observations.}
\label{tab:study-overview}
\setlength{\tabcolsep}{3pt}
\begin{tabular}{@{}p{0.17\linewidth}p{0.23\linewidth}p{0.53\linewidth}@{}}
\toprule
Cohort & Observed & Endpoint status \\
\midrule
Audit &20/20 &\raggedright Complete evidence binding, answer consistency, and deployed reachability: all UNKNOWN. \tabularnewline
Study A &14/20 &\raggedright Four $Q=0$, ten $Q=\mathrm{NULL}$; six NOT\_RUN. Accepted scores, $B$, and $V$: all NULL. \tabularnewline
\midrule
Task & Grading coverage & Decisive evidence / package endpoint \\
\midrule
P017 &104/104 (full) &\raggedright 45 calls, two files, 36 votes reconstruct 30/104. \tabularnewline
P087 &109/109 (full) &\raggedright Required operand private; rubric cites public prompt. $Q=0$. \tabularnewline
P006 &121/121 (full) &\raggedright Source and reference use different six-month periods. $Q=0$. \tabularnewline
P050 &108/108 (full) &\raggedright Five failures use an unsettled discount order. $Q=\mathrm{NULL}$. \tabularnewline
P003 &0/128 (none) &\raggedright Report precedes last disclosure; no final handoff. $E=\mathrm{NULL}$. \tabularnewline
\bottomrule
\end{tabular}
\end{table}

P050 exposes a different limit: five native failures depend on a discount
order that the source leaves unsettled, while the report states its chosen
order and sensitivity. One assessor questions criterion applicability; the
second is incomplete, leaving aggregate $Q$ NULL. P003 lacks a final handoff
after the latest disclosure. These cases require rule clarification or
completed output evidence, rather than imputed negative scores.
The cohort summaries in Table~\ref{tab:study-overview} show the scope of these
diagnostics: they identify defects without estimating a population defect
rate or establishing a validated subset. Appendix~\ref{app:terminal-record}
retains every consumed opportunity, unobserved task, and endpoint.

\paragraph{A reporting-rule counterexample.}
A source-informed precipitation counterexample on development records shows
why aggregate agreement alone cannot validate reporting-set membership.
Deliberately treating every nonmissing numeric record as a report retained
zeros flagged as unreported, contrary to the task. Monthly sums stayed unchanged,
but report and recorded-zero counts increased; the task's required counts
distinguish this wrong rule. These are record statistics, not physical dry-day
counts or complete precipitation totals. The counterexample is not an observed
solver or scorer error.

\subsection{Scorer responses to controlled evidence changes}
\paragraph{Controlled changes isolate what the scorer responds to.}
LLM preference evaluation has documented position, verbosity, and
self-enhancement biases~\citep{zheng2023judging}. Our fixed-output controls
instead vary the evidence needed for a specific criterion. In the synthetic
PDF task, J7 (weight 10) requires unobscured text and table cells: the normal
PDF receives 3/3 passes and the masked PDF 3/3 failures, although both retain
extractable text. The separate J8 heading criterion passes in both.
In the earlier recording, removing section markers changes the layout verdict while preserving the
table verdict. This selective response supports local evidence sensitivity
(Appendix~\ref{app:worked-obligation}).

For interactive HTML, a later recording performs six actual click
walkthroughs: content changes in 3/3 functional controls and in none of the
no-operation controls. Its 21/24 valid ballots and three unknowns remain
separate from an earlier 23/24 recording using prepared interaction evidence;
the configurations differ (Appendix~\ref{app:judge-controls}).
A travel control changes the reference target itself.
The explicit durations sum to 380 minutes rather than 320; correcting the
reference reverses the arithmetic judgments on unchanged outputs
(Table~\ref{tab:reference-repair}). One ballot per cell identifies this
observed reversal, not its repeatability.

\begin{table}[!htbp]
\centering\small
\caption{\textbf{Reference-target intervention on fixed outputs.}
Arithmetic verdicts on unchanged incorrect (320 min) and correct (380 min)
outputs; one ballot per cell, with expected contrasts withheld.}
\label{tab:reference-repair}
\begin{tabular}{@{}lcc@{}}
\toprule
Reference target & Incorrect output & Correct output \\
\midrule
Original 320 min &Pass &Fail \\
Corrected 380 min &Fail &Pass \\
\bottomrule
\end{tabular}
\end{table}

\paragraph{Operational human feedback is narrower than validation.}
\label{sec:human-agreement}
The October 11 snapshot contains 306 quality reviews: 35 object to solution
reasonableness and 16 to rubric reasonableness. Machine-visible endorsement
is 7,539/7,815 (96.47\%), covering 7,810/20,893 answer--criterion records
(37.38\%). Nonrandom selection, visible machine decisions, and 404 unresolved
stored binary-verdict inconsistencies prevent interpreting endorsement as
independent scorer accuracy. The audit reports higher endorsement for \texttt{fail} than \texttt{pass} records, but hypothetical conflict exclusion can reverse this ordering (Appendix~\ref{app:human-platform}).
With recorded endorsement responses fixed, strict reversal first becomes attainable at 114 hypothetical machine-state recodings under the same common-population and binary-conflict model (Appendix~\ref{sec:verdict-correction-sensitivity}).

\subsection{Source-bound execution and package assessment}
\paragraph{A source-bound solver comparison.}
The conditional development experiment provides finer execution evidence
on one authenticated imported package
(Table~\ref{tab:conditional-native-main}). Under the original 24-turn ceiling,
the weak solver delivered a qualified output while the strong solver exhausted
its allowance without a final answer. A fresh supplemental strong conversation
with a 48-turn ceiling returned after 28 SDK turns. Its score 0.926 and the
retained weak score 0.851 describe different resource conditions; they do not
establish equal-resource superiority or the causal effect of additional turns.
Each score aggregates three ballots on one output, not three solver trials.

\begin{table}[!htbp]
\centering\small
\caption{\textbf{One package, with the original failure retained.}
Requests are physical solver/judge requests (S/J). Both scored outputs have
three valid ballots on the same native 10-criterion, weight-94 rubric.
No final answer means no score, not zero.}
\label{tab:conditional-native-main}
\setlength{\tabcolsep}{3pt}
\begin{tabular}{@{}p{0.40\linewidth}rrr@{}}
\toprule
Configured condition & Turn cap & S/J & Score \\
\midrule
Weak (\texttt{gpt-5.6-luna}) &24 &19/13 &0.851 \\
Strong (\texttt{gpt-5.6-sol}) &24 &24/0 &--- \\
\midrule
Supplemental strong &48 &21/12 &0.926 \\
\bottomrule
\end{tabular}
\par\smallskip\raggedright
The original 24/24 condition has no paired score difference. Strong48 versus
Weak24 is an unequal-budget observation. SDK turns and physical requests
are different counters.
\end{table}

The initial 56-request unit includes the failed strong run; the 39-request
supplement includes an unsuccessful package-comparison stage. Their
SDK-reported costs are USD 9.045275 and USD 6.276490, respectively, not
reconciled invoices. Appendix~\ref{app:conditional-partial-result} retains stage
costs and usage, and Appendix~\ref{app:conditional-strong-supplement} the
supplemental accounting.

A later common-reference assessment accepted one evidence-bound submission
supporting all 14 obligations, totaling 100/100 weight. This assesses the
constructed package against its locked source reference; the solver outputs
were not mounted in that view. It is one qualifying model assessment, not 14
experiments, independent factual certification, or a score for either solver.
Its rubric is separate from the native 10-criterion, weight-94 instrument.
The assessment used three requests and reported USD 0.37681. The cumulative
history retains 153 physical requests, 152 COMPLETE and one historical UNKNOWN
(Appendix~\ref{app:conditional-common14-result}).

\paragraph{Execution and score traceability.}
\label{sec:historical-execution}
Eleven recovered historical runs contain tool use and delivered artifacts.
This is a linkage count, not an 11/20 success rate on the later frozen sample.
For P017, the instruction-to-output chain binds 45 tool calls and results,
two files, and 36 votes over 12 criteria; the recorded score reconstructs
as 30/104. Its later supplementary exposure instead retains 44/104 native
and 37/97 evidence-revalidated weight. The full version distinctions,
coverage exclusions, and archived configurations remain in
Appendix~\ref{app:historical-traceability}.

\subsection{Recorded configuration performance and domain profiles}
\label{sec:current-dailylife}
\noindent\begin{minipage}{\linewidth}
\setlength{\intextsep}{3pt}
\paragraph{Complete configuration results.}
Table~\ref{tab:current-dailylife} reports all 26 author-selected configurations.
GPT-6-Astra-max-Codex has the largest displayed DailyLife aggregate, 83.1\%;
Claude Opus 5-max records 81.4\%. GPT-5.6-Sol-max records 77.2\% with Codex
and 78.8\% with CC. These are platform-adjusted configuration observations,
not native criterion or supplementary accepted scores
(Section~\ref{sec:evaluation-setup}). Without verified common budgets or backend
identity, these differences do not isolate model ability or a harness effect.
\begin{aabtable}[H]
\centering\small
\caption{\textbf{All 26 reported DailyLife configurations (October 2026 snapshot).}
Source labels, order, and one-decimal aggregates are preserved. Missing harness
suffixes mean unspecified. DL IDs identify provenance, not rank. Model citations
do not authenticate runs; family-only references are marked
(Appendix~\ref{app:model-reference-scope}).}
\label{tab:current-dailylife}
\begin{tabular}{@{}lp{0.49\linewidth}p{0.26\linewidth}r@{}}
\toprule
ID & Source configuration & Model reference & DailyLife \\
\midrule
DL-01 & GPT-6-Astra-max (2026.09.04)-Codex & \citep{openai2026astra6} & 83.1\% \\
DL-02 & GPT-5.6-Sol-max (2026.07.09)-Codex & \citep{openai2026gpt56} & 77.2\% \\
DL-03 & GPT-5.6-Sol-max (2026.07.09)-CC & \citep{openai2026gpt56} & 78.8\% \\
DL-04 & GPT-5.5-high (2026.04.24) & \citep{official2026entryf} & 73.6\% \\
DL-05 & Claude Opus 5-max (2026.07.24) & \citep{anthropic2026opus5} & 81.4\% \\
DL-06 & Claude Opus 4.8-max (2026.05.29) & \citep{official2026entrye} & 66.2\% \\
DL-07 & Gemini 3.8 Flash Preview-high (2026.09.02) & \citep{official2026gemini38} & 75.1\% \\
DL-08 & Gemini 3.1 Pro Preview-high (2026.02.19) & \citep{official2026entryu} & 23.6\% \\
DL-09 & Grok-4.6-high (2026.08.13) & \citep{xai2026grok46} & 74.3\% \\
DL-10 & Grok-4.5-high (2026.07.08) & \citep{official2026entryd} & 74.4\% \\
DL-11 & Qwen-3.8-Max-xhigh (2026.09.02) & \citep{official2026qwenmodels} & 74.8\% \\
DL-12 & Qwen-3.8-Max-xhigh (2026.08.03) & \citep{official2026qwenmodels} & 74.7\% \\
DL-13 & Qwen-3.8-Flash (2026.08.26) & \citep{official2026qwenmodels} & 68.1\% \\
DL-14 & Qwen-3.8-27B (2026.08.14) & family: \citep{official2026qwenmodels} & 61.4\% \\
DL-15 & Qwen-3.7-Max-high (2026.05.20) & family: \citep{official2026qwenmodels} & 61.5\% \\
DL-16 & Qwen-3.7-Plus-high (2026.06.02) & \citep{official2026entrys} & 52.6\% \\
DL-17 & Deepseek-V4.1-Flash-max (2026.09.08) & \citep{official2026deepseekapi} & 77.9\% \\
DL-18 & Deepseek V4 Pro-max (2026.08.13) & \citep{official2026deepseekapi} & 74.1\% \\
DL-19 & Deepseek V4 Flash-max (2026.07.31) & \citep{official2026deepseekapi} & 67.2\% \\
DL-20 & GLM-5.3-max (2026.08.14) & \citep{official2026glm53} & 75.7\% \\
DL-21 & GLM-5.3-Flash-xhigh (2026.08.26) & \citep{official2026glm53flash} & 70.1\% \\
DL-22 & GLM-5.2-max (2026.06.15) & \citep{official2026glm52docs} & 75.7\% \\
DL-23 & Kimi-0902-Preview (2026.09.07) & family: \citep{official2026kimifamily} & 72.3\% \\
DL-24 & Kimi-K2.7-Code (2026.06.12) & \citep{official2026kimi27docs} & 56.7\% \\
DL-25 & Hunyuan-4-Preview (2026.08.28) & family: \citep{official2026hunyuanfamily} & 64.4\% \\
DL-26 & Seed-2.1-Pro-260628-high & \citep{official2026entryi} & 66.3\% \\
\bottomrule
\end{tabular}

\end{aabtable}

\end{minipage}\par

\noindent\begin{minipage}{\linewidth}
\begin{aabfigure}[H]
\centering
\includegraphics[width=0.98\textwidth,trim=9bp 22bp 0bp 13bp,clip]{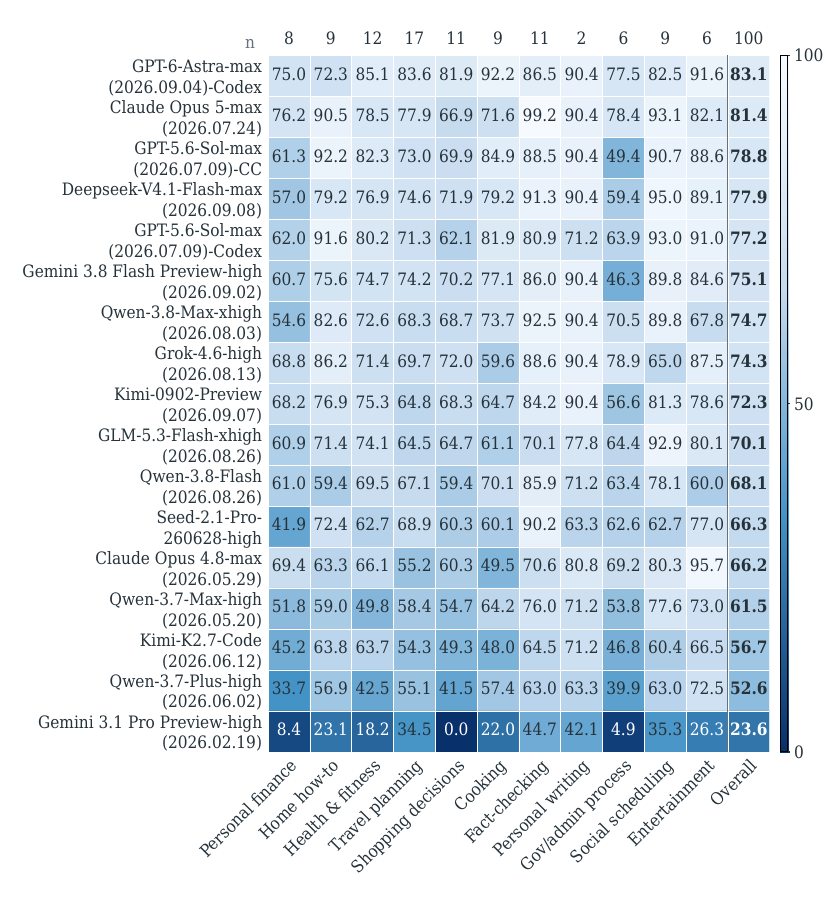}
\caption{\textbf{Domain profiles for the 17 configurations with complete records.}
Rows follow descending unrounded platform-adjusted DailyLife overall, with
DL-ID breaking ties; Table~\ref{tab:current-dailylife} retains all 26 totals.
Nine domain breakdowns are unavailable
(Appendix~\ref{app:current-dailylife-plot}).
The 11 domains use the same 20260710 exercise versions. Darker blue denotes
lower scores; displayed percentages use one decimal, retaining source precision
in the analysis. These profiles support the common-domain-weight analysis,
not a matched-budget or model-only ranking.}

\label{fig:current-dailylife}
\end{aabfigure}
\newcommand{\AABDomainCount}{17}
\newcommand{\AABPositiveSupportedCount}{7}
\newcommand{\AABBoundaryOnlyCount}{3}
\newcommand{\AABUnsupportedCount}{7}

\paragraph{Global support under domain weights.}
Using the same domain weights for every competitor, \AABPositiveSupportedCount{} of the \AABDomainCount{} configurations can co-lead under strictly positive weights. Another \AABBoundaryOnlyCount{} have boundary-only support, requiring zero weight on at least one domain; \AABUnsupportedCount{} are unsupported even by nonnegative weights. These are full-cohort configuration classifications, not independent task samples. Figure~\ref{fig:current-dailylife}'s domain crossings therefore need the whole-cohort comparison below, not just pairwise inspection.
\end{minipage}\par

The positive-supported set is a strict subset of the pairwise-nondominated set. For the additional exclusions, exact convex combinations of retained competitors improve at least one domain without worsening any. Under every strictly positive weighting, at least one constituent therefore outranks the excluded configuration. Domain crossings alone do not establish attainable leadership against the whole cohort.

Support permits ties and establishes existence within this declared linear preference family, not the size or practical plausibility of a weight region or the benchmark's adjusted aggregate. The additional exclusions are target-wise boundary cases: arbitrarily small hypothetical coordinate changes can restore positive support to a given target. This does not estimate evaluation error or assert a common perturbation for all targets. Appendix~\ref{app:domain-linear-support} gives the exact certificates, proof, and source-data checks.

\subsection{Selection sensitivity in retained construction scores}
\label{sec:selection-sensitivity}
The construction scores are sensitive to the rule used to summarize repeated
measurements.
Across 305 retained groups containing 876 observations, minimum-based and
mean-based decisions differ in 32 groups at $\tau=0.55$:
192 minima versus 160 means fall at or below the threshold.
The mean gap is 0.0964. Adaptive repetition and incomplete version identity
prevent attributing this difference to judge noise alone. Adjacent revisions
also change their rubrics, so a falling construction score need not measure
greater valid difficulty (Appendices~\ref{app:minimum-selection}
and~\ref{app:adjacent-revision}).

These observations concern the measurement used during construction. Whether
diagnostic feedback improves the independently assessed final package is a
separate question, considered in Section~\ref{sec:comparison-limits}.

\FloatBarrier
\section{Related Work}
\label{sec:related_work}
The construction problem is best distinguished by what each method produces
and how that output is checked, rather than by scores from incompatible suites.
\begin{aabtable}[H]
\centering\small
\caption{\textbf{Nearest approaches to benchmark construction.} The comparison
identifies the object each method constructs and its stated consistency mechanism;
it is not a cross-system performance experiment.}
\label{tab:nearest-construction}
\setlength{\tabcolsep}{4pt}
\begin{tabular}{@{}p{0.24\linewidth}p{0.30\linewidth}p{0.41\linewidth}@{}}
\toprule
Approach & Constructed object & Consistency or feedback mechanism \\
\midrule
AutoBencher~\citep{li2025autobencher} & Datasets with declared properties & Optimizes dataset descriptions against those properties. \\
BenchAgents~\citep{butt2024benchagents} & Generated benchmark tasks & Planning, generation, verification, and evaluation with developer feedback. \\
Meta AutoBench~\citep{imani2025autobench} & Parameterized physics problems & Python solutions, numerical and dimensional checks, corrective feedback. \\
Anchor~\citep{ivanov2026anchor} & Instructions, environments, solutions, verifiers & Derives components from a shared constraint-optimization specification. \\
TaskCraft~\citep{shi2025taskcraft} & Tool-grounded tasks & Expands atomic tasks in depth and width with incremental verification. \\
AgentSynth~\citep{xie2025agentsynth} & Computer-use tasks & Composes executed subtasks. \\
TRACE~\citep{guo2025trace} & Evolved benchmark tasks & Uses trajectories and reproduction-based validation. \\
BenchMint & Tasks with heterogeneous evidence and artifact outputs & Joint task/rubric authoring; checks evidence access, reference derivation, and version-linked grades. \\
\bottomrule
\end{tabular}
\end{aabtable}

\paragraph{Constructing tasks and their checks.}
The nearest approaches differ in what makes a generated task checkable
(Table~\ref{tab:nearest-construction}). AutoBencher and BenchAgents organize
generation around benchmark properties and feedback. Meta AutoBench and
Anchor build on programmatic or formal specifications. TaskCraft, AgentSynth,
and TRACE ground construction in tool use, executed subtasks, or reproduced
trajectories. BenchMint focuses on heterogeneous documents, private user facts,
and required output artifacts, where a formal answer oracle is not available
for every criterion. Its workflow links sources, references, and criteria.
The diagnostics examine whether accessible evidence, private references,
and recorded grades are consistent for each requirement.

\paragraph{Feedback and downstream utility.}
Autodata~\citep{kulikov2026autodata} couples feedback-driven data generation
with final quality and rubric checks, then evaluates downstream training.
AutoPlay~\citep{ramrakhya2025autoplay} likewise tests the training utility of
exploration-generated tasks. Self-Challenging Language Model
Agents~\citep{zhou2025selfchallenging} links generated instructions to
verification functions and success/failure examples. For BenchMint, the utility
question is whether feedback improves the selected final packages.
Answering it requires assessment measurements that were not used to select
those packages.

\paragraph{Reliable agent evaluation.}
AndroidWorld~\citep{rawles2025androidworld} distinguishes repetition from task
variation; $\tau$-bench~\citep{yao2025taubench} tests consistency across
user--agent interactions. HAL~\citep{kapoor2025hal} jointly examines models,
scaffolds, costs, and logs, while Agent-as-a-Judge~\citep{zhuge2025agentjudge}
inspects behavior and progress. Our fixed-output controls vary the evidence
or reference target for individual criteria and inspect the resulting verdicts,
complementing these execution-level analyses.

\FloatBarrier
\section{Discussion and Limitations}
\label{sec:discussion}
BenchMint makes the relationship between a task's evidence, expected outputs,
and scoring criteria explicit during construction. The results show why this
relationship must also be checked after construction: complete grading can
coexist with an inaccessible operand or an incorrectly derived reference.
Fixed-output controls show that a scorer can respond to the relevant evidence,
but also follow an incorrect reference. Joint authoring therefore provides
bindings that can be inspected; it does not by itself certify them.

\subsection{Two construction effects remain unestablished}
\label{sec:comparison-limits}
\paragraph{Joint versus separate authoring.}
Study B compares joint task/rubric authoring (JC) with separate generation and
validation (GV). It consumed 7/8 planned opportunities and 115 physical calls.
For source 2, JC stopped after one call with all endpoints NULL; GV used 108
calls and delivered outputs, but judged only 54/155 weight, with 47 passing and
101 unjudged. Source-3 GV completed two generation/validation calls before
emission failed the instruction-preservation contract; source-3 JC was not
started. The earlier four consumed opportunities retain their NULL/stopped
records. Unmatched obligations, unequal realized resources, and incomplete
outcomes prevent a comparison of authoring organizations
(Appendix~\ref{app:construction-pair}). This is not an ablation of the full
feedback workflow.

\paragraph{Feedback-directed revision versus independent reconstruction.}
This contrast asks whether a candidate's own diagnostic feedback improves the
selected final package. The intended comparison holds the source, constructor,
candidate count, selection rule, and resource budget fixed, and assesses
obligations locked before construction using measurements not used for selection.
The later two-source block attempted 2/4 candidate positions and qualified 0/4:
26 HTTP starts produced 25 COMPLETE responses and one UNKNOWN, with 0/36 judge
ballots and 0/4 final-assessment stages completed. A separate rubric-reserved
run attempted three positions and qualified 0/4 despite 60 completed requests;
all three saved drafts had empty rubrics, and the fourth position lacked its
required prior feedback. These are construction failures and missing assessment
outcomes, not estimates of a zero or negative feedback effect
(Appendices~\ref{app:prospective-interface-capacity}
and~\ref{app:rubric-reserved-construction}).

The later imported-package execution and common-reference assessment answer
narrower questions. They do not complete either construction comparison.
A feedback-effect estimate still requires matched final packages, full
failure-inclusive accounting, and assessment evidence not reused for selection.
The separately frozen technical continuation likewise supplies no paired
task-score difference (Appendix~\ref{app:terminal-followup-r797}).

\subsection{What final-package assessment must distinguish}
\paragraph{Answer agreement and policy coverage.}
A separate authored fictional repair-cafe task illustrates a limit of testing
on a fixed input. Its physical request order already matches the mandated sort
keys, and its rejection conditions never overlap. Retaining row order or
changing rejection precedence can therefore reproduce the same allocation
table with the downstream scheduler held fixed. Unlike the observed P087/P006
defects, this is a source-specific identifiability argument, not a failed
package assessment or a solver experiment. Table agreement alone cannot show
which policy the package retained; package content and the applicable source
rules also matter (Appendix~\ref{app:source_policy_identifiability}).

\paragraph{Collection-wide validity.}
The 20-package audit leaves complete evidence binding, answer consistency,
deployed reachability, and solver feasibility unresolved. Study A's four
recorded $Q=0$ outcomes identify defects, while the other ten consumed
opportunities retain $Q=\mathrm{NULL}$ and six remain NOT\_RUN.
Neither the 100 selected packages nor the 11 recovered historical executions
establish a validated subset or a collection-wide usability rate.
The common-reference assessment supports one package's 14 obligations under
a model-based protocol, not independent factual certification of the collection.

\paragraph{Scoring, interaction, and generalization.}
Independent judge accuracy remains unmeasured. Machine-visible endorsement
covers 37.38\% of the answer--criterion frame, and the causes of 404 stored
binary-verdict inconsistencies remain unresolved. Shared model-family bias,
incomplete version binding, and adaptive repetition limit generalization from
the diagnostic and historical records. Frozen evidence can be wrong; live
fallback, solver sampling, and judging can vary. Disclosure authorization does
not establish generated-text compliance or successful elicitation
(Appendix~\ref{app:disclosure-control}).

Questionnaire counts, recruitment, expertise, and seed-to-task links remain
unverified. Representativeness, user-feedback benefit, and transfer to other
models or task populations are unmeasured. The public bundle supports
documentary analysis and controller replay, but lacks artifacts needed for
full construction, native judging, and final-package execution
(Table~\ref{tab:configuration-crosswalk}). Its reproducibility scope is narrower
than an executable public release of the full benchmark.

\section{Conclusion}
BenchMint constructs tool-using tasks and weighted rubrics from shared evidence,
linking what a model can access to what it must deliver and how it is scored.
Daily-Life documents the resulting packages and their recorded evaluations.
The strongest evidence is local: fully graded tasks can contain unsupported
requirements, and fixed-output controls distinguish responses to evidence from
responses to an incorrect reference. These findings make final-package checks
a necessary complement to construction-time feedback. Collection-wide validity
and the effects of authoring organization and diagnostic feedback remain
unestablished; the current execution and assessment records do not replace
matched comparisons of independently assessed final packages.

\FloatBarrier
\section*{Reproducibility Statement}
The accompanying analysis archive (\path{anc/analysis/}) provides capability
scores, plotting and verification scripts, stored reference-repair outcomes, and
retrieval/disclosure source replay. A frozen claim ledger binds reported quantities
to their sources. Appendix~\ref{app:public-reproduction} gives copyable commands;
Table~\ref{tab:configuration-crosswalk} identifies public inputs, stored outcomes,
and unavailable artifacts. Full construction, native judging, and final-package
execution cannot be reproduced from this bundle.
The authors used AI tools for literature searches, manuscript preparation, implementation, and feedback on the methods.
\Needspace{6\baselineskip}
\section*{Ethics Statement}
The authors confirm the required permissions for research use and public release of
the accompanying analysis materials, and compliance with applicable human-data
ethics requirements.

The archived release-scope check omitted consultation-draft notes marked against
redistribution and adoption, retained seven unchanged sanitized P017 projections,
and recorded text, link, and PDF-metadata checks. These are scoped disclosure and
integrity checks, not a corpus-wide license, consent, or ethics certification.
Private originals remain withheld. Access to a source does not establish rights
to redistribute it or a derived task. The documentary material is intended for
studying evaluation, not for medical, financial, or legal advice or deployment.

Questionnaire counts, recruitment, expertise, and links to published tasks remain
unverified. Operational reviewer recruitment and full historical instructions
are not documented in the available records. Human intake and machine-visible quality
control must not be treated as independent validation.
\par
\bibliographystyle{plainnat}
\bibliography{revision_references}
\FloatBarrier
\Needspace{10\baselineskip}
\appendix
\section*{Appendix Guide}
The appendices separate construction details, package observations, scoring
controls, and historical measurements. For implementation and score definitions,
start with Appendix~\ref{app:implementation}; for the fixed 20-package sample
and all Study A outcomes, Appendix~\ref{app:example}; for fixed-output controls,
Appendix~\ref{app:judge-controls}. Historical executions are in
Appendix~\ref{app:detailed-evaluation}, and statistical panels and operational
reviews are in Appendix~\ref{sec:historical-statistics}.
The incomplete feedback comparisons and the separate imported-package assessment
are in Appendices~\ref{app:prospective-interface-capacity}--\ref{app:conditional-common14-result}.
Table~\ref{tab:evidence-index} maps cohorts to exact endpoints and locations.
Status definitions are in Appendix~\ref{app:state-glossary}; offline commands
and their reproduction limits are in Appendix~\ref{app:public-reproduction}.
\begin{aabtable}[!htbp]
\centering\small
\caption{Evidence guide: cohorts, endpoints, and where to find them.
Rows retain separate versions and denominators; they do not form a validated release.}
\label{tab:evidence-index}
\begin{tabular}{@{}p{0.25\linewidth}p{0.28\linewidth}p{0.40\linewidth}@{}}
\toprule
Question / evidence & Unit / denominator & Result, protocol, and boundary \\
\midrule
Final-package validity & Fixed 20 of 100; 14 consumed opportunities & Supplementary frozen versions: Tables~\ref{tab:study-a-scope} and~\ref{tab:study-a-current}; all accepted scores and $B,V$ remain NULL. Failures and unobserved cases remain in Appendix~\ref{app:terminal-record}. \\
Study B: construction comparison & 8 opportunities; 7 consumed; 115 calls & Appendix~\ref{app:construction-pair}. No accepted score; source-3 JC unspent. Independent of Study A. \\
\raggedright Feedback comparison & \raggedright 4 planned positions in each later trial; 0/4 qualified in each & \raggedright Appendices~\ref{app:prospective-interface-capacity} and~\ref{app:rubric-reserved-construction}. Incomplete construction, not a measured feedback effect. \tabularnewline
\raggedright Imported-package assessment & \raggedright One package; native weight 94, common-reference weight 100 & \raggedright Appendix~\ref{app:conditional-partial-result}. Original strong failure retained; supplemental solver budgets differ. Appendix~\ref{app:conditional-common14-result} reports one qualifying model assessment. \tabularnewline
Local scoring sensitivity & Two synthetic tasks, eight criterion instances per recording & Tables~\ref{tab:judge-controls} and~\ref{tab:judge-controls-r310}; distinct configurations, not pooled repetitions. \\
Reference intervention & One selected task, four conditions & Table~\ref{tab:reference-repair}; Appendix~\ref{app:reference-repair}. \\
Historical executions & 11 recovered linked runs & Table~\ref{tab:historical-execution}. Historical P017: $30/104$; supplementary P017: native $44/104$, evidence-supported $37/97$ (Appendix~\ref{app:archive-verification}). Historical and supplementary P068 have different judge runtimes. \\
Later terminal follow-up & Two sources; four original rows & Appendix~\ref{app:terminal-followup-r797}. Source-0/source-1 are not Study B indices. Original source-0 $S=0,Y=0$ and source-1 NULL rows remain; continuation $E,B,Q,V,Y$ are NULL, with B exposed but unobserved. \\
Human observations & 306 quality reviews; 7,815 endorsement review records & Tables~\ref{tab:human-quality}, \ref{tab:human-frame}, and~\ref{tab:human-endorsement-october}. September/October share endorsements; July remains separate (Table~\ref{tab:human-agreement}). \\
Platform configurations & 26 totals; 17 complete vectors & Table~\ref{tab:current-dailylife}; Figure~\ref{fig:current-dailylife}. Historical panel: Table~\ref{tab:model-configurations}. \\
\bottomrule
\end{tabular}
\end{aabtable}

Component validation has distinct units and incomplete denominators
(Appendix~\ref{app:module-validation}); only screening has a fully recorded
numerator and denominator, 55/62. The selected adjacent-revision pairs and repeated
measurement groups have their full accounting in Appendix~\ref{sec:historical-statistics}.
Figure~\ref{fig:historical-repeats} retains observed spans, not confidence intervals.

Unique historical units are in Table~\ref{tab:cohort-inventory}; configuration details
and reproduction limits are in Table~\ref{tab:configuration-crosswalk}.

\section{Implementation Details and Version Semantics}
\label{app:implementation}
\paragraph{State glossary.}
\label{app:state-glossary}
Section~\ref{sec:state-glossary} defines NULL, UNKNOWN, and NOT\_RUN;
Appendix~\ref{app:judge-admission} defines native completeness.
PASS and FAIL mean that the named check is supported or
violated. A \emph{conditional score} divides passed weight by judged weight and
is undefined when judged weight is zero. A \emph{native score} is the scorer's
stored aggregate under its own configuration.
An \emph{admitted calibration score} passes the native admission checks; this
need not imply full rubric coverage. An \emph{accepted score} is the supplementary
study's distinct endpoint, requiring its evidence and full-coverage contract
(Appendix~\ref{app:judge-admission}). File delivery and conditional ratios do not
supply an accepted score. The package endpoints $S,E,B,Q,V,Y$ are defined below.

Figure~\ref{fig:detailed-construction} expands Figure~\ref{fig:construction}'s
solver/private boundary: it shows which evidence the solver can access,
which user facts can be disclosed, and what the judge can inspect.
\begin{aabfigure}[H]
\centering
\includegraphics[width=\textwidth]{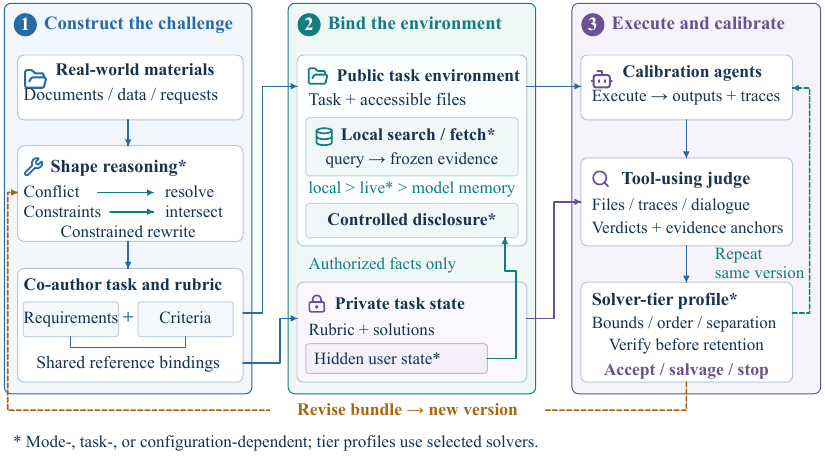}
\caption{\textbf{Evidence access and judgment within the construction workflow.}
This view expands Figure~\ref{fig:construction}'s public/private split. Shared reference bindings
connect task requirements, reasoning challenges, and rubric criteria. Public inputs
include frozen retrieval and controlled disclosure for applicable tasks; private
state supplies authorized user facts and judge criteria. Agents produce artifacts
and traces for tool-based judging. Internal calibration either retains an output,
revises the bundle into a new version, or repeats execution of the same version.
The evidence-priority relation governs task facts, not measured model quality.
Configuration-dependent paths are marked with an asterisk. Internal retention
precedes independent usability assessment.}
\label{fig:detailed-construction}
\end{aabfigure}

The following details expand the configurable operations in Section~\ref{sec:method}.
They describe system routes and the version boundaries of their artifacts.
Optional source expansion creates additional seeds and is accounted for separately
from the default construction route.

\paragraph{Origin of task requirements and scenarios.}
The authors report four seed channels: expert and user questionnaires about
practical task needs, crowd-test cases, reward-plan and expert-platform submissions,
and internal review with distilled failure reports. Project documentation
corroborates case/questionnaire intake and raw-seed quality checks, but not a
particular product, collection interface, or usage-log origin.
Screening examines authenticity, substantive need, suspected synthetic origin,
and authorability. Intake supplies scenario choices and reference pointers;
scoring uses the captured task evidence, not questionnaire responses as ground truth.
Respondent counts, recruitment, expertise, and seed-to-final-task links are
unavailable, so no published task or count is attributed to a questionnaire channel.
Raw questionnaires, identities, and internal source links are not distributed.

\subsection{Evidence capture and task authoring}
\paragraph{Captured evidence and version identity.}
Grounding resolves source pointers into local evidence snapshots. Captured documents
and attachments carry their source location, local file location, content hash, and
retrieval metadata. These records fix the evidence used during authoring and allow
later checks against the same captured content. Refreshing a source during repair
updates the capture and its retrieval record, creating a changed evidence version.
Frozen therefore refers to a captured version, not an unchangeable construction history.
Self-contained tasks can proceed without external retrieval;
live tool access, when enabled, remains part of the execution configuration. Source
reliability and the correctness of derived reference values are assessed separately
from snapshot identity.

\paragraph{Package fields and private state.}
The bundle in Section~\ref{sec:method} is a conceptual grouping, not the storage schema:
\texttt{simulator\_spec} is a separate specification field, emitted under
\texttt{evaluation/simulator/spec.json} and read by the disclosure controller.
Its private facts are not judge-only state. A source binding records a source location,
captured content identity, and the referenced fact or derivation; a disclosure binding
additionally identifies the condition under which a private fact becomes available.
Rubric-entry indices identify the scoring obligations, while capability labels may
repeat across entries. These relationships are carried in specifications and authoring
context, rather than asserted to be a separately validated dependency graph.
A change to any bound surface requires a new version record. Reusing only a task ID,
round tag, or capability label cannot establish an unchanged package.

\paragraph{Task and rubric responsibilities.}
The task author uses the seed and frozen evidence to produce the instruction,
constraints, materials, and executable specification. Within this step, a dedicated
rubric-generation component receives the task context, reference bindings, and available
evidence. It produces weighted criteria with check points, failure examples, and derived
reference targets, and can revise them in response to bounded quality feedback. Supplied author
rubrics have a mode-specific role: conversion treats them as an alignment constraint,
whereas regeneration uses them as a reference and rederives factual values from the
evidence. These are distinct authoring responsibilities; agreement between generators is
recorded separately from independent validation.

\subsection{Verification, calibration, and version changes}
\paragraph{Build verification and repair.}
The default build route is composition, grounding, task and rubric authoring, emission,
and verification. The verifier returns \texttt{PASS}, \texttt{WARN}, or \texttt{FAIL},
together with reasons and a failure category. Its checks cover required structure,
reference bindings, evidence accessibility, and supported forms of answer leakage.
For example, missing required input material and a supplied file matching a reference
answer require different repairs. A material format that the leakage scanner cannot
read is reported as a warning. The default batch route permits warnings and attempts
category-specific repair for selected failures before rerunning the build and verifier;
remaining failures stop that source attempt. Holistic quality review is an additional,
explicitly invoked operation.

\paragraph{Configured tier profile.}
For admitted normalized scores $q_a\in[0,1]$, configured tiers
$\mathcal A_1,\ldots,\mathcal A_T$ run from intended weaker to stronger.
The controller uses $z_j=\max_{a\in\mathcal A_j}q_a$, the best score in tier $j$.
When profile filtering is enabled and every required score is available, it requires
\begin{equation}
\begin{gathered}
\ell\le z_T\le\tau,\qquad z_{j+1}\ge z_j\ (j<T),\\
z_T-z_1\ge\delta.
\end{gathered}
\label{eq:calibration-profile}
\end{equation}
Here $\ell,\tau$ bound the strongest tier's target score and $\delta$ sets
minimum endpoint separation; adjacent tiers may tie.
The tier ordering and thresholds are construction configuration choices, not a model
ranking inferred from a generated task.
For the stated configured protocol, ordinary calibration uses
Equation~\ref{eq:calibration-profile}; eligible posterior salvage additionally checks
strict ordering between the weakest and strongest retained scores. Neither establishes
independent validity.

\paragraph{Resampling the same task version.}
The construction controller can collect additional executions when tiers invert, the strongest-tier
score lies just above the ceiling, or a judge-disagreement diagnostic overlaps it.
This branch preserves the task version, takes the median of valid scores for each
agent, and reevaluates the tier profile. If any agent has fewer than two valid samples,
it retains the preceding decision. The disagreement diagnostic is derived from split
criterion votes and serves as a resampling trigger, not a confidence interval.

\paragraph{Revising task artifacts.}
Revision instead changes the candidate artifacts. Feedback combines the preceding
instruction and rubric with criterion verdicts, judge reasons, and trace evidence such
as failed material reads. A failure accompanied by an inaccessible attachment directs
repair toward material access before further difficulty adjustment. Other feedback can
revise task difficulty or the targeted criteria. The author then regenerates the
specification and the emitter writes the revised package.
Table~\ref{tab:construction-control} summarizes these distinct operations.

\begin{aabtable}[!htbp]
\centering
\small
\caption{\textbf{Implemented diagnostic operations.} Observed conditions trigger
artifact repair, repeated execution, revision, retention checks, or stopping.}
\label{tab:construction-control}
\begin{tabular}{@{}p{0.29\linewidth}p{0.64\linewidth}@{}}
\toprule
Observed condition & Operation and affected object \\
\midrule
Material or verification defect & Repair relevant artifacts; rebuild and reverify. \\
Unstable calibration signal & Collect new executions of the same task version. \\
Difficulty profile missed & Revise task artifacts using criterion and trace feedback. \\
Calibration condition met & Check grounding and computed references before retention. \\
Limit or unresolved failure & Stop, or salvage an eligible recorded version. \\
\bottomrule
\end{tabular}
\end{aabtable}

\paragraph{Reference checks, rescoring, and version records.}
Before acceptance or salvage, grounding checks and a separate model-based recomputation
of declared computed targets can trigger further repair. Confirmed numerical mismatches
are distinguished from values the recomputation cannot establish. Certain rubric
repairs also request rescoring of existing trajectories; rescoring is distinct from a
new agent execution. In this path, the calibration decision can still use pre-repair
scores; fresh execution is needed to attribute calibration to the final version.
Round snapshots preserve specifications and calibration summaries,
and execution records contain an execution identifier and hashes of the instruction,
rubric, materials, search corpus, simulator specification, and oracle manifest.

\paragraph{Stopping and retention outcomes.}
The loop is bounded by its round limit and, when enabled, a wall-clock limit. It can
accept a calibrated candidate, salvage an eligible recorded version, or terminate
without retaining one. Salvage remains a separate internal outcome. Version-linked
records distinguish the retained artifacts from the versions that supplied calibration
scores; unresolved joins remain explicit.

\paragraph{Archived retention routes.}
The following phases summarize historical branches; they do not assert
retrospectively enforced version or coverage checks.
\begin{enumerate}
\setlength{\itemsep}{2pt}\setlength{\parsep}{0pt}
\item \textbf{Emit and measure.} Within the configured budget, emit the package,
execute solvers, and admit valid judge scores. Compute the rounded ratio of
graded weighted passes to graded weight; keep authored weight and exclusions separate.
\item \textbf{Repeat a version.} When resampling is triggered, rerun that same
version. Use per-agent medians when enough valid samples exist; otherwise
retain the preceding decision.
\item \textbf{Revise the package.} Requested material repair or hardening
regenerates and emits a new version, then continues the revision loop.
\item \textbf{Check before retention.} Before acceptance or eligible salvage,
check grounding and recompute declared reference targets. Repair confirmed
mismatches where the route permits. Some rubric repairs rescore existing
trajectories while the decision can still retain pre-repair scores.
\item \textbf{Select or stop.} Record qualified snapshots. Eligible Daily-Life
tasks continue posterior hardening within the round budget. Restore the qualified
package with the lowest strongest-tier score, breaking ties by greater tier
separation; otherwise use eligible salvage or terminate without output.
\end{enumerate}
The last selection rule applies to the Daily-Life posterior-candidate route;
ordinary acceptance and budget salvage are separate branches. Changing instructions,
materials or search captures, tools/disclosure configuration, rubric text or weights,
or reference targets invalidates attribution of earlier scores to an unchanged
package. A rescore reuses a trajectory; it does not test behavior under changed inputs.
Neither retaining a previous decision after insufficient resampling nor retaining
pre-repair scores establishes final-version calibration. Complete score summaries
also need not imply full graded weight (Appendix~\ref{app:judge-admission}).

\emph{Recommended final-version requirement, not historical enforcement:}
Appendix~\ref{app:release-contract} specifies the positive-weight, coverage,
version/configuration, and unused-for-selection checks, with a read-only metadata
checker. Full-rubric claims require all authored weight to be graded. The checker
is separate from the original controller and does not supply fresh measurements
or retrospectively establish calibration for any archived package.

\subsection{Local evidence access and its limits}
\label{app:retrieval-control}
The task-fact precedence rule is
\begin{equation}
E_{\mathrm{local}}\succ E_{\mathrm{live}}\succ K_{\mathrm{model}},
\label{eq:evidence-priority}
\end{equation}
where $\succ$ denotes precedence, not measured source quality. Local materials
fix task-specific facts; enabled live search supplies missing background;
model memory cannot replace required retrieval or overwrite frozen values.
The emitter builds a search index of candidate titles, URLs, and snippets and a URL-to-page
index over successful nonempty captures. The MCP search server normalizes case and
whitespace, attempts an exact key match, then uses token and Chinese-substring scores
with a unique-best-match condition in the main implementation. The Daily-Life
implementation instead merges tied best matches, deduplicating by URL. Fetch normalizes
the requested URL and reads the stored page body. For fixed indices and query, these operations follow fixed code paths;
they do not invoke a language model to generate retrieval results. In both frozen
implementations, an absent \texttt{\_search\_index.json} loads as an empty mapping and
\texttt{search} returns \texttt{error}; an unmatched query returns \texttt{miss}.
An absent \texttt{\_web\_index.json} also loads as an empty mapping, but \texttt{fetch}
returns \texttt{miss}, just as for an unknown URL, even when search succeeds and the
captured page exists. A missing page referenced by an existing fetch entry returns
\texttt{error}. Thus the frozen source does not uniformly distinguish unavailable
retrieval infrastructure from absence of matching evidence.

The corpus is placed outside the ordinary workspace-file presentation to elicit
search/fetch use. This placement is an interaction convention, not a security guarantee
against arbitrary filesystem exploration. Its contents are permitted task evidence.
Rubrics, answers, and simulator-private facts have a separate exclusion boundary.
Online search is attached at runtime when available, with the solver prompt generated
from the same availability decision. Local-tool descriptions and miss responses can
nevertheless suggest live search without checking attachment; these hints do not
establish tool availability. Without that tool the prompt reports the missing
channel instead of asking the model to imitate web retrieval. Live results must not
overwrite frozen task-specific values. The precedence rule is in the solver prompt;
the judge receives the task's rubric and reference evidence, rather than an assumed
identical prompt rule.

This policy defines the benchmark's information setting. It does not claim that a local
page is universally more accurate than a current online source or that agents always
obey the instruction. Captured content can be wrong or outdated. Retrieval replayability,
evidence-policy adherence, and score stability are therefore separate measurements.

\subsection{Disclosure control and conversational rendering}
\label{app:disclosure-control}
With private facts $\mathcal H$, visible history $h_{\le t}$, and controller
state $\sigma_t$ containing rules, prior disclosures, and triggers, authorization
and rendering are separate operations:
\begin{equation}
\begin{aligned}
\mathcal D_t&=\operatorname{Disclose}(\sigma_t,h_{\le t})\subseteq\mathcal H,\\
y_t&=\operatorname{Render}(h_{\le t},\mathcal D_t).
\end{aligned}
\label{eq:controlled-disclosure}
\end{equation}
Here $\mathcal D_t$ includes authorized facts and prior disclosures; $y_t$ is
the rendered response. Configuration and persona are implicit. Authorization
does not by itself establish compliance of generated text.
Hidden facts support opening disclosure, keyword-triggered questions or topics, and
never-disclosed fields. Progress rules can volunteer eligible facts after delayed
clarification or a stall, with their reasons retained in the trace. Eligibility,
never-disclose protection, and suppression of already committed disclosures remain
rule constrained. The main implementation uses lexical question-key matching.
Daily-Life additionally calls \texttt{\_semantic\_asked\_keys} for undisclosed
\texttt{on\_ask}/\texttt{on\_probe} candidates without lexical matches outside the
opening turn. With a backend, it sends the utterance and candidate keys/aliases, not
their private values, to \texttt{backend.generate}; returned keys are restricted to
the candidate set and can authorize disclosure. No backend, invalid output, or an
exception yields no semantic keys. Never-disclosed and already disclosed facts are
excluded from these candidates.

\paragraph{Lexical trigger boundary.}
In the supplied frozen source, \texttt{on\_ask} accepts a lexical topic match,
including a negated mention. With the archived synthetic fixture and
\texttt{backend=None}, both implementations release \texttt{authorized\_budget}
for ``What is the budget?'' and ``I will not ask about the budget.'', logging
\texttt{on\_ask@asked} in both cases. Unrelated turns can instead release that fact
through stall recovery; the Daily-Life stall branch does so even with
\texttt{hold\_volunteer=True}. The trace therefore records the branch that authorized
release, not a validated distinction between a mention and an actual request.
A volunteered fact is available without having been actively elicited. Clarification
assessment must distinguish actual requests, mentions, and volunteered assistance
using the interaction content, rather than infer elicitation from availability or
this trigger label. These supplied-source counterexamples do not test generated
responses or estimate prevalence in Daily-Life. The frozen implementations and
original fixture results are retained unchanged.

Deterministic disclosure replay requires fixed specification, inputs, and full
controller state on the lexical-only route; the semantic route additionally requires
frozen matching decisions. Reinvoking a stochastic matcher need not reproduce the
same disclosure set. The supplied replay uses \texttt{backend=None}; it does not
establish that semantic decisions are frozen or logged for replay. Response generation
remains model based. The implementation tests unauthorized lexical, entity, and number matches
and checks that required numeric or filename anchors are actually conveyed. Failed
checks can trigger rewriting or a template carrying the authorized values. Suspected
simulator assistance is logged for review. These checks do not prove complete semantic
coverage or human-equivalent behavior.

For supported DOCX and XLSX outputs, an optional user-feedback step presents image
previews reconstructed from extracted text and cell values to a model, together with
the visible dialogue. These bounded previews support content-level feedback but do not
preserve native document layout or establish formatting quality; Agent-as-Judge requires
separate rendered evidence for layout assessment.

\subsection{Declared reasoning mechanisms and verification limits}
\label{app:design-details}
\paragraph{Declared mechanisms and their evidence checks.}
The author declares planted mechanisms in the specification; the rubric author receives
these declarations alongside source captures, visible material paths, solution approach,
and reference files. Structured \texttt{twist\_audit} records encode the claimed source
conflict, candidate set, time-dependent values, or missing subgoal. The verifier tests
necessary conditions for these claims against shipped and captured material. Table~\ref{tab:semantic-levers}
summarizes the distinction between an intended reasoning challenge and what the checks
can establish. These are construction mechanisms, not measured effects on model scores.

\begin{aabtable}[t]
\centering\small
\caption{\textbf{Semantic difficulty mechanisms and bounded checks.} Corpus checks
inspect concrete support for declared mechanisms; they do not prove reasoning depth,
uniqueness, or exhaustive absence of an answer.}
\label{tab:semantic-levers}
\begin{tabular}{@{}p{0.22\linewidth}p{0.33\linewidth}p{0.38\linewidth}@{}}
\toprule
Mechanism & Intended challenge & Implemented evidence check \\
\midrule
Source conflict & Resolve incompatible source values. & Declared values must occur in the relevant corpus. \\
Constraint intersection & Identify the candidate satisfying all constraints. & The answer and declared near-miss candidates must be present. \\
Multi-hop evidence & Combine distributed clues and eliminate distractors. & Target, features, and distractors must occur in the corpus. \\
Temporal reasoning & Apply the specified reference date. & Declared stale and current values must both be supported. \\
Ambiguous reference & Resolve a description among plausible entities. & The clue must not already name the answer; target and distractors must exist. \\
Unsupported subgoal & State that requested information is unavailable. & A declared missing probe must not be found by the implemented corpus matcher. \\
\bottomrule
\end{tabular}
\end{aabtable}

The mechanism vocabulary is shared across authoring, schema validation, and verification.
Prompt instructions and semantic linting guide difficulty selection, but quality advice
is not universally a blocking gate. Some optional authoring policies require explicit
configuration. In particular, matching a near-miss entity does not prove that it violates
exactly one constraint, and absence of a lexical probe is not proof of semantic
unsolvability. Such claims require execution and independent assessment.

\paragraph{Rewriting with preservation constraints.}
Scratch generation authors from source directions; regeneration can restructure an
existing task and rederive factual values; conversion treats supplied rubric intent as
an alignment constraint. On a feedback round, the prompt distinguishes difficulty
adjustment from corrective rewriting. A corrective instruction takes precedence while
compatible scenario anchors and previously established requirements are carried forward.
The new specification passes through rubric generation and emission again. Maintaining
an earlier scenario is an instruction and provenance obligation, not a proven guarantee
that a model preserves every semantic constraint.

\paragraph{Layered checks and their execution scope.}
Emission distinguishes structural errors from quality advice and an absent rubric.
A package can be emitted with a missing rubric warning, but the verifier rejects it as
ungradable. The normal batch build invokes the verifier after emission and can route
selected failure categories to rebuilding. The calibration loop invokes grounding
verification before acceptance or salvage, after the solver--judge measurements. Checks cover reachable materials, declared
reference sources and values, supported leakage patterns, and declared semantic twists.
Warnings expose unreadable formats and unsupported checks. Model-based rubric review
and recomputation serve additional roles; a registered optional quality hook must not be
reported as a gate on every construction run. The study records which checks actually
executed for each version and requires independent assessment of final usability.

\subsection{Agentic judging and score admission}
\label{app:judge-admission}
\paragraph{Evidence access and tool use.}
The construction loop calls a mounted scoring entry point with the emitted item and
solver run directory. The judge receives private rubric and reference bindings plus
access to the produced workspace, tool events, and conversation. Its configured tools
include document readers, spreadsheet comparison, trace and transcript readers, and
optional visual or external-fact checks. The inspected judge tool surface has no general code executor or automatic
computed-specification checker; declared metrics and artifact paths are context for
inspection, while the construction loop has a separate recomputation guard.
Tool availability does not establish that a particular ballot used a tool: saved tool calls and evidence anchors provide that record.
A required visual surface or process transcript can be unavailable, in which case the
criterion is excluded rather than assigned an invented outcome.

\paragraph{Ballots and evidence-sensitive aggregation.}
In the batch route, one judge conversation can inspect evidence for multiple criteria;
separate invocations supply ballots, without an independence guarantee.
For a binary criterion $c$ that reaches aggregation in solver run $a$, let
$\mathcal B_c$ contain normalized returned records from non-spoiled ballots that
include that criterion. Invocation errors spoil a ballot; omission of $c$ is an
abstention for $c$. Counts $n_c^+$ and $n_c^-$ record pass and fail votes.
Let $d_c^+=1$ when a passing record is labeled deep, and $d_c^-=1$ when a failing
record is labeled deep and has a nonblank evidence anchor. Otherwise these flags are zero.
For criteria not already excluded for unavailable evidence ($\mathrm E$), the binary aggregation rule used in Equation~\ref{eq:criterion-score} is
\begin{equation}
\widehat y_{a,c}=\begin{cases}
\mathrm M, & |\mathcal B_c|=0,\\
0, & n_c^+\le n_c^-\ \text{or}\ (d_c^-=1\ \land\ d_c^+=0),\\
1, & \text{otherwise}.
\end{cases}
\label{eq:evidence-sensitive-vote}
\end{equation}
Thus ties fail, and a pass majority can be overridden by an anchored deep dissent
when no passing record is deep. The design prioritizes a localized, anchored failure
over unelaborated agreement: a majority may overlook one decisive defect. This
asymmetry can also amplify a mistaken dissent; its advantage over majority voting
has not been established. These labels and anchors are judge-reported, so the rule
does not independently validate their truth. An unavailable evidence
surface can also exclude a criterion before voting. Criterion indices distinguish
authored rubric entries even when they share a capability category. Comparative
criteria instead use a configured ordinal route whose aggregation rule is not
specified by the available archive. Equations~\ref{eq:criterion-score}
and~\ref{eq:evidence-sensitive-vote} therefore define binary scoring only;
the supplement cannot reproduce ordinal aggregation. The notation separates scientific states even though native records serialize
both $\mathrm E$ and $\mathrm M$ as \texttt{excluded}; their reason flags determine
completeness. It does not change any recorded outcome.

\begin{aabtable}[!htbp]
\centering\small
\caption{\textbf{Binary scoring states and propagation in the inspected implementation.}
Native completeness concerns criterion outcomes, not full planned-ballot coverage.}
\label{tab:score-states}
\begin{tabular}{@{}p{0.24\linewidth}p{0.34\linewidth}p{0.35\linewidth}@{}}
\toprule
Observed state & Criterion / native summary & Admission consequence \\
\midrule
Some missing or failed ballots; at least one valid criterion vote & Aggregate surviving votes by Equation~\ref{eq:evidence-sensitive-vote}. The criterion remains in $D_a$. & Lost ballots alone need not set incompleteness; report planned and observed votes separately. \\
All surviving ballots omit a criterion & $\mathrm M$; no-verdict exclusion, absent from $D_a$. & Reported missing judgment sets incompleteness and prevents native admission. \\
All judge ballots fail & $\mathrm M$ for every affected criterion; error exclusion, absent from $D_a$. & Reported call failure sets incompleteness and prevents native admission. \\
Required evidence unavailable & $\mathrm E$; evidence exclusion, absent from $D_a$. & Evidence-only exclusions need not lower reported completeness; admission may cover less than $W$. \\
Zero judged weight & Conditional ratio undefined; native stored value 0.0. & No explicit positive-weight admission check. Evidence-only exclusion can pass the summary checks; 0.0 must not be read as all criteria failed. \\
Supplementary frozen study & Native pass/fail become 1/0; all other criterion states remain NULL. & Locked-weight score requires $D_a=W$. The final accepted-score field remains NULL; coverage alone does not promote it. \\
\bottomrule
\end{tabular}
\end{aabtable}

\paragraph{Native completeness and admission.}
The summary counts graded criteria $G$, budget-excluded criteria $B$, reported
call-error criteria $F$, and no-verdict criteria $N$. Its data-completeness field is
$\operatorname{round}_3(G/(G+B+F+N))$, or 1.0 when that denominator is zero;
these are reason-flag counts, not rubric weights. The incomplete flag is
$F+N>0$. Evidence-only exclusions contribute to none of these counts.
Native admission requires the expected typed fields, $B=0$, an incomplete flag of
false, reported completeness at least 1.0, and a numeric score in $[0,1]$.
It also rejects identifiable all-fallback judging paths. Missing path records
cannot authenticate actual judging, and the wrapper does not independently
recount ballots or require $D_a>0$. Thus native admission is a conditional-score
contract, not a full-coverage or accuracy guarantee. A calibration vector additionally
requires an admitted score for each configured solver.

\paragraph{Supplementary accepted score and validity.}
The frozen-package study retains the full authored denominator. Its locked-weight
score is NULL unless every criterion weight has a valid numeric verdict and a bound
native aggregate. Credible scoring $V$ additionally requires the frozen evaluator
configuration, linked evaluator views, calls and outputs, valid evidence anchors,
and observed evidence/answer/rubric support $B$. Complete coverage or agreement
between AI assessors alone establishes neither $B$ nor $V$. Independent human
calibration remains a scientific limitation, not a prerequisite added to this
operational definition; shared solver/judge model-family bias remains unresolved.

Structural validity $S$, observed execution $E$, evidence support $B$, model-assessed
package quality $Q$, and credible scoring $V$ are separate endpoints. The retained
case records support the reported $Q$ statuses and their stated rationales, but the
distributed material does not give a complete rule for combining assessors' valid,
missing, or conflicting opinions. We therefore reproduce the frozen status field
without reconstructing that rule or reclassifying any package. This reporting
boundary is separate from the later common-reference assessment of one package
(Appendix~\ref{app:conditional-common14-result}).
The usable-package conjunction $Y=S\land E\land B\land Q\land V$ is zero if any
observed component is zero, one only if all are one, and otherwise NULL.
A delivered file alone does not establish the full execution endpoint or package
validity. Appendix~\ref{app:terminal-record} gives the retained frozen outcomes;
none is promoted by this definition, and historical native scores are not
reclassified as supplementary accepted scores.

Judge outputs guide construction; the fixed-output controls
in Appendix~\ref{app:judge-controls} separately probe criterion-specific scoring behavior.

\Needspace{16\baselineskip}
\subsection{Reproducing the distributed analyses}
\label{app:public-reproduction}
The accompanying analysis archive (\path{anc/analysis/}) supports offline checks of
archived inputs, scorer-control artifacts and recorded ballots, arithmetic, and
controller fixtures. From that directory, install the declared dependencies and run:
\begin{quote}
\small
\begin{verbatim}
python -m pip install -r requirements.txt
python diagnostic/verify_arithmetic.py
python -B scorer_controls/verify.py
python analysis_source/adjacent_revision/excerpts/verify.py
python -B controller_checks/reproduce.py \
  --output LOCAL_CHECKS.json
\end{verbatim}
\end{quote}
Separately, \path{anc/figure_source/current_dailylife/plot.py} reconstructs
Figure~\ref{fig:current-dailylife} as a sorted subset of 17 complete platform-report
associations; the canonical 26-row aggregate records remain unchanged. Domains retain
20260710 exercise versions and original report precision. Nine unverified complete
breakdowns are omitted only from the figure. Associations do not independently attest request parameters or backend
weights. Reuse is retained as lineage, not independent replication.
Archived raw model-comparison code is retained for
provenance and is not a public figure input. The case-excerpt verifier checks distributed quotations, translation
hashes, and locators as artifact integrity; it does not reproduce the construction
pipeline or authenticate withheld originals. The arithmetic check verifies the duration
sum. Scorer checks inspect synthetic originals and redacted native observations;
private originals of redacted records remain withheld. Optional \verb|--render|
requires PyMuPDF for local PDF raster checks. Native judging is not rerun.
Controller replay compares all fixture observations with the recorded production-source
results. Its literal fixtures exclude semantic question matching and the absent-fetch-index
boundary. These commands make no model calls or private-service requests; dependency
installation may require network access. Renderer versions and PDF metadata may change
figure bytes without changing the numerical comparisons.

The archive's \texttt{SHA256SUMS.json} records a digest for each distributed file except
itself. The separate \texttt{reference/release-inventory.json} supplies anonymized package
identifiers, construction modes, domains, and archived artifact hashes.
\texttt{endorsement/reproduce.py} recomputes the pooled endorsement rate, item-macro
rate, and cluster-bootstrap interval from anonymous per-item counts. Individual review
records and the private task corpus are not distributed. These offline checks reproduce
reported analyses and controls, rather than the full construction pipeline.

The release audit in Appendix~\ref{app:release-audit} has a standard-library-only
aggregate replay. From \path{anc/analysis/}, run:
\begin{quote}\small
\begin{verbatim}
cd supplement/release20
python -B reproduce.py --check
\end{verbatim}
\end{quote}
Its full inventory, fixed sample, anonymized receipts and file hashes preserve all
failures and unknowns. Raw task/corpus files and complete environments are not
distributed; these receipts do not substitute for runnable full-task inputs.

The released contribution is therefore documentary analysis and controller fixtures,
not an end-to-end benchmark baseline or an executable full-task release.
The accompanying analysis archive, manuscript source, and seven P017 content projections have
different scopes (Appendix~\ref{app:archive-verification}); a review packet may omit
archival provenance without establishing its absence from the original archive.
No general public access route to the private corpus is documented. Access would
require the source holder's permission separately for the original material and any
derived task; this paper does not promise such access or authorize redistribution.

\subsection{A final-version diagnostic audit}
\label{app:release-contract}
The audit connects a selected package's required outcomes to accessible evidence,
reference derivations, and the scope of its grades. Its unit is one frozen package
and declared solver--judge configuration. Recorded failures motivate the following
procedure for evaluating the shared-evidence obligations in
Section~\ref{sec:saved-contract}.

\paragraph{Freeze evidence and scoring obligations together.}
For each required output, bind the criterion identifier and positive weight to the
fact it requires, its permitted evidence locator or disclosure route, and the
reference derivation. Check access through the solver's interfaces and reconstruct
the target under the stated period and assumptions. Mark each obligation supported,
contradicted, or unobserved, retaining the counterexample or missing input
(Appendix~\ref{app:terminal-record}). Freeze digests for instructions, materials,
search corpus, disclosure rules and state, references, rubric, and tools; bind
prompts, runtime, requested and observed model labels, sampling, seed status,
token/time limits, retries, concurrency, ballots, and aggregation to that version.
Unavailable settings remain explicit; a changed component defines a new measured
object.

\paragraph{An inspectable worked obligation.}
\label{app:worked-obligation}
The fixed-output document control provides a complete documentary trace for J7,
the obligation that body text and inventory cells remain inside the visible
page and unobscured. Its weight is 10. The supplied synthetic
\path{source_materials/stock.csv} contains items A and B with counts 12 and 7;
the instruction requires a two-page PDF. The visibility target follows from
that instruction and the rubric's render-based check, not from the judge's
eventual verdict. Reading the CSV or extracting PDF text alone cannot establish
the required visual property.

The normal and masked final PDFs have separate content digests. The latter
places an opaque layer over the data cells on both pages while retaining
extractable text. Thus its final artifact contradicts J7 even though the numbers
remain recoverable as text. The retained native record has three J7 passes for
the normal PDF and three failures for the masked PDF. J8 independently tests
the headings and section boundaries and passes in both conditions.
The audit decision is therefore a J7 failure on the masked artifact, not a
failure of source availability or of the separate section-marker obligation.

The accompanying \path{anc/analysis/supplement/worked_obligation/WORKED-AUDIT.json} binds each
step to the exact instruction, rubric, CSV, final PDFs, prepared page renders,
and recorded ballots in \path{anc/analysis/analysis_source/scorer_controls/}. The full control
inventory retains all five variants and all 24 planned criterion ballots,
including the no-operation HTML timeout. This worked trace reexpresses the
earlier synthetic diagnostic; it adds no judge call, independent final-package
assessment, or accepted score for a real-task subset.

\paragraph{Specify the construction contrast.}
A comparison of feedback-directed revision with independent reconstruction gives
both conditions the same public evidence, task requirements, constructor model,
resource limits, candidate count, and selection rule. Only the feedback condition's
second candidate receives its own first candidate and eligible diagnostic probe.
A missing probe cannot be treated as score feedback. The protocol fixes common
final obligations before construction and requires assessment of each emitted
package against them. Changing its native rubric does not redefine that target.
The contrast concerns diagnostic feedback, separately from Study B's comparison
of joint versus separate authoring organizations.

\paragraph{A concrete, privately runnable input.}
We assembled a private development input from the retained official US-station
GSOD file and its data dictionary: 355 supplied daily records from 2023, with
April and October fixed as the analysis periods. The task requires CSV and Markdown
outputs for an author-proposed retrospective station comparison. It specifies calendar
coverage, missing-value treatment, temperature conversion, precipitation inclusion
rules, and evidence-linked interpretation through 14 common evaluation obligations
with total weight 100. In particular, numeric precipitation availability is
distinguished from a reported amount: an I-coded zero is not evidence of a dry
day, and sums of reports need not cover complete calendar months. Identical
materials and definitions appear in both prepared condition inputs.
The input reuses a project-exposed source block under a private noncommercial
research permission basis; it adds neither an independent sample nor a completed
construction comparison.

\paragraph{Separate reference assessment, selection, and measurement.}
The executable arithmetic check is author-derived, not an independent reference
judgment; an earlier source-only lock also leaves final-package assessment open.
Final-obligation assessment must first reconstruct targets from public evidence
without constructor answers, then compare its locked determinations with that
package. Commit the selected package before final solver measurement, retain all
measurements used for selection, and distinguish fresh execution from rescoring
a stored trajectory. Final measurements cannot be used to choose among candidates.
These records expose provenance without establishing statistical independence.

\paragraph{Keep every attempt and the full scoring scope.}
Retain selected, unselected, failed, incomplete, and nonadmitted candidates and
phases. For binary scoring, report authored weight $W$, pass/fail-judged weight
$D$, and all excluded or missing identities and reasons. Interpret the conditional
score only when $D>0$; a full-authored-scope claim requires $D=W$, with like
obligations for a comparison. At $D=0$, preserve the native stored 0.0 while
marking the ratio undefined. Final-package evidence and the required ballot
coverage must also be established; otherwise the accepted comparison remains
NULL. The prepared input supports this measurement design; diagnostic utility and
calibration remain unmeasured by its arithmetic checks.

\paragraph{Failure-inclusive execution of a frozen development case.}
\label{app:failure-inclusive-development}
A separate prospectively bound trial on 17 October 2026 used an already-exposed
NOAA-data task with frozen input, source, and configuration records. It did not
execute the 2023 station input or the 14-obligation target prepared above. The plan comprised two
solver conditions, two conditionally admitted score phases, and three fixed ballots
per score (six planned slots), with one round and no retries. The requested solver
labels were \texttt{haiku4.5} and \texttt{gpt5.5}; \texttt{gpt-5.6-sol} judged
admitted outputs. Both solvers started. The former terminated at its budget limit
and remains an incomplete attempt with a NULL score; its score phase and three
ballots were not admitted. The latter completed, and all three admitted ballots
were valid, with zero failed ballots. Neither solver attempt was excluded. The
completed condition had a native weighted score of 1 and total weight 71; it did
not supply the missing comparison condition. The native item
\path{autobench-table-noaa-months-82200001} was graded on ten identifiers:
A3, B4, B3, B1, A4, D1, D7, C5, C15, and E4, each with three valid ballots.
The 71 is that native record's scoring denominator, not 71 out of the later
100-weight common target. No O01--O14 assessments or 29 units of missing
common-target weight are inferred from it. The two task identities and scoring
contracts are distinct.

All started phases had reported costs: \$4.03432 for the incomplete solver,
\$2.77664 for the completed solver, and \$5.648585 for the three judge attempts,
totaling \$12.459545. The control record retains the native accounting-completeness
label \texttt{UNKNOWN}, documented as a default label rather than a missing
reported component in this trial. Invoice and helper-serving costs remain unknown.
The plan nominally allocated \$26 (\$4 per solver and \$9 shared across each
three-ballot score), with a \$40 soft reserve. The incomplete solver exceeded its
\$4 threshold by \$0.03432; these controls are not hard invoice caps.

The record therefore preserves an incomplete solver and three conditional
nonadmissions alongside a successfully graded output, rather than reducing the
trial to its successful condition. This is a development example on one exposed
task, not a complete paired comparison or evidence of model separation,
final-version calibration, or benchmark-wide utility.

\paragraph{Saved-specification binding example.}
\par\smallskip\noindent\textbf{A saved restaurant contract.}\par\smallskip
\noindent\begin{minipage}{\linewidth}
\small\setlength{\tabcolsep}{3pt}
\begin{tabular}{@{}p{0.20\linewidth}p{0.75\linewidth}@{}}
\toprule
Evidence & Merchant names, ratings, and per-person prices are unverified; do not invent them. \\
Output & Separate pending itinerary-CSV records; dining-information gaps in the Markdown report. \\
Checks & Y4: no invented facts; E7: explicit CSV uncertainty; C5: explicit report uncertainty. \\
\bottomrule
\end{tabular}
\end{minipage}\par\smallskip
\noindent This saved-specification example illustrates binding, not completed execution.
Its version mismatch precludes borrowing historical scores (Appendix~\ref{app:example}).

\paragraph{Read-only metadata companion.}
The accompanying \path{anc/analysis/supplement/final_version_contract/release_contract.py}
implements a read-only, standard-library metadata check for
\texttt{aab-final-version-record/1}. It checks seven-component identity,
configuration-digest consistency, declared selection/measurement separation,
positive judged weight, exact denominator arithmetic, and exclusion identities.
It rejects incomplete records. Run it on a record with:
\begin{quote}\small
From \path{anc/analysis/}, run:
\begin{verbatim}
python -B supplement/final_version_contract/release_contract.py RECORD.json
python -B -m unittest discover \
  -s supplement/final_version_contract \
  -p test_release_contract.py
\end{verbatim}
\end{quote}
The synthetic tests exercise metadata consistency, and the checker always returns
a NULL accepted score. Obtainability and derivability require the substantive
checks above. File authenticity, semantic correctness, source rights, and the truth
of model/judge records remain outside this metadata check. It executes no agent
and leaves the native controller, binary aggregation, admission, and supplementary
endpoints unchanged. Section~\ref{sec:discussion} specifies the additional evidence
needed for benchmark validation and construction-benefit claims.

\subsection{Development component checks and measurement gaps}
\label{app:module-validation}

Section~\ref{sec:retrospective} characterises constructed packages, scorer behaviour and
solver measurements rather than the individual construction stages. Table~\ref{tab:module-validation}
collects the procedures and available counting information from a development note.
Only screening records a numerator, denominator, and matching rule together.
Rates lacking their observation count or evaluator are not used as performance
evidence; their original reported values remain in the archived development record.
Per-item observations, the model checkpoint, measurement date, and blinding protocol
are unavailable for these checks.

\paragraph{Score definitions.}
\emph{Tagging} compares machine- against human-assigned taxonomy labels field by field: a
single-label field scores $100\%$ on exact match and $0\%$ otherwise, and a multi-label field
earns credit proportional to the correct labels recovered but $0\%$ if any extra label is
emitted. The note does not establish an item-level denominator or an overall
aggregation rule for its nine field summaries. \emph{Screening} compares the stage's three-way
verdict---accept, needs-human, reject---against a human label on each question, counting exact
agreement only. The source note (revision 2752) reports 55/62 exact three-way
matches (88.71\%); this is development agreement, with no established blind protocol
or evaluator independence. It is not independent accuracy. \emph{Rewriting}, \emph{hardening} and \emph{conversion} to multi-turn were
measured on one described $29$-task sample, each under its own weighted composite: rewriting
scores $0.6\times$objective-label hit rate $+\;0.4\times$subjective-label realisation rate;
hardening scores $0.5\times$instruction-change hit rate, then $0.1\times$ each of
main-line/boundary consistency, instruction--material match, difficulty validity and
moderation, clarity/answerability/verifiability, and output completeness; conversion scores
$0.3\times$atomic-requirement coverage $+\;0.25\times$hard-constraint retention
$+\;0.15\times$attachment-reference accuracy $+\;0.15\times$ordering plausibility
$+\;0.1\times$context carry-over $+\;0.05\times$hallucination-free rate. Those three are
composites on different weightings, not pass rates, and are not comparable with one another.
\emph{Rubric generation} reports the share of generated criteria that human
review accepted without edits against all five of description correctness, coverage of the
task's capability points, absence of requirements beyond the task and its frozen evidence,
weight reasonableness, and executability from that evidence; the unit is one criterion.
\emph{Quality validation} ran the eight-dimension checker twice over a described $36$-task
set---once on tasks confirmed defect-free, once after planting defects at recorded locations
with recorded expected dimensions, with out-of-scope plants not counted as hits---so its two
rates have different units and stay separate.

\begin{aabtable}[t]
\centering\small
\caption{\textbf{What the development component records support.}
Observation units and missing evaluation information are kept distinct.
Uninterpretable percentages are omitted from the performance column; the one
fully counted agreement is development evidence, not independent accuracy.}
\label{tab:module-validation}
\begin{tabular}{@{}p{0.15\linewidth}p{0.27\linewidth}p{0.14\linewidth}p{0.34\linewidth}@{}}
\toprule
\raggedright Stage &
\raggedright Observation unit; recorded sample &
\raggedright Quantitative evidence &
\raggedright Counting status \tabularnewline
\midrule
\raggedright Tagging &
\raggedright Taxonomy label field, $9$ fields; item-level $N$ not recorded &
\raggedright No item-level estimate &
\raggedright Partial-credit agreement per field, not task success \tabularnewline
\addlinespace[2pt]
\raggedright Screening &
\raggedright Question; $N=62$ &
\raggedright $88.71\%$ verdict agreement ($55/62$) &
\raggedright Numerator, denominator and match criterion all recorded. No confusion matrix, so
no per-class rates \tabularnewline
\addlinespace[2pt]
\raggedright Rewriting &
\raggedright Task, described $29$-task sample; per-operation $N$ not recorded &
\raggedright Not used as performance evidence &
\raggedright Two-component weighted composite; evaluator not recorded \tabularnewline
\addlinespace[2pt]
\raggedright Hardening &
\raggedright As above, same described sample &
\raggedright Not used as performance evidence &
\raggedright Six-component weighted composite; evaluator not recorded \tabularnewline
\addlinespace[2pt]
\raggedright Conversion to multi-turn &
\raggedright As above, same described sample &
\raggedright Not used as performance evidence &
\raggedright Six-component weighted composite, not a conversion success rate; evaluator not
recorded \tabularnewline
\addlinespace[2pt]
\raggedright Rubric generation &
\raggedright Generated criterion, from rubrics for $27$ sampled tasks &
\raggedright No rate with a known denominator &
\raggedright Criterion count not recorded; the $27$ tasks cannot replace that denominator \tabularnewline
\addlinespace[2pt]
\raggedright Quality validation, clean controls &
\raggedright Defect-free task in the described $36$-task set; clean subset not recorded &
\raggedright No false-positive-rate bound &
\raggedright Control-set size not recorded; a false-positive bound cannot be recovered
\tabularnewline
\addlinespace[2pt]
\raggedright Quality validation, planted defects &
\raggedright Planted defect in the same set; plant count not recorded &
\raggedright No rate with a known denominator &
\raggedright Plant count and dimensional composition are not recorded; no detection rate follows \tabularnewline
\bottomrule
\end{tabular}
\end{aabtable}

The note records no quantified check for the remaining stages: expansion is marked untested in
that development cycle, environment construction is described as basic functionality only
(local database generation, mocked local pages, one local search service), and intake, emit
and the external runner carry only qualitative statements.

These development cohorts remain separate from the historical endorsement
records (Appendix~\ref{app:human-agreement}) and the fixed release sample
(Appendix~\ref{app:release-audit}).
None of the rows above identifies end-to-end construction yield or judge accuracy against
independent reference judgments.

\subsection{Source-policy identifiability}
\label{app:source_policy_identifiability}
This source-specific argument concerns an authored fictional repair-cafe
allocation task, not one of the observed Daily-Life package failures or the
precipitation reporting-rule counterexample in Section~\ref{sec:local-controls}.
It asks whether the frozen input can distinguish policy rules, with the
downstream scheduler held fixed; it does not compute a new allocation or
evaluate a constructed package.

We examined whether allocation-table correctness on the frozen authored source can distinguish preservation of mandatory policy rules from alternative rules that give the same answers. The source requires output ordered by \texttt{arrival\_order} and then \texttt{request\_id}. Its physical request order already matches this ordering. Consequently, sorting by the declared keys and retaining the physical row order produce identical output sequences on this input. The fallback \texttt{request\_id} comparator is also inactive because the source contains no tied \texttt{arrival\_order} values. Correct output order alone therefore does not identify which rule the package retained.

Rejection priority has a related identifiability limit. The source gives withdrawal precedence over unsafe requests, followed by unsupported item classes. Each frozen record activates at most one rejection predicate. Reordering these predicates therefore preserves every rejection label and the eligible-request set. With the downstream scheduling procedure held fixed, the alternative classifiers can produce identical allocation tables. This argument does not require computing a reference allocation. It concerns the precedence of overlapping rejection conditions, not omission of a rejection predicate: every declared rejection type has an instance in the source.

\paragraph{Scope of the supplied computational attachments.}
The accompanying analysis archive is under \path{anc/analysis/}. Domain-support
source data and exact-certificate material are under \path{anc/domain-support/};
the separate symbolic perturbation construction is under
\path{anc/boundary-support/construct_boundary_support.py}. The retained selection
functionals and machine-state recoding analyses are under
\path{anc/selection-functionals/} and \path{anc/verdict-recoding/}, respectively.
These materials support the documented analysis,
controller, and certificate checks. They do not include the private construction
and scoring archives or the full-task environments described above, and they do
not reproduce the end-to-end construction pipeline. Descriptions of archived
commands do not grant access to withheld task data or private services; source
and derived-task redistribution permissions remain separate requirements.

\section{Daily-Life Instantiation and Task Examples}
\label{app:example}
\subsection{Audited collection: composition and selection}
The daily-life application is an instantiation of BenchMint. Its source-processing,
authoring, execution, and evaluation artifacts belong to the system described in this
paper.

\paragraph{Archived release composition and provenance.}
The audited archival release contains \num{pre.n_items} packages across \num{pre.n_domains}
domains, each with metadata, a seed, an instruction, a rubric, reference files, and an
identifier-matched saved authoring specification. Saved seeds record
\num{pre.n_scratch} scratch, \num{pre.n_regen} regeneration, and \num{pre.n_convert}
conversion outputs. The supplement's \texttt{reference/release-inventory.json}
provides anonymous package identifiers, domains, construction modes, and artifact hashes.
The inventory documents archived versions; it does not include the private task materials.

\paragraph{Selection and measurement scope.}
We keep implementation snapshots, release cohorts, and execution versions
explicit because measurements collected under different configurations should not
be pooled merely because they share the system name. The archival release includes
human-supplied materials and selection using observed
model scores and domain quotas. Its size measures selected production outputs, not
source-to-output yield or independently assessed usability. The capability matrix uses
a separate earlier evaluation cohort with a different domain distribution.

\subsection{Fixed-sample offline release audit}
\label{app:release-audit}
To check final-release traceability beyond illustrative tasks, we froze a sample of
\num{release20.sampled} packages from the actual \num{release20.population}-package
inventory before inspecting task contents. We stratified by domain and generation
mode; interaction mode was absent from the inventory. Allocation used
$n_h=\lfloor \num{release20.sampled}N_h/\num{release20.population}\rfloor$, assigning the remainder by descending fractional part
then ascending SHA256 of \texttt{seed:domain:generation\_mode}. Within each stratum,
we selected the smallest SHA256 of \texttt{seed:public\_id:saved\_spec\_sha256}, with
seed \texttt{r209-release20-v1}, no replacement and no outcome-dependent selection.
All \num{release20.unsampled} unsampled packages remain explicitly \texttt{NOT\_RUN}
for this supplementary audit, not for their historical construction or platform use.
Core files, saved specifications and solution-tree digests matched the inventory for
every selected package before execution; the historical field named
\texttt{reference\_tree\_sha256} binds the solution tree.

Table~\ref{tab:release-audit} is the canonical package-status table.

\begin{table}[htbp]
\centering\small
\caption{\textbf{Mechanical audit of the fixed 20-package sample.}
P/F/U mean PASS/FAIL/UNKNOWN; each row has denominator 20 and grouped checks
are separate. The other 80 packages are NOT\_RUN in this audit.
Opening \num{release20.formats.PASS}/\num{release20.formats.supported_locations}
checked file locations leaves \num{release20.formats.unsupported_locations}
unsupported-format locations uncovered.}
\label{tab:release-audit}
\begin{tabular}{@{}p{0.61\linewidth}rrr@{}}
\toprule
Check & P & F & U \\
\midrule
Core/spec/solution binding & \num{release20.package_checks.mapping_authentication.PASS} & \num{release20.package_checks.mapping_authentication.FAIL} & \num{release20.package_checks.mapping_authentication.UNKNOWN} \\
Instruction, rubric, spec structure (each) & \num{release20.package_checks.instruction_structure.PASS} & \num{release20.package_checks.instruction_structure.FAIL} & \num{release20.package_checks.instruction_structure.UNKNOWN} \\
Literal local search & \num{release20.package_checks.production_local_mcp_search.PASS} & \num{release20.package_checks.production_local_mcp_search.FAIL} & \num{release20.package_checks.production_local_mcp_search.UNKNOWN} \\
Literal local fetch & \num{release20.package_checks.production_local_mcp_fetch.PASS} & \num{release20.package_checks.production_local_mcp_fetch.FAIL} & \num{release20.package_checks.production_local_mcp_fetch.UNKNOWN} \\
Complete format opening & \num{release20.package_checks.safe_file_openability.PASS} & \num{release20.package_checks.safe_file_openability.FAIL} & \num{release20.package_checks.safe_file_openability.UNKNOWN} \\
Deployed workspace reachability & \num{release20.package_checks.public_workspace_reachability.PASS} & \num{release20.package_checks.public_workspace_reachability.FAIL} & \num{release20.package_checks.public_workspace_reachability.UNKNOWN} \\
Complete evidence binding & \num{release20.package_checks.deterministic_evidence_binding.PASS} & \num{release20.package_checks.deterministic_evidence_binding.FAIL} & \num{release20.package_checks.deterministic_evidence_binding.UNKNOWN} \\
Complete rubric/answer consistency & \num{release20.package_checks.deterministic_rubric_answer_consistency.PASS} & \num{release20.package_checks.deterministic_rubric_answer_consistency.FAIL} & \num{release20.package_checks.deterministic_rubric_answer_consistency.UNKNOWN} \\
Semantics, dialogue, services, feasibility (each) & \num{release20.package_checks.semantic_correctness.PASS} & \num{release20.package_checks.semantic_correctness.FAIL} & \num{release20.package_checks.semantic_correctness.UNKNOWN} \\
\bottomrule
\end{tabular}
\end{table}

Direct calls to the pinned production classes used sorted literal index keys for
search and the union of indexed and returned URLs for fetch. Success requires a
nonempty result list or returned text, respectively. All
\num{release20.operations.search.total} searches were nonempty; of
\num{release20.operations.fetch.total} fetches,
\num{release20.operations.fetch.PASS} returned nonempty text. P027's remaining fetch
returned an empty string. P017, P056 and P076 have no local indices and remain unknown.
PDF/CSV/JSON/XLSX file-opening checks passed at
\num{release20.formats.PASS} of \num{release20.formats.supported_locations} file
locations; P054's CSV did not open under the tested UTF-8 encoding;
alternative encodings were not assessed. Another \num{release20.formats.unsupported_locations}
locations had unsupported formats. Locations can repeat content; calls and files
are not independent task observations. No result-selected retesting was added.

The supplement preserves the full sample-status table, anonymized operation receipts,
hashes and sampling metadata. Public replay reaggregates these records; it does not rerun the withheld raw
corpus or provision task environments. For this fixed-sample offline audit, deployed
reachability and semantic usability remain unresolved; these checks do not establish
whether the same tasks ran in their original platform environments. This audit is neither human review nor a measure of judge accuracy,
construction yield or causal construction benefit.

\subsection{Supplementary final-package outcomes and assessment coverage}
\label{app:terminal-record}
The supplementary study (Study A) retains its original denominator of 20 frozen
packages. At the P050 evidence freeze, 14 opportunities are consumed:
P099, P090, P068, P039, P047, P060, P087, P085, P017, P019, P001, P006, P003, and P050.
\label{app:unobserved-freeze}
The remaining six have no supplementary execution observation in this manuscript; no rerun or replacement
is added here. These time-specific observations do not revise the earlier mechanical
snapshot in Table~\ref{tab:release-audit}.
\label{app:early-frozen}
P099 and P090 retain failed or unsupported execution outcomes and NULL scores;
P090 has no eligible final reply and its full weight 152 remains unjudged.
Earlier P039 and P047 failure/NULL records are also retained; the completed P047
observation below does not overwrite them. None of these prior records contributes
a positive success or validity label. In the separate paired
construction study, seven of eight original opportunities are consumed; no accepted
score is established, and source-3 JC remains unstarted.
These observations do not establish comparative construction benefit.

\begin{table}[!htb]
\centering\footnotesize
\setlength{\tabcolsep}{2pt}
\caption{Fixed Study A scope at the P050 paper freeze, in original order: 14 consumed opportunities, not 14 successes. Each task appears once.}
\label{tab:study-a-scope}
\begin{tabular}{@{}llcll@{}}
\toprule
Task & Obs. & S/E/B/Q/V/Y & Judged/authored & Page \\
\midrule
P099 & C* & N/N/N/N/N/N & ?/? & \pageref{app:early-frozen} \\
P090 & C* & N/N/N/N/N/N & 0/152 & \pageref{app:early-frozen} \\
P068 & C & N/N/N/0/N/0 & 100/124 & \pageref{app:p068-observation} \\
P039 & C* & N/N/N/N/N/N & ?/? & \pageref{app:early-frozen} \\
P047 & C & N/N/N/0/N/0 & 81/135 & \pageref{app:p047-reference} \\
P060 & C & 1/1/N/N/N/N & 40/91 & \pageref{app:p060-observation} \\
P087 & C & 1/1/N/0/N/0 & 109/109 & \pageref{app:p087-observation} \\
P085 & C & 1/N/N/N/N/N & 0/145 & \pageref{app:p085-observation} \\
P017 & C & 1/1/N/N/N/N & 104/104 n; 97/104 e & \pageref{app:p017-observation} \\
P019 & C & 1/N/N/N/N/N & 0/169 & \pageref{app:p019-observation} \\
P001 & C & 1/1/N/N/N/N & 17/125 & \pageref{app:p001-observation} \\
P006 & C & 1/1/N/0/N/0 & 121/121 & \pageref{app:p006-observation} \\
P003 & C & 1/N/N/N/N/N & 0/128 & \pageref{app:p003-observation} \\
P050 & C & 1/1/N/N/N/N & 108/108 & \pageref{app:p050-observation} \\
P054 & NR & N/N/N/N/N/N & ?/? & \pageref{app:unobserved-freeze} \\
P056 & NR & N/N/N/N/N/N & ?/? & \pageref{app:unobserved-freeze} \\
P037 & NR & N/N/N/N/N/N & ?/? & \pageref{app:unobserved-freeze} \\
P033 & NR & N/N/N/N/N/N & ?/? & \pageref{app:unobserved-freeze} \\
P027 & NR & N/N/N/N/N/N & ?/? & \pageref{app:unobserved-freeze} \\
P076 & NR & N/N/N/N/N/N & ?/? & \pageref{app:unobserved-freeze} \\
\bottomrule
\end{tabular}
\par\smallskip\raggedright
C = consumed; C* retains the early failed/unsupported or failure/NULL observation (P099, P090, P039), without inferring an endpoint zero.
NR = NOT\_RUN: no supplementary execution observation at this freeze, not task failure.
N = NULL; ? = UNKNOWN weight, not zero. Accepted scores and human-truth labels are NULL throughout.
S/E/B/Q/V/Y denote structure, execution, evidence support, package quality, credible scoring, and protocol conjunction; 1/0 are recorded endpoint values, not human validation.
P017's n/e coverage denotes native/evidence-revalidated weights for one task; its original E=NULL and separate offline E=1 are retained on the cited page.
The older mechanical snapshot (Table~\ref{tab:release-audit}) is unchanged; its UNKNOWN checks are not these supplementary endpoints.
P047/P068 retain their partial assessor-view limitations. No rows, exposures, or versions are pooled.
Raw pass/fail/unjudged weights remain in Table~\ref{tab:study-a-current}.
\end{table}

The supplementary frozen P068 execution,\phantomsection\label{app:p068-observation}
 an air-fryer fact-checking task, produced
its Markdown artifact and self-contained handoff. All 47 pages of its three supplied
PDFs were visible in solver observations. This execution differs from the historical
P068 case in Section~\ref{sec:historical-execution}; their outcomes are not pooled.
The supplementary frozen P068 scoring retains nine pass, two fail, and three NULL
verdicts over 14 criteria: 124 total weight, 100 judged, 80 positive, and 24 unjudged.
Its summary 0.8 is conditional on graded weight; its accepted score remains NULL.
No replacement aggregate is reported. The three NULL criteria are C5 (weight 10),
E7 (7), and C13 (7); preserved criterion evidence also identifies two omitted safety
instructions for C13 without changing the original NULL assignment.

Both isolated AI assessors identify J2's local-only prohibition as inconsistent with
the task and workspace permissions for authoritative supplementation. The native J2
failure is retained. Both applicability labels are $Q=0$, yielding the frozen
conjunction $Y=0$; remaining success and validity fields stay NULL. Their supplied
views omit required PDF/source contents and per-file public access mappings, leaving
feasibility, reference correctness, and evidence obtainability unresolved.
Solver traces were intentionally excluded from package-quality assessment.
Missing source observations limit assessment coverage and do not prove impossibility.
The assessors used the same model family, with isolated contexts and without each
other's ballots or native verdicts; they supply neither human truth nor independent
scoring accuracy. The public analysis bundle does not reproduce these executions.

\paragraph{Completed frozen P047 execution and reference identity.}
\label{app:p047-reference}
P047 produced the required Markdown file and an 8,051-byte final handoff.
All 67 recorded model responses completed: nine solver, 56 judge, and two
package-assessment responses. Response completion is not criterion coverage.
Table~\ref{tab:study-a-current} reports the weighted outcomes without rescoring.
The six unjudged criteria remain J9 (10), D7 (10), C5 (10), C15 (7), J6 (7),
and J2 (10). The bounds assign only that unjudged weight to zero or one; they are
identification bounds for the raw rubric sum, not confidence intervals or a
validity-adjusted score.

\begin{table}[!htb]
\centering\footnotesize
\setlength{\tabcolsep}{3pt}
\caption{Final-package endpoints and scoring coverage. P/F/U are pass/fail/unjudged
weights; S/E mean structure/execution and Q/Y mean package quality/protocol
conjunction. JC is joint construction and GV is separate generation and validation; N means NULL, never zero. All accepted scores are NULL. Study A retains
20 tasks (14 consumed through P050); Study B retains eight opportunities (seven consumed).}
\label{tab:study-a-current}\label{tab:frozen-p047}
\begin{tabular}{@{}lrrrrcc@{}}
\toprule
Case & P & F & U & Judged/authored & S/E & Q/Y \\
\midrule
P047 & 71 & 10 & 54 & 81/135 & N/N & 0/0 \\
P068 & 80 & 20 & 24 & 100/124 & N/N & 0/0 \\
P060 & 30 & 10 & 51 & 40/91 & 1/1 & N/N \\
P087 & 62 & 47 & 0 & 109/109 & 1/1 & 0/0 \\
P017 native & 44 & 60 & 0 & 104/104 & 1/1 & N/N \\
P017 evidence & 37 & 60 & 7 & 97/104 & 1/1 & N/N \\
P085 & 0 & 0 & 145 & 0/145 & 1/N & N/N \\
P019 & 0 & 0 & 169 & 0/169 & 1/N & N/N \\
P001 & 0 & 17 & 108 & 17/125 & 1/1 & N/N \\
P006 & 101 & 20 & 0 & 121/121 & 1/1 & 0/0 \\
P003 & 0 & 0 & 128 & 0/128 & 1/N & N/N \\
P050 & 64 & 44 & 0 & 108/108 & 1/1 & N/N \\
Source-2 JC & N & N & N & N & N/N & N/N \\
Source-2 GV & 47 & 7 & 101 & 54/155 & 1/1 & N/N \\
Source-3 GV & N & N & N & N & N/N & N/N \\
\bottomrule
\end{tabular}
\par\smallskip\raggedright
$B$ (evidence support) and $V$ (credible scoring) are NULL throughout.
P017's two rows describe one attempt: native grades versus revalidated evidence
coverage; $E=1$ is the separate offline observation. Its native ratio is $44/104$;
evidence bounds are $[37/104,44/104]$. Neither is an accepted score.
For a given row, conditional score is P/(P+F) when P+F$>0$; bounds are
[P/full,(P+U)/full]. P001's bounds are $[0,108/125]$; its conditional zero
is not a full-task score. Rows are not pooled; earlier failures remain in
Appendix~\ref{app:terminal-record}. P006/P003/P050 details appear below. P050's coincident bounds $[64/108,64/108]$ describe native coverage, not semantic-validity confidence.
\end{table}

Table~\ref{tab:study-a-current} collects the supplementary outcomes. P047's raw
conditional ratio is $71/81$, with bounds $[71/135,125/135]$; P068's are $80/100$
and $[80/124,104/124]$. Neither conditional ratio is an accepted score.

The frozen government capture is a portal homepage, whereas the final reference
and rubric require the identity of a specific deductions-policy document.
The reference also retains a precise date omitted by the final instruction.
The inspected captures exactly match the original frozen release bytes.
An earlier saved specification describes captures with different hash prefixes
and sizes; those versions cannot substitute for the tested sources.
This is a tested reference/source-identity counterexample despite successful file
delivery, not evidence of failure to execute or proof that the task is impossible.

Two isolated AI package assessors identify reference-correctness and
criterion-applicability violations. Each ballot has 18 anchors checked against
original bytes; evidence obtainability and feasibility remain unknown.
They are separate model requests, not independent human labels or truth;
shared model-family bias and the validity requirements in
Appendix~\ref{app:judge-admission} remain. The original $Q=0$, $Y=0$, and
$S/E/B/V=\mathrm{NULL}$ are unchanged. Historical P047's $20/101/135$ accounting
in Table~\ref{tab:historical-execution} belongs to a different configuration/version
exposure and is not pooled with this result.
The public supplement's \texttt{p047-reference-identity.json} records source and
terminal digests with these aggregate counts, without personal data or private URLs.

\paragraph{Frozen P060: observed execution and partial scoring.}
\label{app:p060-observation}
The October 12 observation used P060's original frozen package and public inputs;
the solver did not receive its private solution or rubric. It produced the required
XLSX and a 6,953-byte final handoff. The delivered workbook was opened and its contents inspected. These observations establish $S=1$ for structure only
and $E=1$ for execution; workbook formula correctness is a separate question.

Four of ten criteria have numeric verdicts: A4, A3, and I12 pass (10 weight each),
and B1 fails (10). B5 (7), G2 (10), G5 (7), C1 (10), J3 (7), and J6 (10) remain
NULL. Thus 51/91 weight is unjudged. The native conditional ratio $30/40=0.75$
is not an accepted score. The raw identification bounds $[30/91,81/91]$ assign
the missing weight to zero or one; they are not confidence intervals.

Evidence support $B$ remains NULL because applicability of canonical food choices
is unresolved: the task permits selecting inventory foods and alternatives,
whereas B1/G2 specify particular choices. The recorded B1 failure is retained;
this mismatch in specificity is neither a demonstrated contradiction nor a
verified formula error. $Q,V,Y$ and the accepted score remain NULL. No package-quality
assessment was observed. This one execution establishes neither executable-task
yield nor scorer reliability, and supplies no result for other cases.
The sanitized \texttt{p060-observation.json} preserves the ten criterion states,
full denominators, output and raw-record digests; private evidence is not distributed.

\paragraph{Frozen P087: public evidence and private source provenance.}
\label{app:p087-observation}
The accepted observation records $S=1,E=1,B=\mathrm{NULL},Q=0,V=\mathrm{NULL},Y=0$;
the accepted score remains NULL. All 13 native criteria are judged, with 109/109
weight covered and raw score $62/109$. This complete native coverage does not
resolve evidence support or scoring validity. Duplicate criterion identifiers are
separately bound by occurrence; they do not demonstrate a scoring error.

The private original seed contains the 5-million-yuan (500 in ten-thousand-yuan
units) operand. The released public prompt does not, whereas the rubric attributes
that operand to a literal in the prompt. Public/private visibility and source-version
alignment are therefore material to criterion applicability. The frozen 12366
capture contains navigation, login and footer text, not the substantive policy
attributed to it; a separate local announcement provides partial policy support.
These observations do not establish that the threshold is legally false or that
the real-world task is impossible. The sales arithmetic is consistent conditional
on the threshold; legal applicability, complete rolling-period circumstances and
exact filing dates remain unresolved. Both isolated AI ballots record $Q=0$;
their agreement is a protocol result, not independent truth. The report is present,
but its date and filing-cycle assertions are not independently verified.

\paragraph{Frozen P085: missing completion and coverage.}
\label{app:p085-observation}
The eighth consumed task (original zero-based index 7) has $S=1$;
$E,B,Q,V,Y$ and the accepted score remain NULL. No episode-final handoff or
final-artifact receipt is established. None of its three required deliverables---a
property-agreement document, legal-assessment and procedure workbook, and
information-gap/self-check document---is available for assessment. Intermediate
dialogue is not a completed task handoff. All 17 native criteria, with total weight
145, remain unjudged: judged weight is 0/145 and no conditional score is defined.
Missing completion and coverage do not establish package invalidity or a zero score.
The later P017 observation below does not revise this missing outcome.

\paragraph{Frozen P017: execution, native grades, and evidence coverage.}
\label{app:p017-observation}
The ninth consumed task (original zero-based index 8) requires a dinner-planning
workbook and a three-month dietary explanation. Its original terminal retains
$S=1$ and $E,B,Q,V,Y$ and accepted score NULL. A separate offline examination
of the same complete public-input view, final response, and exact XLSX/DOCX export
establishes $E=1$: both required files are present and parsable. This observation
neither evaluates spreadsheet formulas nor certifies dietary correctness. It adds
no solver attempt or judge vote and does not overwrite the original terminal.

All 12 native criteria have actual recorded grades: passing weight 44, failing
weight 60, and no missing native grade, totaling 104. Thus the native conditional
ratio is $44/104$. Revalidation asks a different question: whether each grade's
cited evidence satisfies its existing artifact-coverage obligation. A4 (weight 7)
cites the document but lacks the required workbook receipt; earlier workbook reads
do not replace that citation. Its evidence-supported state remains unknown, while
the other 11 criteria cover 97/104 weight (37 pass, 60 fail). The evidence bounds
$[37/104,44/104]$ assign the seven unsupported weight to zero or one; they are
not confidence intervals, missing native votes, or an accepted score. The original
native grades remain unchanged. $B,V,Q,Y$ and accepted score remain NULL.

A source-bound check illustrates why execution is insufficient. The public request
says ``current season'' without a planning date, while the locked rubric selects
summer. Conditional on that selection, the recorded source operands give mean oil
21.0\,g and salt 4.6\,g; the reference workbook labels the corresponding means
28.4 and 21.9. These literal and arithmetic discrepancies leave reference support
and applicability unresolved. They neither determine medical truth nor generalize
to all packages; no reference is rewritten or criterion regraded.
One of the two predeclared package-assessment ballots records reference-correctness
and criterion-applicability violations; the other is unobserved. No aggregate $Q$
or human label is inferred. The historical P017 row in
Table~\ref{tab:historical-execution} is a separate exposure and is not pooled here.
The later P019 and P001 observations retain these separate native and evidence
states; they do not add a P017 attempt or vote.

\paragraph{Frozen P019: budget-limited final handoff.}
\label{app:p019-observation}
The tenth consumed opportunity (original zero-based index 9) is an interactive
learning-goal and seven-day action-planning task. Its one attempt uses a shared
32-request solver budget across dialogue invocations, not a renewed budget per
turn. All 32 requests complete, spanning seven dialogue invocations and 38 tool
actions, but no episode-final handoff follows the last disclosure. Earlier Markdown
and workbook files precede the disclosed daily time and intended practical use;
their existence does not establish a final response to those requirements.
Thus $S=1$, while $E,B,Q,V,Y$ and accepted score remain NULL. All 20 criteria
(169 weight) are unjudged, with no conditional score.

The evaluated dialogue condition retains tool history within each invocation,
user/assistant text across turns, and a persistent filesystem. This specifies the
observed information available to the solver; it does not establish equivalence
to other persistent-session conditions or predict success under them.
For these P019/P001 attempts, the fixed solver and judge are GPT-6-Astra with high
reasoning effort. The one-attempt protocol permits no automatic retry, up to 32
solver requests in 1,800 seconds, 128 judge requests, and one vote per criterion;
solver/judge output limits are 16,384 tokens per request. Missing endpoints are
observations under this condition, not evidence of task impossibility.

\paragraph{Frozen P001: executable outputs and partial verdict coverage.}
\label{app:p001-observation}
The eleventh consumed opportunity (original zero-based index 10) requires a
living-room plan that combines spatial constraints, retained furniture, product
evidence, and changing household needs. Its complete Markdown plan and two distinct
generated layout PNGs are present, with a final handoff at solver call 29.
The copied original floorplan is not a third generated layout. Original $S=1,E=1$
establish required outputs and handoff, not design, child-safety, or reference
correctness. There are 70 actual calls (29 solver, 41 judge): 69 completed and one
without a completed response. Package-quality assessment is unspent; $B,Q,V,Y$
and accepted score remain NULL.

Five native verdict submissions are recorded. Y11 (true), A10 (false), and A2
(false) were excluded because their evidence bindings were not accepted; they
remain excluded. B1 (false, weight 7) and D4 (false, weight 10) are the two accepted
evidence-bound verdicts. They concern an omitted price-difference calculation and
an omitted product-evidence boundary about rounded corners, respectively; these
are recorded judge findings, not independently established truth. F4 has no verdict,
and eight remaining criteria are unattempted. Neither category contributes a vote
or a correctness failure.

Across 14 criteria and 125 total weight, validated coverage is 17/125: zero passing,
17 failing, and 108 unjudged weight. The conditional native aggregate of zero is
not a score over all 125 weight. Assigning the unjudged weight to zero or one gives
identification bounds $[0,108/125]=[0,0.864]$, not a confidence interval.
Later offline attribution checks add no scientific judgments and do not promote
the three excluded submissions. At that observation, 11 of 20 opportunities were consumed;
it neither completes the experiment nor estimates cohort success or judge accuracy.

\paragraph{Frozen P006: full native coverage with a source/reference mismatch.}
\label{app:p006-observation}
The twelfth consumed original opportunity requires a Markdown report distinguishing
what the supplied electrical-service and standards pages support about household
faults. All 63 requests complete: eight solver, 53 judge, and two isolated assessor
requests. The required Markdown and final handoff at solver call 8 establish
$S=1,E=1$. All 14 native criteria, totaling 121 weight, have evidence-bound
verdicts: 101 passing, 20 failing, and zero unjudged weight. The descriptive native
ratio is $101/121=0.8347107438$ (stored rounded aggregate 0.835).
$B,V$ and accepted score remain NULL; $Q=0,Y=0$. There are no human labels.

Both assessor ballots record evidence obtainability 1, reference correctness 0,
criterion applicability 0, and feasibility UNKNOWN. The private oracle specifies
January--June 2026, but the complete frozen 95598 page and native retrieval show
February--July 2026. Both intervals contain six months; equal counts do not resolve
the wrong period. The rubric's six-month count and June headline are separately
compatible with the source. Whether tariff enumeration belongs in this
fault-service boundary report is a separate applicability judgment; matching
quotes does not settle it, and the original ballots remain unchanged.

The two native failures each carry weight 10. E7 concerns omission of the price
module. B1 concerns omission of the four-category arithmetic sum 52,116, despite
the correct subtotal $1,724+452=2,176$ and first-three-category sum
$2,176+46,592+710=49,478$. Adding the fourth category gives
$49,478+2,638=52,116$; this arithmetic does not establish disjoint categories or
a unique-standard count. These are the recorded criterion verdicts, not a new
evaluation of their semantic appropriateness.

Feasibility UNKNOWN describes the evidence available to the isolated assessors;
it does not negate the observed execution endpoint $E=1$. Offline verification
binds each verdict to its criterion, completed response, evidence receipt, and
quoted content. It does not certify semantic correctness, electrical safety,
source authority, or human agreement. At the P006 observation, Study A had consumed 12 of its original 20
opportunities; Study B remains at seven of eight and 115 calls. All earlier
failures, NULLs, and partial-evidence distinctions retain their original denominators.

\paragraph{Frozen P003: report existence without a final handoff.}
\label{app:p003-observation}
Original index 12 (native LIV-0107) is the thirteenth consumed opportunity.
All 32 solver calls completed across seven turns. The unchanged solver allowance
ended before another response to the latest user disclosure: assistant message 40
precedes user message 44. Thus the existing Markdown report is not a final handoff.
$S=1$ binds structure; $E,Q,Y,B,V$, native and accepted scores, and human labels
remain NULL. No judge or package-assessment call was made. All 14 criteria,
totaling 128 weight, remain unjudged: zero judged weight, undefined conditional
score, and coverage bounds $[0,1]$. This is missingness under finite evaluation, not
criterion failures or evidence that the task is impossible.

\paragraph{Frozen P050: complete native coverage with unsettled rule semantics.}
\label{app:p050-observation}
Original index 13 (native LIV-0055) is the fourteenth consumed opportunity.
The task requires a cross-platform digital-device comparison in Markdown.
Exact public-input, private-heldout, source, and final-handoff bindings establish
$S=1,E=1$: the final response is solver ordinal 10 and the required report has
17,515 bytes. The solver sees the instruction and public workspace, with source
access through native search/fetch; the private reference is withheld.
There are 68 calls: ten solver, 56 judge, and two package-assessment calls.
All 12 native criteria have evidence-bound VALUE verdicts: seven pass (64 weight),
five fail (44), and none unjudged, totaling 108. The native ratio
$64/108=0.5925925926$ rounds to 0.593. Its coincident coverage bounds
$[64/108,64/108]$ describe native verdict coverage only, not confidence in semantic
validity. $Q,Y,B,V$, accepted score, and human labels remain NULL.

Every native failure (G4, D8, D1, D2, B5) cites the 0.2375 difference between
alternative discount sequences. The frozen reference gives 787.5075 for earbuds
and 25098.2075 total; the submitted report gives 787.2700 and 25097.97.
G4 also cites education-order differences of 25 for the laptop and 15 for the
monitor. Decimal arithmetic reproduces the alternative formulas; it establishes
neither the correct settlement rule nor optimality. The source permits combined
discounts but leaves relevant internal order and eligibility ambiguous. The final
report explicitly states its order and reports sensitivity. Solver disagreement
does not establish that the reference is wrong; the oracle and native verdicts
remain unchanged.

Assessment call 67 yields one valid anchored ballot: criterion applicability is
violated; feasibility, reference correctness, and evidence obtainability are
UNKNOWN. It questions the compulsory 88 membership fee and the applicability of
exact numeric targets under the source conditions. Call 68 does not yield a valid
second ballot within the shared 300-second assessment window. The protocol
therefore produces no aggregate $Q$; the first ballot is retained as evidence,
not substituted for the missing aggregate. Context-isolated judgments from the
same requested model family do not provide independent human validation.
Exact artifact binding, complete native coverage, package validity, and independent
judge calibration remain four distinct claims. Study A retains all 20 packages;
14 consumed opportunities are not 14 successes.

\paragraph{Construction comparison: retained versions and partial outcomes.}
\label{app:construction-pair}
Study B retains its original eight opportunities and separate v1/v2 strata.
The four earlier consumed opportunities and their NULL/stopped records remain
unchanged. Source 2 contributes two v2 observations: JC uses one physical call and
retains a failed/stopped outcome with $S,E,B,Q,V,Y$ all NULL; GV uses 108 calls
(one generator, one validator, 17 solver and 89 judge calls) and records $S=1,E=1$,
with $B,Q,V,Y$ and full score NULL. Package-quality assessment was not called.
Source-3 GV adds two completed generation/validation calls, followed by failure
of the instruction-preservation contract at package emission. All its endpoints
remain NULL; source-3 JC is unstarted. The failure class supplies no quality score
or certified underlying cause. Seven opportunities are consumed, accounting for
115 physical calls (113 prior plus two); this is a frozen snapshot of eight planned
opportunities, not experiment completion. Call counts are resource accounting,
not independent experimental observations or evidence of comparable realized cost.

Source-2 GV has five passing, one failing and 11 unjudged/excluded criteria out of 17.
Passing weight is 47, failing weight 7 and NULL weight 101, totaling 155; graded
weight is 54. The native $47/54$ conditional ratio excludes 101/155 of the weight.
The full score remains NULL; $[47/155,148/155]$ assigns the missing weight to zero
or one and is a partial-identification interval, not a confidence interval.
The passing criteria are B2, B1\#1, B6, I10 and G4\#2; B1\#2 fails.
A3\#1, A2, D7, G4\#1, B5, C1, C2, J3, J4, J6 and A3\#2 remain NULL.
The delivered spreadsheet and document establish observed execution, not correctness.

This pair changes joint six-surface authoring versus separate generator and validator
organization. It does not isolate the entire adaptive hardening workflow. Changed
rubrics require alignment of the underlying task obligations before criterion-level
comparison; unmatched criteria and missing outcomes cannot support a method ranking.
No accepted-score gain, construction benefit or full-workflow validity follows.

\subsection{Separately frozen comparison and terminal technical continuation}
\label{app:terminal-followup-r797}
This follow-up retains two repeated-source descriptive pairs ($N=2$), with four
original arm--source rows frozen in R783. Its source indices 0 and 1 are local
to this follow-up, not Study B's source-2/source-3 identifiers. Table~\ref{tab:terminal-followup-original}
preserves the original endpoints; no later technical continuation replaces them.
The v3 arm labels call the native GV-v2 and JC-v2 methods, respectively.

\begin{table}[!htbp]
\centering\small
\caption{Original four-row follow-up freeze, separate from Study B.
N denotes NULL, not zero. Calls are durable client admissions.}
\label{tab:terminal-followup-original}
\begin{tabular}{@{}llccr@{}}
\toprule
Source & Arm & S/Y & E/B/Q/V & Calls \\
\midrule
0 & GV-v3 & 0/0 & N/N/N/N & 2 \\
0 & JC-v3 & 0/0 & N/N/N/N & 1 \\
1 & JC-v3 & N/N & N/N/N/N & 1 \\
1 & GV-v3 & N/N & N/N/N/N & 0 \\
\bottomrule
\end{tabular}
\end{table}

The source-0 rows retain their recorded structure failures from a missing or
out-of-root input, not a judgment that the underlying tasks are impossible.
Source-1 JC retains its provider-fault/NULL record; source-1 GV was not admitted.
No between-arm task-score subtraction is established.

A subsequent downstream-only technical continuation left construction and these
four rows unchanged. R796 stopped before a new admission because a required
resource was missing. After a resource-layout repair, R797 attempted source-0 GV
execution but ended with \texttt{TERMINAL\_NO\_RETRY}; downstream JC was not
admitted. The continuation's $E,B,Q,V,Y$ remain NULL. Its evidence-support
assessment was already exposed and remained unobserved; it cannot supply an
independent blind $B$ or the original protocol's paired success.

The ledger retains four earlier admissions with 131,072 reserved output tokens
and six additional admissions with 98,304 reserved output tokens: ten admissions
and 229,376 reserved tokens in total. Five new responses completed; admission 10
failed. A preliminary saved-response diagnosis (R798) identifies
\nolinkurl{too_many_requests} and \nolinkurl{rate_limit_exceeded}, a provider token-rate
rejection that the adapter surfaced as
\nolinkurl{STREAM_COMPLETION_OR_SCHEMA_INVALID}.
No post-fault admission is recorded. These are client-side admissions and output
reservations, not measured provider executions, actual token usage, monetary cost,
or independent task observations; upstream executions, retries, and charges are
unknown. The interruption supplies no package-quality verdict, accepted score,
or comparative construction effect.

\subsection{Multi-city travel: task environment and version boundary}
\paragraph{Saved artifacts and intended outputs.}
The multi-city travel task begins with booking materials, weather references, restaurant
pointers, and incomplete user preferences. Its saved specification contains task
instructions, a weighted rubric, evidence bindings, a reference solution, and a private
user-simulator specification. The intended deliverables are an itinerary CSV and a
Markdown report. The three-row example in Section~\ref{sec:saved-contract} maps the saved evidence gap
to output obligations and criteria; it does not establish completed execution.

\paragraph{Historical version mismatch.}
Selected solver--judge records are bound to the task-artifact hashes of their historical
version. For one time-window value, the saved authoring specification records
\num{case.saved_window_min} minutes, while the retained emitted rubric records
\num{case.emitted_window_min}; its hash differs from that historical binding.
The example describes the saved specification, and we do not attach historical
scores to the changed rubric. The discrepancy is a versioned construction defect, not an accepted numeric target
for this manuscript.

\paragraph{Relation to the reference-correction study.}
The recorded four-condition intervention in Appendix~\ref{app:reference-repair}
uses a separate frozen task and supplies the reference-correction evidence; historical
scores from this example are not used to measure that effect.

\subsection{Six-day travel: archived requirements and evidence limits}
\paragraph{Task requirements and translation.}
A distinct second travel-planning task, separate from the opening multi-city case,
makes the package representation concrete.
Its archived instruction and rubric specify a task in which the agent must
reconcile fixed bookings, family-chat updates, unclear handwritten notes,
and official attraction information when preparing a six-day itinerary and budget.
Table~\ref{tab:historical-example} summarizes selected requirements and one criterion
from that payload. The English text is a condensed translation; output names are
described by role rather than substituted as literal English filenames.

\begin{aabtable}[!htbp]
\centering
\small
\caption{\textbf{Archived requirements and criterion links in the six-day travel task.}
Requirements and rubric entries illustrate how task artifacts depend on shared evidence.}
\label{tab:historical-example}
\begin{tabular}{@{}p{0.23\textwidth}p{0.72\textwidth}@{}}
\toprule
Component & Archived content, condensed and translated \\
\midrule
Task requirement & Respect fixed bookings and the latest family constraints. Do not infer budgets or dietary restrictions from illegible notes. \\
Source materials & Three travel-chat screenshots, two handwritten notes, and itinerary and booking information. The payload also lists four attraction reference pages. \\
Required outputs & A six-day itinerary document and a budget spreadsheet in \texttt{output/}; the spreadsheet has seven specified columns and a verifiable-total row. \\
Rubric structure & Thirteen weighted criteria, with total weight 118. \\
Criterion A10 & Extract reliable information from images and decline to use low-confidence handwriting; weight 7. \\
A10 checks & Recover chat timing, speakers, and decisions; identify the notes' lack of reliable fields and units; avoid inventing a budget, dietary restriction, or schedule. \\
Linked criteria & B4 checks whether a venue's price is explicitly verifiable; B1 uses that basis for the total; G5 checks consistency across outputs; D7 requires evidence attribution. \\
\bottomrule
\end{tabular}
\end{aabtable}

\paragraph{Interpreting the evidence requirements.}
This example illustrates why instruction, evidence, and rubric must be checked together.
The intended response to an unreadable note is to acknowledge the uncertainty; a
confidently invented number violates the requirement even if the output files are
well formed. Likewise, the budget total and narrative itinerary must use the same
price-verification basis. The task's pricing rule is a specified conservative
placeholder, not a claim that the resulting amount is the appropriate ticket purchase.

\paragraph{Runtime mapping and assessment limits.}
This example is not linked to the recovered repeat-score groups and does not establish
independent usability or comparative system performance. The archived instruction requests official-page lookup, while the payload separately
lists reference pages. The runtime mapping between those requests and available
materials must be checked before reuse. The surviving excerpt alone does not establish
that the environment implements every requirement, that the reference facts are
correct, or that the item passed independent review. It serves as an artifact example; the excerpt has no joined solver-score panel or
verified repair trajectory.
\FloatBarrier

\section{Fixed-Output Controls for a Subsequent Scorer Version}
\label{app:judge-controls}
\subsection{Fixed-output scoring: protocol, completion, and evidence}
\paragraph{Diagnostic population and frozen protocol.}
This separate diagnostic evaluates a subsequent scorer version, without an established
configuration link to the historical endorsement audit, using
\num{prospective.judge_controls_v1_20260912.planned_outputs} fixed output variants
from \num{prospective.judge_controls_v1_20260912.synthetic_tasks} synthetic tasks,
covering \num{prospective.judge_controls_v1_20260912.planned_caps} capability instances
with \num{prospective.judge_controls_v1_20260912.votes_per_cap} ballots each.
Inputs, rubrics, and expected contrasts were frozen before scoring; expected outcomes
were withheld from the judge. Each variant received one scoring run under the fixed
native stack and its existing visual-model fallback policy. Evidence generation
and judge consumption were checked separately. The protocol stopped on missing
ballots or evidence, source drift, or unexpected outcomes, without replacement.
The \num{prospective.judge_controls_v1_20260912.planned_ballots} planned ballots
share tasks, evidence, and judge configuration; they are not independent samples
or independent task observations.

\paragraph{Prepared output contrasts.}
The masked PDF preserves the normal report's extractable text while visually
covering table data cells. Removing section markers instead preserves the body
and tables. The HTML controls contrast an effective click handler with a no-op.
Table~\ref{tab:judge-controls} reports the original verdicts and incomplete ballots.
The accompanying analysis archive supplies the recovered original controls, exact criteria,
prepared screenshots, and redacted native observations with original and distributed
hashes. Its offline verifier rechecks the artifact contrasts and all planned ballot
slots; it does not rerun native judge inference.

\begin{aabtable}[htbp]
\centering\small
\caption{\textbf{Fixed-output scorer controls, earlier recording.} Original native verdicts with
pass/fail/unknown (P/F/U) ballots. Missing ballots remain unknown; repeated ballots
are not independent samples.}
\label{tab:judge-controls}
\begin{tabular}{@{}p{0.24\textwidth}p{0.30\textwidth}p{0.38\textwidth}@{}}
\toprule
Fixed output & Native verdict; P/F/U & Evidence and limitation \\
\midrule
Normal PDF &
J7 pass; \num{prospective.judge_controls_v1_20260912.case_001.J7.pass}/\num{prospective.judge_controls_v1_20260912.case_001.J7.fail}/\num{prospective.judge_controls_v1_20260912.case_001.J7.unknown}
\newline J8 pass; \num{prospective.judge_controls_v1_20260912.case_001.J8.pass}/\num{prospective.judge_controls_v1_20260912.case_001.J8.fail}/\num{prospective.judge_controls_v1_20260912.case_001.J8.unknown} &
Rendered pages; actual image consumption \\
\addlinespace
Working HTML &
J10 pass; \num{prospective.judge_controls_v1_20260912.case_002.J10.pass}/\num{prospective.judge_controls_v1_20260912.case_002.J10.fail}/\num{prospective.judge_controls_v1_20260912.case_002.J10.unknown} &
Prepared click evidence, changed content; on-demand behavior unvalidated \\
\addlinespace
Masked table cells &
J7 fail; \num{prospective.judge_controls_v1_20260912.case_003.J7.pass}/\num{prospective.judge_controls_v1_20260912.case_003.J7.fail}/\num{prospective.judge_controls_v1_20260912.case_003.J7.unknown}
\newline J8 pass; \num{prospective.judge_controls_v1_20260912.case_003.J8.pass}/\num{prospective.judge_controls_v1_20260912.case_003.J8.fail}/\num{prospective.judge_controls_v1_20260912.case_003.J8.unknown} &
Text extractable; cells visually covered \\
\addlinespace
Removed section markers &
J7 pass; \num{prospective.judge_controls_v1_20260912.case_004.J7.pass}/\num{prospective.judge_controls_v1_20260912.case_004.J7.fail}/\num{prospective.judge_controls_v1_20260912.case_004.J7.unknown}
\newline J8 fail; \num{prospective.judge_controls_v1_20260912.case_004.J8.pass}/\num{prospective.judge_controls_v1_20260912.case_004.J8.fail}/\num{prospective.judge_controls_v1_20260912.case_004.J8.unknown} &
Headings/dividers/frames absent; body retained \\
\addlinespace
No-operation HTML &
J10 fail; \num{prospective.judge_controls_v1_20260912.case_005.J10.pass}/\num{prospective.judge_controls_v1_20260912.case_005.J10.fail}/\num{prospective.judge_controls_v1_20260912.case_005.J10.unknown} &
Prepared click evidence shows unchanged content; one final ballot missing (allocation incomplete) \\
\bottomrule
\end{tabular}
\end{aabtable}

\FloatBarrier

\paragraph{Completed and missing ballots.}
All \num{prospective.judge_controls_v1_20260912.attempted_outputs} variants were
attempted; \num{prospective.judge_controls_v1_20260912.complete_outputs} completed
their ballot allocation and \num{prospective.judge_controls_v1_20260912.partial_outputs}
remained partial. Of \num{prospective.judge_controls_v1_20260912.planned_ballots}
planned ballots, \num{prospective.judge_controls_v1_20260912.valid_ballots} were valid:
\num{prospective.judge_controls_v1_20260912.pass_ballots} pass and
\num{prospective.judge_controls_v1_20260912.fail_ballots} fail;
\num{prospective.judge_controls_v1_20260912.unknown_ballots} remained unknown.
The no-operation HTML has no valid final ballot. Its native aggregate is fail and
its summary marks completion, but the required ballot allocation is incomplete.
The missing ballot is neither a pass nor a fail and is not imputed.

\paragraph{Interaction evidence coverage.}
The earlier HTML judgments used prepared interaction evidence; they do not establish
on-demand browsing. The later recording below evaluates actual click behavior.

\paragraph{Reported model identities, cost coverage, and interpretation.}
The agent and visual tools requested \texttt{gpt-5.6-sol}; SDK responses disclosed
that name, while visual responses disclosed
\path{deployment-gpt-5.6-sol-platform-global}. These labels do not establish
hidden backend equivalence. Complete visual-tool and invocation charges were
unavailable, so full-stack cost is unknown. These task-dependent controls establish
neither general scoring accuracy nor repair or transfer benefit, and do not
authenticate the historical audit's scorer configuration.

\subsection{A later recording on the same fixed outputs}
\label{app:judge-controls-r310}
\paragraph{Protocol, coverage, and the unobserved criterion.}
A later recording reran the same
\num{prospective.judge_controls_r310_20260928.cases_total} frozen synthetic outputs as
\num{prospective.judge_controls_r310_20260928.planned_ballots} single-criterion
conversations, one criterion per conversation, over
\num{prospective.judge_controls_r310_20260928.criteria_total} criterion instances with
\num{prospective.judge_controls_r310_20260928.browse_votes_per_html} ballots each.
Every role, including the visual role, requested one routing alias, and each ballot was
capped at \num{prospective.judge_controls_r310_20260928.per_ballot_http_max} HTTP
requests, \num{prospective.judge_controls_r310_20260928.per_ballot_tool_max} tool calls,
and \num{prospective.judge_controls_r310_20260928.per_ballot_seconds} seconds including
cleanup. Of the planned ballots,
\num{prospective.judge_controls_r310_20260928.valid_ballots} are valid
(\num{prospective.judge_controls_r310_20260928.pass_ballots} pass,
\num{prospective.judge_controls_r310_20260928.fail_ballots} fail) and
\num{prospective.judge_controls_r310_20260928.unknown_ballots} remain unknown, so
\num{prospective.judge_controls_r310_20260928.criteria_complete} of
\num{prospective.judge_controls_r310_20260928.criteria_total} criteria completed their
allocation. Those \num{prospective.judge_controls_r310_20260928.unknown_ballots}
belong to one layout criterion with no completed ballot allocation. That criterion is excluded rather than failed, and its variant stays
incomplete even though its graded native score is maximal, so
\num{prospective.judge_controls_r310_20260928.cases_complete} of
\num{prospective.judge_controls_r310_20260928.cases_total} variants are complete.
Unknown ballots remain in the planned denominator and are not imputed.

\begin{aabtable}[htbp]
\centering\small
\caption{\textbf{Later fixed-output scorer-control recording.} Same five frozen
synthetic artifacts as Table~\ref{tab:judge-controls}, separately recorded. Native
per-criterion verdict with pass/fail/unknown (P/F/U) ballots out of
\num{prospective.judge_controls_r310_20260928.browse_votes_per_html} per criterion.
Unknown ballots stay unknown: they are never read as failures or successes. Repeated
ballots share artifacts and one configuration and are not independent samples.}
\label{tab:judge-controls-r310}
\begin{tabular}{@{}p{0.22\textwidth}p{0.31\textwidth}p{0.39\textwidth}@{}}
\toprule
Fixed output & Native verdict; P/F/U & In-browser and visual evidence \\
\midrule
Normal PDF &
J7 pass; \num{prospective.judge_controls_r310_20260928.case_001.J7.pass}/\num{prospective.judge_controls_r310_20260928.case_001.J7.fail}/\num{prospective.judge_controls_r310_20260928.case_001.J7.unknown}
\newline J8 pass; \num{prospective.judge_controls_r310_20260928.case_001.J8.pass}/\num{prospective.judge_controls_r310_20260928.case_001.J8.fail}/\num{prospective.judge_controls_r310_20260928.case_001.J8.unknown} &
Rendered pages; screenshot digest equals the decoded model input image \\
\addlinespace
Working HTML &
J10 pass; \num{prospective.judge_controls_r310_20260928.case_002.J10.pass}/\num{prospective.judge_controls_r310_20260928.case_002.J10.fail}/\num{prospective.judge_controls_r310_20260928.case_002.J10.unknown} &
Actual click walkthrough in all
\num{prospective.judge_controls_r310_20260928.browse_votes_per_html} ballots; content
changed in \num{prospective.judge_controls_r310_20260928.functional_dom_changed} \\
\addlinespace
Masked table cells &
J7 fail; \num{prospective.judge_controls_r310_20260928.case_003.J7.pass}/\num{prospective.judge_controls_r310_20260928.case_003.J7.fail}/\num{prospective.judge_controls_r310_20260928.case_003.J7.unknown}
\newline J8 pass; \num{prospective.judge_controls_r310_20260928.case_003.J8.pass}/\num{prospective.judge_controls_r310_20260928.case_003.J8.fail}/\num{prospective.judge_controls_r310_20260928.case_003.J8.unknown} &
Text still extractable; data cells visually covered \\
\addlinespace
Removed section markers &
J7 pass; \num{prospective.judge_controls_r310_20260928.case_004.J7.pass}/\num{prospective.judge_controls_r310_20260928.case_004.J7.fail}/\num{prospective.judge_controls_r310_20260928.case_004.J7.unknown}
\newline J8 excluded; \num{prospective.judge_controls_r310_20260928.case_004.J8.pass}/\num{prospective.judge_controls_r310_20260928.case_004.J8.fail}/\num{prospective.judge_controls_r310_20260928.case_004.J8.unknown} &
Layout criterion never observed; variant incomplete despite its graded native score \\
\addlinespace
No-operation HTML &
J10 fail; \num{prospective.judge_controls_r310_20260928.case_005.J10.pass}/\num{prospective.judge_controls_r310_20260928.case_005.J10.fail}/\num{prospective.judge_controls_r310_20260928.case_005.J10.unknown} &
Actual click walkthrough in all
\num{prospective.judge_controls_r310_20260928.browse_votes_per_html} ballots; content
changed in \num{prospective.judge_controls_r310_20260928.noop_dom_changed} \\
\bottomrule
\end{tabular}
\end{aabtable}

\FloatBarrier

\paragraph{Actual browser execution and visual evidence.}
Unlike the earlier recording, on-demand browsing executed: all
\num{prospective.judge_controls_r310_20260928.browse_walkthroughs} click walkthroughs,
one per HTML ballot, were actually performed with no page errors. The functional control
changed the page content in
\num{prospective.judge_controls_r310_20260928.functional_dom_changed} of
\num{prospective.judge_controls_r310_20260928.browse_votes_per_html} ballots and the
no-operation control in
\num{prospective.judge_controls_r310_20260928.noop_dom_changed}, matching the recorded
verdicts. For the visual criteria,
\num{prospective.judge_controls_r310_20260928.image_records} image records bind each
source artifact to its screenshot and to the decoded model input image.

\paragraph{Requested configuration, budgets, and interpretation.}
The recording issued \num{prospective.judge_controls_r310_20260928.http_requests} HTTP
requests (\num{prospective.judge_controls_r310_20260928.http_text} text,
\num{prospective.judge_controls_r310_20260928.http_vision} visual) and
\num{prospective.judge_controls_r310_20260928.known_total_tokens} known total tokens;
\num{prospective.judge_controls_r310_20260928.usage_unknown_requests} request never
reported usage, so full-stack cost is unknown and is not imputed. The requested alias is
a harness routing label, not an authenticated backend identity or a vendor model
citation; Appendix~\ref{app:model-configurations} lists the requested configuration.
Because the requested model, the batching, and the runtime all differ from the earlier
recording, no single change is isolated as the cause of the observed browsing difference, and this
recording repairs no earlier number: both remain reported separately. Like the earlier
one, it establishes criterion-specific responses on frozen synthetic artifacts and
supplies no human agreement, independent deliverable usability, general scoring accuracy,
or construction benefit. The public bundle republishes these per-slot records with an
offline recomputation of every count above; its digests bind the distributed bytes only
and, because the original model inputs are withheld, support reaggregation rather than a
native replay.

\subsection{Selected arithmetic correction: a paired reference-target intervention}
\label{app:reference-repair}
\paragraph{Case selection and the corrected target.}
This exploratory follow-up uses \num{repair_case.n_packages} package selected after
an earlier arithmetic failure was observed. The earlier verdicts only motivated
selection; every cell in Table~\ref{tab:reference-repair} records a native
judgment collected for that intervention. Integer recomputation of the explicit visit durations gives
$120+3\times60+4\times20=\num{repair_case.corrected_minutes}$ minutes.
The intervention changes the rubric's expected total and the equivalent private
reference duration from \num{repair_case.original_minutes} to
\num{repair_case.corrected_minutes} minutes. All other package files, task instructions,
evidence, and both answer artifacts remain byte-identical. This is a deterministic
reference correction, not an independently assessed autonomous constructor repair.

\paragraph{Frozen conditions and judge configuration.}
The protocol and input hashes were frozen before scoring. Conditions cross original
versus corrected targets with fixed incorrect versus correct outputs, in a prespecified
shuffled order. Each receives \num{repair_case.ballots_per_condition} native batch ballot,
using the same Agent-as-Judge implementation, \texttt{single\_judge\_v1} profile,
and requested \texttt{gpt-5.6-sol} model; caches, retries, and auxiliary visual-model
calls are disabled in every condition. The judge receives the normal package and
answer, without experimental labels or expected verdicts. The endpoint is criterion B1,
not the correctness of every output claim or the overall weighted score. Other criteria
can vary across fresh judgments.

\paragraph{Completed comparison.}
All
\num{repair_case.n_conditions} arithmetic cells completed and the input hashes remained
unchanged. The accompanying analysis archive contains the criterion outcomes, source and protocol digests, and an
offline arithmetic verification script.

\paragraph{Within-case interpretation and replication.}
Let $p_{r,o}$ denote the pass probability under a repeated-judgment model for reference
$r$ and fixed output $o$. The observed contrast
$(\widehat p_{\mathrm{corrected},380}-\widehat p_{\mathrm{original},380})
-(\widehat p_{\mathrm{corrected},320}-\widehat p_{\mathrm{original},320})=2$
is computable, but one ballot per cell supplies no within-cell replication for
estimating repeatability or sampling variance. Even under a Bernoulli model for
repetition of a fixed cell, a single pass has the exact two-sided 95\% interval
$[0.025,1]$; a single fail has $[0,0.975]$. These model-based bounds are not
additional observations or population-accuracy evidence. Exact hidden backend
identity is not established by the requested model label.

A replication must keep both output artifacts, both reference variants, the judge
configuration, and the B1 endpoint fixed. Independent blocks should contain all
four conditions in randomized order with reset contexts; invalid returns remain
separate from pass and fail. A precision target, fixed sample size or valid
sequential rule, and feasible operating budget must precede new observations.

\subsection{Deterministic retrieval and disclosure checks}
\label{app:controller-checks}
\paragraph{Frozen fixtures and production entry points.}
A behavioral protocol with synthetic evidence and user facts was frozen before invoking
the original production functions. The main and Daily-Life snapshots were tested through
\path{SearchServer.do_search/do_fetch} and
\path{ControlledUserSimulator.decide/commit}, without a language-model backend,
online retrieval, or monkeypatching of controller logic. Table~\ref{tab:controller-checks}
reports every planned case. Each result is an assertion about the specified fixture,
not a population accuracy estimate or a human assessment.

\begin{aabtable}[htbp]
\centering\small
\caption{\textbf{Direct checks of production information-access functions.}
The same frozen fixtures exercise both implementation snapshots.}
\label{tab:controller-checks}
\begin{tabular}{@{}p{0.65\textwidth}cc@{}}
\toprule
Predeclared behavior & Main & Daily-Life \\
\midrule
Exact query returns captured evidence & Pass & Pass \\
Unrelated query returns no fabricated result & Pass & Pass \\
Fetch returns the captured plain-text content & Pass & Pass \\
Opening decision exposes the designated public fact & Pass & Pass \\
Matching clarification releases the authorized fact & Pass & Pass \\
Unrelated clarification releases no additional fact & Pass & Pass \\
Never-disclose fact remains outside disclosed state & Pass & Pass \\
Committed fact is not issued again on the repeated query & Pass & Pass \\
\bottomrule
\end{tabular}
\end{aabtable}

\paragraph{Source provenance and replay.}
The supplement contains the fixed fixtures, assertions, original observations, and
source-derived offline replay. Distributed source copies preserve executable
statements; original and distributed source digests are retained. These checks
establish authorized release sets and controller tags on the specified short
sequences. They do not validate the semantic compliance of generated replies or
end-to-end task behavior.

\paragraph{Coverage limits.}
The unrelated-query test uses a short sequence: later stall-recovery policies can
volunteer facts. Fetch equality is tested on plain text, not arbitrary HTML bytes.
Paraphrase robustness, semantic fallback, rendered language-model replies, long
conversations, and task-level usability are outside this diagnostic.
Appendix~\ref{app:disclosure-control} documents the supplied-source negated-mention
and stall-recovery counterexamples separately from these original fixtures.

\section{Retained historical measurements}
\label{app:detailed-evaluation}
The opening appendix index (Table~\ref{tab:evidence-index}) gives canonical results
and protocols; the records below retain historical version and coverage boundaries.
\subsection{Traceability of recovered historical executions}
\label{app:historical-traceability}
Table~\ref{tab:historical-execution} brings together two previously documented
runs and nine further recovered runs. All 11 have recorded solver completion;
11 is a recovery/linkage count within the existing 20-task sample, not a success
or validity rate, and selection is not representative or unbiased. The two earlier
cases retain native rounded scores 0.802 (P039) and 0.763 (P068). P039 used 20 model
rounds, 19 tool rounds, and 35 function calls; P068 used 24, 23, and 48.

\begin{aabtable}[!htbp]
\centering\small
\caption{\textbf{Eleven recovered historical runs: native conditional accounting.}
$N/G/F$ denotes passed weight / graded weight / full rubric weight, not a
platform-adjusted score. NULL lists every excluded criterion. I/R means exact
instruction/rubric bytes; M adds exact primary materials. S denotes search-backed
cases without primary material files; full search-corpus identity is unverified.
Files counts the nine newly recovered runs' top-level deliverables; a dash for
the earlier pair means no new file count is asserted.}
\label{tab:historical-execution}
\begin{tabular}{@{}llrll@{}}
\toprule
Case & $N/G/F$ & Files & NULL & Frozen correspondence \\
\midrule
P006 & 81/108/121 & 1 & J3, J11 & I/R; S \\
P017 & 30/104/104 & 2 & none & I/R/M \\
P033 & 37/101/111 & 3 & J7 & I/R; S \\
P037 & 17/117/134 & 2 & J3, J9 & I/R; S \\
P039 & 81/101/108 & --- & J5 & I/M; rubric differs \\
P047 & 20/101/135 & 1 & J9, J11, J6, J2 & I/R; S \\
P050 & 64/108/108 & 1 & none & I/R; S \\
P054 & 98/108/115 & 3 & J9 & I/R/M \\
P056 & 134/134/141 & 2 & J9 & I/R/M \\
P068 & 87/114/124 & --- & J2 & I/R/M \\
P099 & 125/125/125 & 1 & none & I/R; S \\
\bottomrule
\end{tabular}
\end{aabtable}

The nine additional runs are supported by 51 checked archive objects: 18
instruction/rubric files, 17 primary materials, and 16 final deliverables, including
an OLE spreadsheet for P099. Exact instruction and rubric correspondence holds
for all nine; the material matches apply to P017, P054, and P056. The other six
have recorded search/fetch activity, but no primary material files in their
workspaces. This is historical execution evidence, not independent validation of
frozen final packages. The source job's STOPPED state is distinct from the later
report's DONE state and from these individual completed executions.

For all nine, each graded criterion has three observed valid votes, whereas
archived configuration specifies one vote per criterion. These are stated in the
same units; configuration does not replace observed voting. The 27 sampled judge
requests (three per case) name \texttt{gpt-5.5-2026-04-24} in request bodies and
request arguments. The configured name, generic serving alias, and result-path
label are separate metadata; request identity does not attest backend weights.
The historical solver's maximum reasoning effort remains separate from the
supplementary solver's high-effort configuration.

All nine native summaries mark judging complete, although P006, P033, P037,
P047, P054, and P056 exclude process criteria for unavailable trace evidence.
Their NULL criteria remain outside the conditional denominator. P039 excludes
weight 7 and P068 weight 10 for missing trace evidence; their pass/fail/excluded
counts remain 10/2/1 and 10/3/1. A native complete flag does not imply full rubric
coverage or independent accuracy. Four of the nine rubrics repeat criterion IDs;
occurrence suffixes \#1/\#2 retain their exact statement and weight bindings.
No full-weight score is imputed from the conditional ratios.

Historical runtime scripts, configurations, and execution images differ from the
frozen versions. Exact oracle-manifest correspondence does not establish identity
of the referenced held-out solution bytes; those full bytes and the full deployed
search corpus remain unverified. P039's rubric difference and P068's runtime
boundary also remain explicit. Independent human labels for these exact runs have
not been established; the existing nonblind endorsements remain separate evidence.
All supplementary failures, NULL outcomes, and consumed-once accounting are
unchanged. In particular, historical P099 completion does not reverse its
supplementary frozen failure, and neither historical P039 nor P068 promotes a
frozen accepted score.

\subsection{Offline verification of an archived execution}
\label{app:archive-verification}
P017 was selected for completeness of its archived evidence, not its score;
this selection does not define an unbiased case sample. Exact archived records
connect the original instruction and model request, 45 tool calls and 45 tool
results, the final response, and two delivered Office files. Both files parse and
match archived bytes. Tool pathname references and the final handoff establish
artifact linkage, without claiming to reconstruct every filesystem write.
All 12 criterion records are bound by statement, weight, and occurrence where
IDs repeat. Their 36 observed valid votes reconstruct passed weight 30 and failed
weight 74, giving $30/104=0.288$ after native rounding, with zero excluded weight
and no missing criterion verdicts. The configured one vote per criterion remains
distinct from the three observed votes per criterion.

\paragraph{What the retained ballots can test.}
An offline comparison applies strict majority to these 36 saved votes, counting
the 12 occurrence-resolved criteria once, rather than also counting their mirrored
summary entries. Every verdict agrees with the saved result, retaining passed
weight $30/104$. The pass:fail patterns are five $0{:}3$, four $3{:}0$, and three
$1{:}2$. None has a pass majority with a failing dissent, the only pattern on which
Equation~\ref{eq:evidence-sensitive-vote} can override majority. Its outcomes are
therefore invariant to the unresolved deep/anchor flags on these records.
The projection records the historical \texttt{single\_judge\_v0} method, not proof
that the newer rule ran; ballot depth and anchor eligibility are not retained.
This case supplies no veto-discriminating observation and no independent accuracy
comparison. Agreement here cannot validate the override on other vote patterns.

This verification concerns one historical exposure. The supplementary P017 exposure
retains native $44/104$ and evidence-revalidated $37/97$, including excluded weight
7; neither is replaced by the historical score. The historical held-out solution
bytes were not recovered, complete runtime equivalence is unestablished, and
independent correctness labels are unavailable. Only three archived judge request
samples accompany the 36 vote records; these do not establish all request contexts.

A deterministic source check on five locked summer rows gives mean oil 21\,g
and salt 4.6\,g. These are conditional arithmetic results: season applicability
remains unresolved. The instruction requests the current season without a planning
date; the archived request environment gives September 5, 2026, and the output uses
September. The earlier reference-label mismatch remains a prior observation, not
a newly verified held-out workbook comparison.

\paragraph{Reproducibility entrypoint and projection boundary.}
The supplement at \path{anc/analysis/supplement/historical_p017} retains seven sanitized
projections: instruction, dinner-candidate data, rubric, all criterion records,
final response, and two Office content-XML projections. From \path{anc/analysis/}, run:
\begin{quote}\small
\begin{verbatim}
python -I supplement/historical_p017/verify.py
\end{verbatim}
\end{quote}
This offline command verifies retained-file integrity and native arithmetic only.
It does not authenticate the withheld exact archive, reproduce original Office
layout or metadata, rerun the agent, or independently score correctness.
The source content of \path{notes.md}, containing consultation-draft quotations
marked against redistribution and adoption, is omitted; a hash-bound omission
record replaces it. The original remains immutable in the private exact archive.
Prior private verification covered all eight projections with 21 substitutions;
that evidence is preserved, while the seven retained projections are unchanged.
The public subset is not exact source replay or complete end-to-end reproduction.
Historical held-out bytes remain unavailable.

\FloatBarrier
\section{Historical Statistical Panels}
\label{sec:historical-statistics}
The capability data carry dataset version tag 20260710; the exact collection date is unavailable. They are separate from the October 2026 platform-reported leaderboard and supply no confidence intervals or task linkage for it. These analyses concern different historical populations and are not linked as
a single causal sequence. The opening appendix index gives canonical result locations;
Table~\ref{tab:cohort-inventory} retains additional historical units and linkage. The archive's capability scores, repeated runs,
construction outcomes, costs, and reviews were collected for different purposes and do
not form a single controlled experiment.

\begin{aabtable}[!htbp]
\centering\small
\caption{\textbf{Additional historical units and linkage.} These records describe the same
Daily-Life application, but their denominators and version evidence differ.}
\label{tab:cohort-inventory}
\begin{tabular}{@{}p{0.27\textwidth}p{0.22\textwidth}p{0.45\textwidth}@{}}
\toprule
Cohort & Unit and size & Selection and available linkage \\
\midrule
Linked adjacent revisions & \num{adj.N01} pairs, \num{adj.N02} tasks & Selected from \num{adj.N03} measured pairs; recorded version/run linkage, changing rubrics. \\
Capability panel & \num{cap.n_items} items, \num{cap.n_models} solvers & Platform-reported adjusted aggregates; no established release-version join. \\
Signature-matched repeats & \num{q0.sm_n_groups} item--solver--round groups & Adaptive repeats; capability IDs and weights match, full versions are unverified. \\
Recovered repeat panel & \num{q1.n_groups} score lists & Recovered from a committed vector figure; item and tier identities unavailable. \\
Gate outcomes & \num{po.n_items} records & Retention outcomes; score-to-final-version consistency is unverified. \\
\bottomrule
\end{tabular}
\end{aabtable}

\paragraph{Configuration and denominator crosswalk.}
Table~\ref{tab:configuration-crosswalk} complements the cohort inventory with the
configuration and denominator needed to interpret each endpoint. ``Unavailable'' means
that the archive does not establish the setting; later requested model labels do not
fill historical gaps. Across the construction cohorts, complete authoring model,
prompt, simulator-backend, and budget configurations are unavailable. Source hashes
identify withheld objects but do not supply their contents.

\begingroup\small
\captionsetup{singlelinecheck=false}
\setlength{\LTcapwidth}{\linewidth}
\begin{longtable}{@{}p{0.18\linewidth}p{0.35\linewidth}p{0.40\linewidth}@{}}
\caption{\textbf{Study configurations, denominators, and public evidence.}
These joins are study-specific; no common final-release version join is established.}
\label{tab:configuration-crosswalk}\\
\toprule
Study & Configuration / version key & Denominator and reproduction boundary \\
\midrule
\endfirsthead
\multicolumn{3}{@{}l}{\small Table \thetable\ (continued). Study configurations, denominators, and public evidence.}\\
\toprule
Study & Configuration / version key & Denominator and reproduction boundary \\
\midrule
\endhead
\midrule
\multicolumn{3}{r@{}}{\small Continued on next page.}\\
\endfoot
\bottomrule
\endlastfoot
\raggedright Release / audit & \raggedright Frozen inventory hashes; fixed domain/mode sample. Authoring settings unavailable. & \raggedright Packages per named check; failures and unknowns retained. Inventory and receipts support reaggregation, not environment execution. \tabularnewline
\raggedright Adjacent revisions & \raggedright Saved surface hashes and logged aliases in \texttt{TABLE-DATA.csv}; thresholds, profile flag, and full budgets unavailable. & \raggedright Stored tier scores versus selected-run scores; saved and graded weights separate. Projections and AI-assisted case coding are public; original snapshots are withheld. \tabularnewline
\raggedright Capability panel & \raggedright Fixed solver IDs and recorded names in Table~\ref{tab:model-configurations}; serving checkpoints unavailable. & \raggedright Platform-reported adjusted 100-item aggregates in Table~\ref{tab:model-configurations}; adjusted per-item observations unavailable. \tabularnewline
\raggedright Adaptive repeats & \raggedright Item/solver/round plus capability/weight multiset; full-version identity unavailable. & \raggedright Observed scores within selected groups; the signature retains exclusions and does not fix the graded denominator. Raw repeat archive unavailable. \tabularnewline
\raggedright Recovered repeats & \raggedright Original item, tier, and configuration identities unavailable. & \raggedright Score lists recovered from a vector figure; resampling reuses those observations. \tabularnewline
\raggedright Gate / costs & \raggedright Recorded retention outcomes; score-to-final-version and item-to-cost joins unavailable. & \raggedright Outcome records, observed round counts, or positively priced runs; never cost per usable task. Raw archives unavailable. \tabularnewline
\raggedright Human review & \raggedright Separate dated snapshots; historical judge model/prompt/version mapping unavailable. & \raggedright Endorsed/reviewed criteria, including exclusions and missing verdict fields; pass/fail subset reported separately. Per-item counts reproduce pooled September results, not reviewer/QA subgroups. \tabularnewline
\raggedright Scorer controls, earlier recording (Table~\ref{tab:judge-controls}) & \raggedright Requested \texttt{gpt-5.6-sol}; fixed native stack with visual fallback. & \raggedright Valid versus planned ballots; HTML uses prepared evidence. Original synthetic PDF/HTML controls, exact criteria, prepared images, and redacted native observations are public; offline inspection does not rerun native judging. \tabularnewline
\raggedright Scorer controls, later recording (Table~\ref{tab:judge-controls-r310}; Appendix~\ref{app:judge-controls-r310}) & \raggedright Separately recorded run on the same five frozen synthetic artifacts. Requested routing alias \texttt{gpt-6-astra} for every role, including the visual role; one single-criterion conversation per ballot, capped at \num{prospective.judge_controls_r310_20260928.per_ballot_http_max} HTTP requests, \num{prospective.judge_controls_r310_20260928.per_ballot_tool_max} tool calls, and \num{prospective.judge_controls_r310_20260928.per_ballot_seconds} seconds including cleanup. Requested model, batching, and runtime all differ from the earlier recording, so the two are distinct configurations and neither replaces the other. & \raggedright \num{prospective.judge_controls_r310_20260928.valid_ballots} valid of \num{prospective.judge_controls_r310_20260928.planned_ballots} planned ballots, \num{prospective.judge_controls_r310_20260928.unknown_ballots} unknown and never read as pass or fail; \num{prospective.judge_controls_r310_20260928.criteria_complete} of \num{prospective.judge_controls_r310_20260928.criteria_total} criteria complete. Per-slot records, sanitized amendment, and digests are public in \texttt{scorer\_controls\_r310}; digests bind distributed bytes only and support reaggregation, not native replay. Synthetic diagnostics only: no human agreement and no scoring-accuracy estimate. \tabularnewline
\raggedright Reference repair & \raggedright \texttt{single\_judge\_v1}, requested \texttt{gpt-5.6-sol}; fixed outputs, linked target edit. & \raggedright Arithmetic criterion per condition. Public checker verifies the sum and stored verdicts, not native judging. \tabularnewline
\raggedright Controller checks & \raggedright Two pinned source snapshots; \texttt{backend=None}. & \raggedright Fixture assertions per implementation; public source replay tests authorization and retrieval, not generated dialogue. \tabularnewline
\end{longtable}
\endgroup


\subsection{Historical adjacent-round revision: versions and mechanisms}
\label{app:adjacent-revision}
This case series connects the rewriting, controlled disclosure, and candidate-selection
mechanisms in Section~\ref{sec:method} to saved historical transitions. The linked
stratum contains \num{adj.N01} pairs from \num{adj.N02} tasks, with \num{adj.N11} unique saved versions and
\num{adj.N12} distinct selected runs, selected for recoverable version and run linkage from
\num{adj.N03} measured Daily-Life pairs. Table~\ref{tab:adjacent-revisions} lists every pair.
The release, repeat, capability, human-review, and controlled-reference studies
retain their own populations and interpretations. The saved round numbers identify task versions, not calendar dates; no exact collection date is established for this linked cohort. Current model labels or aggregate scores cannot be substituted into these transitions.

\begin{aabtable}[!htbp]
\centering\footnotesize
\setlength{\tabcolsep}{3pt}
\caption{\textbf{All 19 linked historical adjacent-round revisions (saved construction rounds).} Stored strong ($S$) and weak ($W$) tier scores; $Q$ is loop qualification (T/F); $R$ is the exact retained saved-criterion count; $\Sigma w$ is saved total rubric weight. Scores use changing rubrics, not a common scale. Task labels are descriptive. This historical version-linked cohort is separate from the October 2026 leaderboard (Table~\ref{tab:current-dailylife}); $S/W$ denote saved tiers, not current model names.}
\label{tab:adjacent-revisions}
\begin{tabular}{@{}llccccc@{}}
\toprule
Task / seed & Rounds & $S$ & $W$ & $Q$ & $\Sigma w$ & $R$ \\
\midrule
Medication$^{*}$ / \texttt{03-tFhgk6} & 1$\to$2 & \num{adj.T01.snapshot_smax_pre}$\to$\num{adj.T01.snapshot_smax_post} & \num{adj.T01.snapshot_wmax_pre}$\to$\num{adj.T01.snapshot_wmax_post} & T$\to$F & \num{adj.T01.pre_spec_weight}$\to$\num{adj.T01.post_spec_weight} & \num{adj.T01.exact_retained_body_weight_count} \\
Trekking / \texttt{04-uBQdVu} & 1$\to$2 & \num{adj.T02.snapshot_smax_pre}$\to$\num{adj.T02.snapshot_smax_post} & \num{adj.T02.snapshot_wmax_pre}$\to$\num{adj.T02.snapshot_wmax_post} & F$\to$T & \num{adj.T02.pre_spec_weight}$\to$\num{adj.T02.post_spec_weight} & \num{adj.T02.exact_retained_body_weight_count} \\
Phone selection / \texttt{06-uBWg4J} & 1$\to$2 & \num{adj.T03.snapshot_smax_pre}$\to$\num{adj.T03.snapshot_smax_post} & \num{adj.T03.snapshot_wmax_pre}$\to$\num{adj.T03.snapshot_wmax_post} & F$\to$T & \num{adj.T03.pre_spec_weight}$\to$\num{adj.T03.post_spec_weight} & \num{adj.T03.exact_retained_body_weight_count} \\
Food exchange$^{*}$ / \texttt{07-RS9gRu} & 1$\to$2 & \num{adj.T04.snapshot_smax_pre}$\to$\num{adj.T04.snapshot_smax_post} & \num{adj.T04.snapshot_wmax_pre}$\to$\num{adj.T04.snapshot_wmax_post} & F$\to$F & \num{adj.T04.pre_spec_weight}$\to$\num{adj.T04.post_spec_weight} & \num{adj.T04.exact_retained_body_weight_count} \\
Music rights / \texttt{08-uB1Fgc} & 1$\to$2 & \num{adj.T05.snapshot_smax_pre}$\to$\num{adj.T05.snapshot_smax_post} & \num{adj.T05.snapshot_wmax_pre}$\to$\num{adj.T05.snapshot_wmax_post} & F$\to$F & \num{adj.T05.pre_spec_weight}$\to$\num{adj.T05.post_spec_weight} & \num{adj.T05.exact_retained_body_weight_count} \\
Medical review / \texttt{08-uB7leA} & 1$\to$2 & \num{adj.T06.snapshot_smax_pre}$\to$\num{adj.T06.snapshot_smax_post} & \num{adj.T06.snapshot_wmax_pre}$\to$\num{adj.T06.snapshot_wmax_post} & F$\to$F & \num{adj.T06.pre_spec_weight}$\to$\num{adj.T06.post_spec_weight} & \num{adj.T06.exact_retained_body_weight_count} \\
Medical review / \texttt{08-uB7leA} & 2$\to$3 & \num{adj.T07.snapshot_smax_pre}$\to$\num{adj.T07.snapshot_smax_post} & \num{adj.T07.snapshot_wmax_pre}$\to$\num{adj.T07.snapshot_wmax_post} & F$\to$F & \num{adj.T07.pre_spec_weight}$\to$\num{adj.T07.post_spec_weight} & \num{adj.T07.exact_retained_body_weight_count} \\
Gift email$^{*}$ / \texttt{09-KAYuJA} & 1$\to$2 & \num{adj.T08.snapshot_smax_pre}$\to$\num{adj.T08.snapshot_smax_post} & \num{adj.T08.snapshot_wmax_pre}$\to$\num{adj.T08.snapshot_wmax_post} & F$\to$F & \num{adj.T08.pre_spec_weight}$\to$\num{adj.T08.post_spec_weight} & \num{adj.T08.exact_retained_body_weight_count} \\
VAT registration / \texttt{10-Lk8GTn} & 2$\to$3 & \num{adj.T09.snapshot_smax_pre}$\to$\num{adj.T09.snapshot_smax_post} & \num{adj.T09.snapshot_wmax_pre}$\to$\num{adj.T09.snapshot_wmax_post} & F$\to$F & \num{adj.T09.pre_spec_weight}$\to$\num{adj.T09.post_spec_weight} & \num{adj.T09.exact_retained_body_weight_count} \\
Visa / \texttt{10-LkJQw9} & 1$\to$2 & \num{adj.T10.snapshot_smax_pre}$\to$\num{adj.T10.snapshot_smax_post} & \num{adj.T10.snapshot_wmax_pre}$\to$\num{adj.T10.snapshot_wmax_post} & F$\to$F & \num{adj.T10.pre_spec_weight}$\to$\num{adj.T10.post_spec_weight} & \num{adj.T10.exact_retained_body_weight_count} \\
Visa / \texttt{10-LkJQw9} & 2$\to$3 & \num{adj.T11.snapshot_smax_pre}$\to$\num{adj.T11.snapshot_smax_post} & \num{adj.T11.snapshot_wmax_pre}$\to$\num{adj.T11.snapshot_wmax_post} & F$\to$F & \num{adj.T11.pre_spec_weight}$\to$\num{adj.T11.post_spec_weight} & \num{adj.T11.exact_retained_body_weight_count} \\
Resume tailoring / \texttt{11-OIZb1W} & 1$\to$2 & \num{adj.T12.snapshot_smax_pre}$\to$\num{adj.T12.snapshot_smax_post} & \num{adj.T12.snapshot_wmax_pre}$\to$\num{adj.T12.snapshot_wmax_post} & F$\to$F & \num{adj.T12.pre_spec_weight}$\to$\num{adj.T12.post_spec_weight} & \num{adj.T12.exact_retained_body_weight_count} \\
Resume tailoring / \texttt{11-OIZb1W} & 2$\to$3 & \num{adj.T13.snapshot_smax_pre}$\to$\num{adj.T13.snapshot_smax_post} & \num{adj.T13.snapshot_wmax_pre}$\to$\num{adj.T13.snapshot_wmax_post} & F$\to$F & \num{adj.T13.pre_spec_weight}$\to$\num{adj.T13.post_spec_weight} & \num{adj.T13.exact_retained_body_weight_count} \\
Parenting / \texttt{11-d9PDs9} & 1$\to$2 & \num{adj.T14.snapshot_smax_pre}$\to$\num{adj.T14.snapshot_smax_post} & \num{adj.T14.snapshot_wmax_pre}$\to$\num{adj.T14.snapshot_wmax_post} & F$\to$F & \num{adj.T14.pre_spec_weight}$\to$\num{adj.T14.post_spec_weight} & \num{adj.T14.exact_retained_body_weight_count} \\
Parenting / \texttt{11-d9PDs9} & 2$\to$3 & \num{adj.T15.snapshot_smax_pre}$\to$\num{adj.T15.snapshot_smax_post} & \num{adj.T15.snapshot_wmax_pre}$\to$\num{adj.T15.snapshot_wmax_post} & F$\to$F & \num{adj.T15.pre_spec_weight}$\to$\num{adj.T15.post_spec_weight} & \num{adj.T15.exact_retained_body_weight_count} \\
Wedding venue / \texttt{11-mRDleW} & 1$\to$2 & \num{adj.T16.snapshot_smax_pre}$\to$\num{adj.T16.snapshot_smax_post} & \num{adj.T16.snapshot_wmax_pre}$\to$\num{adj.T16.snapshot_wmax_post} & F$\to$F & \num{adj.T16.pre_spec_weight}$\to$\num{adj.T16.post_spec_weight} & \num{adj.T16.exact_retained_body_weight_count} \\
Recruitment / \texttt{11-uB8Sjt} & 1$\to$2 & \num{adj.T17.snapshot_smax_pre}$\to$\num{adj.T17.snapshot_smax_post} & \num{adj.T17.snapshot_wmax_pre}$\to$\num{adj.T17.snapshot_wmax_post} & F$\to$T & \num{adj.T17.pre_spec_weight}$\to$\num{adj.T17.post_spec_weight} & \num{adj.T17.exact_retained_body_weight_count} \\
Recruitment / \texttt{11-uB8Sjt} & 2$\to$3 & \num{adj.T18.snapshot_smax_pre}$\to$\num{adj.T18.snapshot_smax_post} & \num{adj.T18.snapshot_wmax_pre}$\to$\num{adj.T18.snapshot_wmax_post} & T$\to$F & \num{adj.T18.pre_spec_weight}$\to$\num{adj.T18.post_spec_weight} & \num{adj.T18.exact_retained_body_weight_count} \\
Raffia$^{*}$ / \texttt{12-eIFsKo} & 1$\to$2 & \num{adj.T19.snapshot_smax_pre}$\to$\num{adj.T19.snapshot_smax_post} & \num{adj.T19.snapshot_wmax_pre}$\to$\num{adj.T19.snapshot_wmax_post} & T$\to$T & \num{adj.T19.pre_spec_weight}$\to$\num{adj.T19.post_spec_weight} & \num{adj.T19.exact_retained_body_weight_count} \\
\bottomrule
\end{tabular}
\par\smallskip\begin{minipage}{0.98\linewidth}\footnotesize
Seed display omits only the literal \texttt{row-}. The companion \texttt{TABLE-DATA.csv} retains full IDs, hashes, logged model aliases, and scored weights. $^{*}$Medication has runtime-opening variation despite saved-spec pointer-only edits; raffia also has saved-spec pointer-only edits. Gift email has an unmatched stored weak-tier value; food exchange has a saved/scored criterion mismatch. See the scope and exception paragraphs.
\end{minipage}
\end{aabtable}

\paragraph{Case selection and provenance.}
Trekking illustrates additional evidence demands; recruitment illustrates mixed
disclosure, eligibility, and output changes with worsening loop qualification.
Selection covers these mechanisms rather than maximizing a score decrease.
Descriptions are English paraphrases of saved specifications, coded with AI
assistance; original historical AI judges produced the scores. This coding adds
neither human annotations nor new judge evaluations. The companion data in
\texttt{analysis\_source/adjacent\_revision} retain stable task IDs, versions, source
hashes, JSON pointers, and the machine-readable numerical claim bindings.
Its \texttt{excerpts/CASE-EXCERPTS.json} adds selected original-language fragments
and labeled AI translations for E1 (trekking) and E2 (recruitment), with explicit
redactions and source-span locators. These fragments support inspection of the
quoted wording; public integrity verification does not authenticate withheld sources.

\paragraph{Trekking: added evidence demands.}
Task \texttt{04-row-uBQdVu}, versions r1--r2, already requires source-based equipment
planning, separate personal and shared equipment, and explicit uncertainty.
The later instruction requires adjudicating conflicting distance, elevation, and
duration, preserving rejected values and reasons, and recording owned quantities.
The designated route PDF supplies the required baseline (constraint index 3).
The revised rubric requires consistent conflict decisions across the workbook
(criterion index 7) and actionable verification entries for evidence gaps
(criterion index 8). The released requirement, baseline, and rubric fragments are
\texttt{E1-r2-02}--\texttt{E1-r2-06}. Separately, the workbook-structure disclosure
flag changes from \texttt{proactive} to \texttt{on\_ask}
(\texttt{E1-r1-08}, \texttt{E1-r2-10}); both saved instructions already name the
same workbook sheets (\texttt{E1-r1-11}, \texttt{E1-r2-12}). This flag change
does not establish newly withheld information or runtime disclosure timing.
The supported binding connects conflict adjudication, the designated route PDF,
and rubric obligations for consistent decisions and verification entries. The strong-tier trajectory is
reported beside changing saved weights in Table~\ref{tab:adjacent-revisions}; it is
not a measurement on a fixed rubric.

\paragraph{Recruitment: mixed revision.}
Task \texttt{11-row-uB8Sjt}, versions r2--r3, changes a CSV and Markdown deliverable
to an Excel workbook and Markdown, and revises non-highest-degree eligibility
wording (constraint index 2). Employment-proof facts are available on request in
both versions; the later fact explicitly states the recorded duration. The revised
rubric (criterion index 0, checkpoint index 0) requires eliciting that duration
and prohibits deriving personal values from the number of simulator entries.
The change concerns disclosed content and its evidence requirement, rather than
simply earlier release. Stored strong-tier score rises from \num{adj.T18.snapshot_smax_pre}
to \num{adj.T18.snapshot_smax_post} while loop qualification changes from true to false.
Here the binding connects a requirement to elicit duration, an on-request fact that
states duration, and a criterion that rejects inference from entry counts. Qualification
is the recorded policy outcome; the saved transition does not establish a successful
repair or independent validation of the eligibility rule.
The released E2 fragments show the output change
(\texttt{E2-r2-01}, \texttt{E2-r3-02}), earlier and later elicitation criteria
(\texttt{E2-r2-05}, \texttt{E2-r3-07}), and unchanged \texttt{on\_ask} flags
(\texttt{E2-r2-08}, \texttt{E2-r3-09}). They expose the required
contribution-month field, not the private recorded duration itself.

\paragraph{Linkage and comparison scope.}
Membership requires agreement of all nonempty recorded surface hashes with run
metadata, recurring logged effective model identifiers, scores matching a stored
tier, exclusion of suffixed runs, and no true judge-incomplete flag. Empty hashes
remain missing; only \num{adj.N13} pairs have all \num{adj.N14} surface hashes on both sides.
These bindings do not reconstruct complete environments or equal effective budgets.
Dynamic runs consume logged openings and later disclosures; saved full instructions
are not assumed to have been sent verbatim. No exact version-bound hardening action
was established for these transitions. The separately recovered \num{adj.N15} within-log
pairs with explicit re-hardening markers lack exact specification-version binding
and are not pooled with this stratum. No raw-seed to first-hardened comparison
is established.

\paragraph{Reading the complete table.}
Strong and weak columns preserve snapshot tier values, not model identities or
binary success. Equation~\ref{eq:calibration-profile} cannot reconstruct historical $Q$
without saved thresholds and route settings; Raffia's final $Q=\mathrm{T}$ at $S=W$
is unresolved, not a proven violation.
The companion CSV separately retains both matched models' scores,
scored denominators, and verbatim logged aliases. For example, the records contain
\texttt{openai/gpt-5.6[1m]} and judge route
\texttt{agent:cc\_sdk:openai/gpt-5.6}; neither authenticates a serving checkpoint
or a mapping to the public model names in Table~\ref{tab:model-configurations}. Across \num{adj.N16} matched model-pair comparisons, scores decrease in \num{adj.N17},
increase in \num{adj.N18}, and tie in \num{adj.N19}. These comparisons share tasks and versions.
All rubrics change, with saved total weights changing in \num{adj.N10} pairs. Exact matching
of complete saved criterion bodies, including evidence bindings, and equal weights
retains \num{adj.N20} correspondences across \num{adj.N21} pairs; the other \num{adj.N22} have none.
Complete emitted rubric bodies are recovered on both sides for \num{adj.N25} pairs:
commonness is exact in saved specifications. This criterion does not establish
semantic equivalence or non-equivalence of rewritten criteria. No common scale,
independent-observation significance test, representative effect, or causal
hardening estimate is inferred.

\paragraph{Specific exceptions.}
Medication and raffia change only evidence-capture pointers within saved specifications.
Medication also has a small realized-opening paraphrase (\texttt{first\_user\_full}),
so it is not an unchanged-runtime or controlled rubric-only rescore. Gift email's
pre-version stored weak-tier value, \num{adj.T08.snapshot_wmax_pre}, is absent from the
selected runs; both selected models score \num{adj.T08.model_1_pre_score}.
Food exchange has \num{adj.T04.pre_spec_criteria} saved criteria but \num{adj.N23} graded criteria:
saved total weights are $\num{adj.T04.pre_spec_weight}\rightarrow\num{adj.T04.post_spec_weight}$,
while scored weights are $\num{adj.T04.model_1_pre_scored_weight}\rightarrow\num{adj.T04.model_1_post_scored_weight}$.
No attribution or imputation repairs these exceptions. The exact common-criterion
audit covers only medication and raffia; it supplies neither a fixed-output rescore
nor a comparison for the substantive rewrites.
\FloatBarrier

\subsection{Historical platform-adjusted aggregates}
Table~\ref{tab:model-configurations} preserves the 21 selected historical configurations
and their platform-reported adjusted 100-item aggregates (dataset tag 20260710).
Fixed IDs retain the source crosswalk. This cohort is separate from the October 2026
26-configuration leaderboard; it supplies no domain estimates for Astra6.
The platform adjustment rule and corresponding adjusted per-item observations are
unavailable. Adjusted-score confidence intervals, paired $p$-values, win rates,
and rank-resampling statistics are therefore unavailable and are not reported.
Historical raw-analysis inputs and code remain archived; their statistics are not
estimates of uncertainty for the adjusted aggregates.

\subsection{Construction logs: repeated scores and empirical resampling}
These task-level construction logs use recorded rubric-weighted scores in $[0,1]$,
not platform-adjusted leaderboard aggregates. They describe measurement and selection
within the construction process, not comparative model ability.
\paragraph{Signature-matched archival repeats.}
Runs in Section~\ref{sec:retrospective} are grouped by item, solver and normalized
round tag after excluding records marked invalid or incompletely judged. Figure~\ref{fig:historical-repeats}'s
signature labels are \texttt{cap\_id} capability categories, not rubric-entry indices. Signatures
sort all recorded (\texttt{cap\_id}, weight) pairs, stripping terminal
\texttt{\#digits} but retaining duplicates and exclusions. Matching establishes only
this multiset's equality, not criterion-text, instruction, environment or configuration
identity. Repeats were adaptive and favored unstable verdicts; groups can share items.
These limits apply to score ranges and scalar-threshold crossings.

\begin{aabfigure}[!t]
\centering
\includegraphics[width=\textwidth]{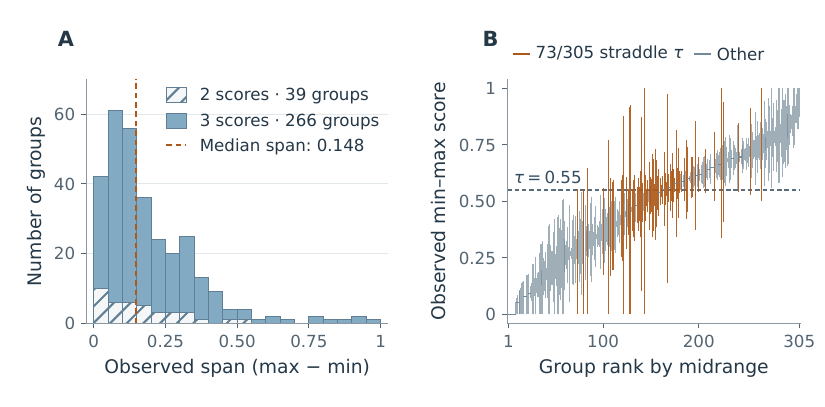}
\caption{\textbf{Observed construction-log spans in selected archival repeat groups.}
Scores are recorded rubric-weighted task values, not adjusted leaderboard aggregates.
Panel A bins observed spans, stacking counts for groups with two and three scores. Panel B sorts the
observed min--max spans by midrange. The \num{q0.sm_n_groups} signature-matched
item--solver--round groups contain \num{q0.sm_n_measurements} scores; \num{q0.sm_n_crossing} groups
satisfy $\min(s)\leq\num{gate.tau}<\max(s)$. Segments show observed minima and maxima,
not confidence intervals. Groups were selected through adaptive repetition,
can share an item, and match multisets of normalized capability labels
(\texttt{cap\_id}) and weights, including excluded rows. This does not authenticate
criterion text or full task versions. Scalar crossing does not establish reversal of the full
multi-condition gate.}
\label{fig:historical-repeats}
\end{aabfigure}

\paragraph{Finite-record effect of minimum selection.}
\label{app:minimum-selection}
We retain all 305 anonymous score lists, comprising 39 lists of length two and
266 of length three. For list $g$, let $\mu_g$ be its mean and $L_g,H_g$ its
extrema. Its observed mean--minimum gap is $\mu_g-L_g$.
A separate functional draws two independent indices uniformly with replacement
from that fixed list. If their scores are $X_1,X_2$, then
\begin{equation}
\begin{split}
\delta_g
&=\mu_g-\mathbb E_{\widehat F_g}[\min(X_1,X_2)]\\
&=\frac{1}{2n_g^2}\sum_{i=1}^{n_g}\sum_{j=1}^{n_g}|x_{gi}-x_{gj}|.
\end{split}
\label{eq:empirical-minimum-decrement}
\end{equation}
Thus $\delta_g=(H_g-L_g)/4$ for $n_g=2$, and
$\delta_g=2(H_g-L_g)/9$ for $n_g=3$, including ties. These identities follow from
$\min(x,y)=(x+y-|x-y|)/2$; the middle score cancels in the three-score case.
We enumerate every ordered pair exactly, without simulation or new judgments.
Independence describes these mathematical draws, not the original observations.

\begin{aabtable}[!htbp]
\centering\small
\caption{\textbf{Selection functionals of the retained score lists.}
Rows stratify by list length $n$; numeric gaps are equal-group means.
The last columns count lists with minimum or mean at most $\tau=0.55$,
and their decision disagreements. Empirical draws reuse retained observations.}
\label{tab:minimum-selection}
\begin{tabular}{@{}lrrrrrr@{}}
\toprule
$n$ & Lists & Mean--min & $\delta_g$ & Min $\le\tau$ & Mean $\le\tau$ & Differ \\
\midrule
2 & 39 & 0.0765 & 0.0383 & 25 & 23 & 2 \\
3 & 266 & 0.0994 & 0.0442 & 167 & 137 & 30 \\
All & 305 & 0.0964 & 0.0434 & 192 & 160 & 32 \\
\bottomrule
\end{tabular}
\end{aabtable}

The median observed gap is 0.0707; the median two-draw decrement is 0.0329.
Extrema alone would bound the three-score mean between $(2L_g+H_g)/3$
and $(L_g+2H_g)/3$, yielding 16--44 possible threshold disagreements across
all groups. The complete retained lists identify 32 exactly.
These are conditional decision differences, not population confidence intervals.
The anonymous projection cannot establish unchanged final versions or distinguish
judging variability from execution, rubric, or task changes. It neither estimates
a defect rate among the 100 packages nor attributes the known defects to minimum
selection. Other qualification conditions are not evaluated by this calculation.

\paragraph{Recovered repeat panel: provenance and selection.}
The separate repeat experiment sampled from a curated corpus with greater representation
near the scalar threshold. The surviving analysis further depends on which executions
produced sufficiently complete repeated scores. After the original scored-run archive
became unavailable, the \num{q1.n_groups} group-level score lists were recovered from a committed
vector figure. Item, tier, and stratum identities cannot be reconstructed.

\paragraph{Empirical resampling results.}
This separately initiated repeat experiment has a surviving panel of
\num{q1.n_groups} recovered score lists and \num{q1.n_measurements} scores,
with \num{q1.crossing_pct}\% of lists crossing the scalar threshold.
Figure~\ref{fig:historical-resampling} preserves every recovered score and the
frozen resampling calculation. Comparing median-threshold decisions from two
independent batches sampled with replacement from each group's observed scores
gives \num{q1.crossing_k1_pct}\% disagreement for single-score batches and
\num{q1.crossing_k5_pct}\% for five-score batches.

\begin{aabfigure}[H]
\centering
\includegraphics[width=\textwidth]{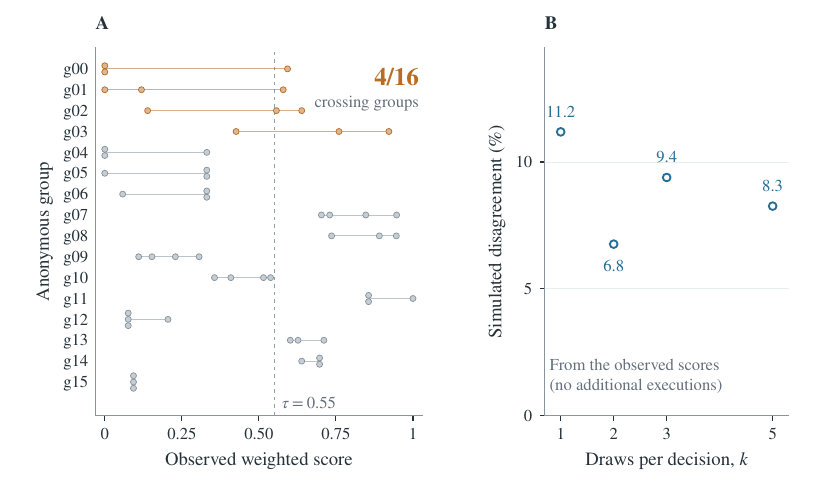}
\caption{\textbf{Construction-log scores and empirical resampling.} Recorded rubric-weighted task scores, not adjusted model aggregates.
The stored values are nonmonotonic: $k=2$ gives $6.8\%$, below $9.4\%$ at $k=3$;
points show discrete settings without interpolating between them. Existing scores are
resampled; no additional executions are represented.
Panel A shows the recovered score lists relative to the scalar threshold
$\tau=\num{gate.tau}$; panel B shows disagreement under empirical resampling.
The \num{q1.n_groups} recovered groups contain \num{q1.n_measurements} scores; duplicate values are offset vertically.
Original item, tier, and stratum identities are unavailable. For each group and $k$,
the frozen calculation draws 4,000 pairs of batches with replacement from the
observed scores and compares their median-threshold decisions
(fixed resampling seed: \texttt{20270118}).
Disagreement estimates are averaged equally across groups. Larger $k$ denotes
more draws from existing scores, not fresh executions, and does not establish
a real-world sampling ceiling.}
\label{fig:historical-resampling}
\end{aabfigure}

\paragraph{Scope of the recovered-panel analysis.}
Larger batches here reuse observed values; they are not additional solver executions.
The panel is selected, small, and lacks recoverable item identities, so it supports
a sensitivity analysis of the observed scores rather than a population noise floor
or a recommendation for the number of fresh runs. The frozen empirical resampling
calculation is retained as a descriptive analysis of those lists; it cannot identify
a population repetition budget or the effect of collecting fresh runs.

\FloatBarrier
\begingroup
\raggedbottom
\subsection{Retention outcomes, round coverage, and judging expenditure}
\paragraph{Recorded outcomes and missing round counts.}
The gate-outcome archive contains \num{po.n_items} records: \num{po.n_accepted} are marked
accepted, corresponding to \num{po.accept_rate_pct}\%, and \num{po.n_salvaged} carry the
salvage outcome, corresponding to \num{po.exhausted_pct}\%. Salvage denotes
retention with a posterior difficulty label rather than a recorded gate pass.
Figure~\ref{fig:outcome-coverage} shows which outcome groups retain round counts.
Score-to-version consistency is not independently established for this archive.
Both recorded outcomes precede independent validity assessment. Round counts are recoverable for
\num{po.rounds_parsed} records, all from the salvage group; accepted records
lack recoverable round counts. The archive therefore does not identify
a rounds-to-acceptance distribution or the benefit of increasing the budget.

\begin{aabfigure}[!ht]
\centering
\includegraphics[width=\textwidth]{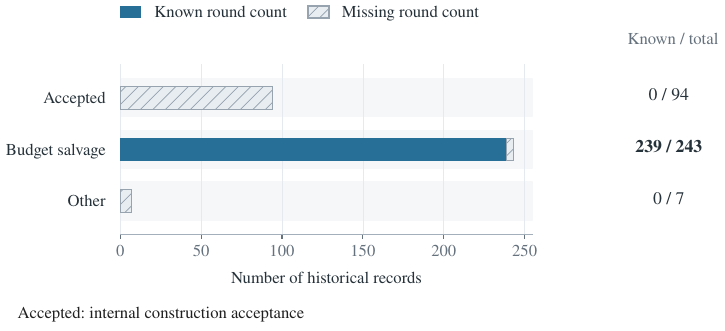}
\caption{\textbf{Archival round-count coverage (344 construction records).} All accepted records lack recoverable
round counts; the recoverable counts belong to the budget-salvage group. This
outcome-dependent missingness prevents the observed round histogram from being
interpreted as a rounds-to-acceptance distribution. Counts are computed directly
from the frozen gate-outcome records. Pale full-width bands are row guides, not record totals; solid and hatched bar segments denote known and missing round counts, respectively.}
\label{fig:outcome-coverage}
\end{aabfigure}

\paragraph{Recorded judging expenditure and missing cost joins.}
The grading archive records total judging expenditure of
\num{cost.total_judge_usd}. Among \num{cost.n_priced} runs with positive
recorded cost, the median is \num{cost.judge_median_usd} per run. Solver
execution and item generation are not separately metered in these figures.
An item-level join between expenditure and gate outcomes is unavailable. These
records support archive expenditure and per-run summaries; estimating
cost per usable output requires complete construction metering and independent
item decisions in the same cohort.

\FloatBarrier
\endgroup
\subsection{October 11 platform audit: quality, endorsement, and coverage}
\label{app:human-platform}
\paragraph{Frame and operational protocol.}
The database snapshot cutoff is October 11, 2026, 09:31:22 UTC. All 11 Daily-Life
version groups are retained, including tasks without review. The frame contains
315 platform tasks, 225 item IDs, and 20,893 answer--criterion records. Assignment
supports ordered/manual allocation and first-come self-claim; the imported tasks
use self-claim with blind evaluation disabled. Review completion has no documented
probability-sampling provenance. Version groups in Table~\ref{tab:human-frame} are
observed platform cohorts, not independent samples or authenticated runtime versions.

\begin{aabtable}[!htbp]
\centering\small
\caption{\textbf{Operational review coverage by Daily-Life version.} Task columns
count containers with at least one endorsement review. Criterion columns count
unique answer--criterion records, including both agreement and disagreement.
Repeated items and multiple criteria per answer prevent interpreting these as
independent observations. Operational selection was not random.}
\label{tab:human-frame}
\begin{tabular}{@{}lrrrr@{}}
\toprule
Version & Task frame & Tasks reviewed & Criterion frame & Criteria reviewed \\
\midrule
Unversioned & 100 & 92 & 13,942 & 1,304 \\
V2.0 & 22 & 22 & 628 & 550 \\
V2.1 & 8 & 7 & 232 & 191 \\
V2.2 & 33 & 32 & 980 & 931 \\
V2.3 & 20 & 15 & 572 & 342 \\
V2.3.1 & 12 & 12 & 346 & 346 \\
V2.4 & 13 & 13 & 406 & 406 \\
V2.5 & 18 & 18 & 666 & 666 \\
V2.6 & 66 & 66 & 2,044 & 2,044 \\
V2.7 & 20 & 20 & 987 & 940 \\
V2.8 & 3 & 3 & 90 & 90 \\
\midrule
All & 315 & 300 & 20,893 & 7,810 \\
\bottomrule
\end{tabular}
\end{aabtable}

\begin{aabtable}[!htbp]
\centering\small
\caption{\textbf{Human task-quality observations.} Each dimension has 306 review
records, covering 302/315 platform tasks and 216 item IDs, attributed to eight
accounts. Selection was operational, not a documented random sample. Positive
opinions do not certify execution of the final package.}
\label{tab:human-quality}
\begin{tabular}{@{}lrr@{}}
\toprule
Quality dimension & Positive & Negative \\
\midrule
Instruction clarity & 300 & 6 \\
Environment completeness & 305 & 1 \\
Solution reasonableness & 271 & 35 \\
Rubric reasonableness & 290 & 16 \\
\bottomrule
\end{tabular}
\end{aabtable}

Of the 306 quality records, 256 are positive on all four dimensions; attribution is to eight accounts.

\paragraph{Visible evidence and labels.}
The current interface exposes rubric statements and weights, check points, failure
examples, reference answers, and machine verdicts, reasons, confidence, and vote
details. The service also supplies the as-tested prompt, stored final response, and
trajectory/product references. Availability does not show that a reviewer opened
all evidence. The UI records agreement and derives \texttt{final\_verdict} by adopting
or inverting the machine verdict. The dashboard sets \texttt{human\_verdict=None},
but the underlying endorsement records do store \texttt{final\_verdict}.
In 404/7,703 binary records this stored value conflicts with the current derivation;
source history confirms the machine-visible formula and backend field validation
without cross-field reconciliation. The historical request and submission-time
machine state are unavailable, so all 404 causal origins remain unknown. Of these
conflicts, 381 are QA-approved and 23 unreviewed: QA approval is neither a new blind
label nor a consistency certificate. Neither an independent label nor its opposite
is inferred. Five answer--criterion records have two reviews (four concordant,
one discordant); these do not constitute an independent interrater study.

\begin{aabtable}[!htbp]
\centering\small
\caption{\textbf{Machine-visible endorsement in the October audit.}
Numerators count agreement decisions; denominators count review records. Explicit
JSON-null machine objects and excluded criteria remain separate from binary verdicts.
Skip records are outside the endorsement denominator.}
\label{tab:human-endorsement-october}
\begin{tabular}{@{}lrrrr@{}}
\toprule
Machine state / review subset & Endorsed & Reviewed & Contested & Endorsement (\%) \\
\midrule
Pass & 2,800 & 3,017 & 217 & 92.81 \\
Fail & 4,629 & 4,686 & 57 & 98.78 \\
Excluded & 102 & 102 & 0 & 100.00 \\
Explicit JSON-null machine object & 8 & 10 & 2 & 80.00 \\
\midrule
All endorsement records & 7,539 & 7,815 & 276 & 96.47 \\
Binary only & 7,429 & 7,703 & 274 & 96.44 \\
\midrule
QA-approved & 6,501 & 6,758 & 257 & 96.20 \\
QA-unreviewed & 1,037 & 1,055 & 18 & 98.29 \\
QA-rejected & 1 & 2 & 1 & 50.00 \\
\bottomrule
\end{tabular}
\end{aabtable}

\paragraph{Incomplete review and QA.}
The 7,815 endorsement records refer to 7,810 unique answer--criterion records;
13,083 frame records have no endorsement. There are also 202 skip records, giving
8,017 answer-stage score records. In the full answer frame, machine states are
9,190 pass, 10,308 fail, 1,058 excluded, and 337 explicit JSON-null objects.
These frame counts are not review counts. Of the 306 quality records,
239 are QA-approved, 66 unreviewed, and one resubmitted. QA reviewers can see
existing evaluator judgments; QA is not a blind second reference. No endorsement
record is marked QA-modified in this snapshot. The quality form also contains an
overall accept/revise/reject field, whose values were not retained in the audit
export; the four dimension ratings are not used to reconstruct it. Account counts
establish stored attribution, not independently verified expertise or independence
from construction. Recruitment, consent, compensation, and full historical reviewer
instructions are not established by these records.

\paragraph{Artifact joins and runtime boundaries.}
The audit joins project/version to task, item--model--criterion, assignment,
review account, score, and QA history; a separate task-to-current-package join
compares rubric text and weights. All 4,716/4,716 current rubric statements and
weights match the database across 315 candidate packages. This establishes current
rubric alignment, not historical execution identity. Current files specify
\path{openai/gpt-5.5} with one vote in 100 packs and \texttt{gpt-5.6-sol}
with three votes in 89; 126 lack a local judge configuration. These are file
observations, not authenticated historical judge calls. The frame contains
19,119 as-tested-prompt hashes, 6,323 final-response hashes, 20,893 trajectory-reference
hashes, and 20,768 product-reference hashes. Reference hashes do not authenticate
referenced content. No local actual-call trace files or actual judge-runtime metadata
were recovered, so the review data cannot be assigned to one verified current judge.

\paragraph{Release-sample component binding.}
For the same 20 packages in Table~\ref{tab:release-audit}, all 20 have an
instruction byte match to at least one platform record, and 14/20 have a rubric
byte match. The unit is a sampled release package with a matching component, not a
review record or execution. Metadata differences and incomplete solution and
saved-specification provenance prevent establishing complete package identity.
These matches leave every mechanical PASS/FAIL/UNKNOWN status in
Appendix~\ref{app:release-audit} unchanged and do not transfer platform quality
opinions to full-package validity.

\paragraph{Sensitivity to stored-verdict conflicts.}
For the published margins, hypothetical conflict exclusion permits either
machine-state ordering of machine-visible endorsement.
The unit is an endorsement review record, not a distinct answer--criterion
record; a conflict flags a stored \texttt{final\_verdict} that differs from
the current agreement-based derivation.
Assuming these flags refer to the binary review-record population in
Table~\ref{tab:human-endorsement-october}, exclusion fixes the pooled denominator, while the numerator
depends on the unknown overlap between conflicts and agreement.
Nonbinary excluded and explicit JSON-null records remain unchanged.
The pooled bounds place as many conflicts as possible among endorsed or
contested binary records.
The machine-fail minus machine-pass endorsement gap is minimized by removing
all contested machine-pass records and assigning the remaining conflicts
to endorsed machine-fail records; interchanging machine-pass and
machine-fail maximizes the gap.
QA totals cover all machine states.
Each endpoint allocation fits within the QA endorsed and contested counts;
retained binary and unchanged nonbinary records fill the remaining capacities.
Each endpoint is therefore separately attainable under all published
machine-state, agreement, QA and conflict-count margins.
The post-exclusion rates and gap sign are not point-identified without the
conflict-by-agreement-by-machine-state cross-tabulation.
These are finite-record sensitivity bounds, not confidence intervals or
scorer-accuracy estimates; unique-criterion coverage and item-cluster
intervals require record-membership maps.

\paragraph{Snapshot lineage and public reproduction.}
All 7,815 Daily-Life endorsement records in the September and October snapshots
form identical multisets on the shared project, task, item, model, rule, and
UTC submission-time keys, with identical shared machine verdicts, agreement,
QA status, and reason-presence values. Reviewer aliases were independently
anonymized and were not compared. The earlier export's \texttt{human\_verdict=None}
does not imply absence of a stored final label. September's exported ``Missing''
category remains historical wording; October directly observes JSON-null machine
objects. October therefore adds a refreshed lineage and coverage audit, including
existing quality observations, rather than new human annotation or an independent
replication. We report no new October interval and retain the dated September
calculation and its original interval unchanged.
The public \path{anc/analysis/analysis_source/human_platform_audit/} directory provides
aggregate counts and an offline arithmetic check. From \path{anc/analysis/}:
\begin{quote}\small
\begin{verbatim}
cd analysis_source
cd human_platform_audit
python3 reproduce.py
\end{verbatim}
\end{quote}
This recomputes tables and coverage from aggregates; it does not reproduce database
extraction or independent annotation. Raw review rows, account aliases, identifiers,
questionnaires, and internal locations are not distributed.

\subsection{Machine-state recoding and recorded endorsement}
\label{sec:verdict-correction-sensitivity}

Hypothetical machine-state recoding can change the endorsement contrast while
preserving the recorded agreement responses. This fixed-population analysis
changes the \texttt{pass}/\texttt{fail} labels used to group those responses;
it does not recompute agreement with a new machine decision. Under the same
common-population and binary-conflict assumptions used for exclusion sensitivity,
at most \(k\) conflict-affected review records could switch between
\texttt{pass} and \texttt{fail}. Recorded responses, quality categories and
nonbinary-state records were held fixed. A discrepancy was only eligible for a
hypothetical change; it was not assumed to require one. The conflict flags concern
stored \texttt{final\_verdict} inconsistencies, not verified errors in machine-state labels.

Let \(a\) and \(b\) be the numbers of endorsed and contested records whose
snapshot machine state is \texttt{pass}, and let \(E\) and \(D\) be their
totals across the binary states. A recoded table with \(j\) endorsed and
\(\ell\) contested \texttt{pass} records requires at least
\[
  e=|j-a|,\qquad n=|\ell-b|,\qquad k_{\min}=e+n
\]
machine-state recodings. This lower bound is attainable by moving records only
in the required direction within each endorsement category. For quality category
\(q\), \(E_q,D_q\) are the endorsed and contested margins across all machine
states, including excluded and JSON-null records, and \(C_q\) is its reported
conflict count. The full inventory \(M=\sum_q C_q\) is fixed; \(k\) is only
an at-most recoding cap. The number of endorsed binary conflicts lies in
\[
\begin{aligned}
 L&=\max\!\left\{\sum_q\max(0,C_q-D_q),\,M-D\right\},\\
 U&=\min\!\left\{\sum_q\min(C_q,E_q),\,E\right\}.
\end{aligned}
\]
Under this model, the recoded table is feasible precisely when its cells
\(j,\ell,E-j,D-\ell\) are nonnegative, both binary groups are nonempty, and
\[
 \max(L,e)\ \leq\ \min(U,M-n),\qquad e+n\leq k.
\]
The identified set includes all joint allocations matching the published
margins, with no additional quality-by-state structural zeros or record-level
linkage restrictions. Under this model, any endorsed conflict total in the
intersection can be distributed across quality categories and then across the
original machine states; the remaining unconflicted margins form a nonnegative
transportation table. This establishes sufficiency within the aggregate model,
not an observed linkage or independence assumption. Constructive endpoint
witnesses accompany the source data.

For each feasible table, the endorsement-rate contrast is
\[
\begin{aligned}
 \Delta(j,\ell)
 &=\frac{E-j}{E+D-j-\ell}-\frac{j}{j+\ell}\\
 &=\frac{E(j+\ell)-j(E+D)}
        {(j+\ell)(E+D-j-\ell)},
\end{aligned}
\]
where positive values mean greater endorsement among \texttt{fail} records.
We enumerated the finite feasible set with exact integer and rational arithmetic.

Under this model, the original ordering was retained for every feasible
allocation with fewer than 114 machine-state recodings; a strict reversal was
attainable at 114 recodings. Table~\ref{tab:verdict-correction-frontier} reports
selected points of the sharp at-most-budget envelope. The complete numerical
frontier is retained. Each retained joint witness starts from an allocation
matching the published state, endorsement, quality and conflict margins. Applying
its declared flips gives the reported recoded table: binary-state margins may
change, while endorsement totals, quality margins and nonbinary records remain fixed.

\begin{table}[t]
\centering
\caption{Fixed-population machine-state recoding sensitivity. Bounds are for
\(\Delta\), recorded endorsement among recoded \texttt{fail} minus
\texttt{pass} records, in percentage points. The recoding cap is counterfactual;
no observed records were relabelled or removed. Display values are rounded,
whereas enumeration and threshold decisions used exact fractions.}
\label{tab:verdict-correction-frontier}
\begin{tabular}{rrr}
\toprule
\shortstack[r]{At most\\recoded states} & Lower bound & Upper bound \\
\midrule
0 & 5.98 & 5.98 \\
113 & 0.04 & 9.08 \\
114 & -0.01 & 9.08 \\
404 & -5.81 & 10.05 \\
\bottomrule
\end{tabular}
\end{table}

All-record endorsement is invariant under machine-state recoding, because
neither the recorded responses nor the reviewed population changes. It therefore
cannot resolve the direction of the recoded binary-state contrast. The attainable
reversal demonstrates sensitivity to hypothetical machine-state assignments,
not which labels are erroneous or how accurately the evaluator judged the records.
Linked adjudication is needed to assess those questions; the original review
remained machine-visible and nonblind.

\subsection{September 20 snapshot: non-blind reviewer endorsement}
\label{app:human-agreement}
\paragraph{Snapshot selection and observation units.}
The anonymized platform snapshot was retrieved on September 20, 2026; the latest
Daily-Life review was submitted on August 27, 2026. It is an expanded historical
audit, not a new annotation experiment. We retain all platform project groups in the Daily-Life application series,
including every named version, irrespective of agreement or platform QA status. The resulting \num{hja2.daily_n} criterion
reviews cover \num{hja2.n_items} distinct item identifiers, \num{hja2.n_tasks}
platform tasks, \num{hja2.n_solver_labels} solver labels, and \num{hja2.n_reviewers}
reviewer accounts. The unfiltered export contains \num{hja2.all_n} reviews across
applications. Repeated items across tasks remain associated within item clusters;
these units are not interchangeable independent sample counts.

\paragraph{Review procedure and interpretation.}
Reviewers saw machine judgments;
the historical export exposes agreement responses and does not establish independent
human pass/fail labels. The newer direct audit recovers a stored final-verdict field
and explains its dependence on visible machine judgments (Appendix~\ref{app:human-platform}). Thus the result measures non-blind endorsement, not blind judge accuracy,
human--human reliability, or package usability. Historical model labels and batches
do not authenticate one common current judge configuration.
Table~\ref{tab:human-endorsement-october} separates endorsement rates by machine verdict.

The canonical counts are in Table~\ref{tab:human-endorsement-october}; its
``explicit JSON-null'' row corresponds to September's historical ``Missing'' label.
The shared records are counted once. The dated September weighting and interval
below remain unchanged.

\paragraph{QA sensitivity and snapshot separation.}
The primary denominator includes all retained decisions. Platform-QA-approved
records contain \num{hja2.qa_approved_agree} endorsements among
\num{hja2.qa_approved_n} reviews (\num{hja2.qa_approved_pct}\%);
QA-unreviewed records contain \num{hja2.qa_unreviewed_agree} of
\num{hja2.qa_unreviewed_n} (\num{hja2.qa_unreviewed_pct}\%), and QA-rejected
records contain \num{hja2.qa_rejected_agree} of \num{hja2.qa_rejected_n}
(\num{hja2.qa_rejected_pct}\%). QA approval is platform metadata, not an
independently elicited pass/fail label. Restricting this snapshot to the project
names used in the July snapshot yields \num{hja2.old_groups_agree} endorsements
among \num{hja2.old_groups_n} decisions (\num{hja2.old_groups_pct}\%); the
snapshots are retained separately and are not pooled or interpreted as a trend.

\paragraph{Weighting and item-cluster uncertainty.}
Equal-item and equal-reviewer averages are \num{hja2.item_macro_pct}\% and
\num{hja2.reviewer_macro_pct}\%. For this September historical snapshot only, the retained interval is
$[\num{hja2.ci_lo_pct},\num{hja2.ci_hi_pct}]\%$. We use
\num{hja2.bootstrap_n} item-cluster resamples with replacement (seed
\num{hja2.bootstrap_seed}), retaining every associated criterion, task, solver,
and reviewer record. Each resample recomputes the pooled endorsement ratio;
the central \num{hja2.ci_level}\% percentile interval conditions on the observed
archive and fixed reviewers. The supplement provides anonymous per-item endorsed/total
counts and \texttt{endorsement/reproduce.py}, preserving the original item order and
random-number procedure. These sufficient aggregate counts reproduce the pooled and
item-macro rates and interval without distributing individual review records.

\FloatBarrier
\subsection{July 30 snapshot: earlier non-blind endorsement audit}
\label{app:human-agreement-july}
\paragraph{Snapshot selection and observation units.}
We recompute agreement from the frozen platform export dated July 30, 2026,
using the four explicitly named Daily-Life project groups. This retains
\num{hja.daily_n} criterion-level reviews and excludes unrelated application
records; the unfiltered archive contains \num{hja.all_n} reviews with
\num{hja.all_agree} endorsements (\num{hja.all_pct}\%). The application-specific
cohort spans \num{hja.n_items} distinct item identifiers,
\num{hja.n_tasks} platform tasks, and \num{hja.n_item_model_task_groups}
task--item--solver groups. These are different units; platform groups are not
independent task counts. Project labels identify historical batches, not a
verified common judge-model, prompt, rubric, or task-version configuration.

\paragraph{Review procedure and endorsement definition.}
A reviewer sees the machine judgment and records agreement or disagreement;
each disagreement in the export has a written reason. We define agreement as
the number of endorsed criterion decisions divided by the number of reviewed
criterion decisions. The export contains no independently elicited human
pass/fail verdicts. We therefore report endorsement rates, without inferring
blind classification accuracy, chance-corrected agreement, or inter-rater
reliability. This is evidence about archived criterion judgments, not a validated
single current scorer configuration or independently assessed package usability.
Table~\ref{tab:human-agreement} reports this earlier snapshot separately.

\begin{aabtable}[!htbp]
\centering\small
\caption{\textbf{Human endorsement by machine verdict in the Daily-Life audit.}
All retained reviews have an agreement decision. ``Excluded'' denotes a
non-scored criterion. Contested pass and fail decisions indicate different
directions of disagreement, without assuming that the human response is an
independent ground-truth verdict.}
\label{tab:human-agreement}
\begin{tabular}{@{}lrrrr@{}}
\toprule
Machine verdict & Reviewed & Endorsed & Contested & Agreement (\%) \\
\midrule
Pass & \num{hja.pass_n} & \num{hja.pass_agree} & \num{hja.pass_disagree} & \num{hja.pass_pct} \\
Fail & \num{hja.fail_n} & \num{hja.fail_agree} & \num{hja.fail_disagree} & \num{hja.fail_pct} \\
Excluded & \num{hja.excluded_n} & \num{hja.excluded_agree} & \num{hja.excluded_disagree} & \num{hja.excluded_pct} \\
\midrule
All & \num{hja.daily_n} & \num{hja.daily_agree} & \num{hja.daily_disagree} & \num{hja.daily_pct} \\
Pass/fail only & \num{hja.binary_n} & \num{hja.binary_agree} & \num{hja.binary_disagree} & \num{hja.binary_pct} \\
\bottomrule
\end{tabular}
\end{aabtable}

\paragraph{Weighting and uncertainty.}
The primary rate weights reviewed criteria equally. Equal-item and equal-reviewer
averages are \num{hja.item_macro_pct}\% and \num{hja.reviewer_macro_pct}\%,
respectively; individual reviewer rates range from
\num{hja.reviewer_rate_min_pct}\% to \num{hja.reviewer_rate_max_pct}\%.
For this earlier snapshot, we draw
\num{hja.bootstrap_n} bootstrap samples of item identifiers with replacement
(seed \num{hja.bootstrap_seed}), retaining all associated criteria, tasks,
solver labels, and reviewer records within each sampled item. Each replicate
recomputes the pooled ratio; the central \num{hja.ci_level}\% percentile range
is $[\num{hja.ci_lo_pct},\num{hja.ci_hi_pct}]\%$. This quantifies sensitivity to resampling the observed item mixture,
conditional on the fixed reviewers.

The retained public records do not reconstruct historical reviewer recruitment,
consent and compensation procedures, or full instructions; endorsement counts
do not supply these procedural details.

\paragraph{Selection and annotation limits.}
Non-random review selection, exposure to
machine decisions, shared reviewers, and unavailable first-pass revision histories
limit extrapolation to new items or reviewers. No matched criterion has multiple
independent reviewers in the retained export. Independent blinded annotation
and human--human agreement would strengthen this audit, but cannot be recovered
by relabeling the existing endorsement records.
\subsection{Current configuration provenance}
\label{app:current-configurations}
Table~\ref{tab:current-dailylife} preserves all 26 selected source configurations,
including dated revisions, effort suffixes, explicit Preview labels, and the distinct
Codex and CC observations. Unspecified harnesses remain unspecified. The source
snapshot was modified October 10, 2026 and copied October 11, 2026. These labels do
not independently establish general availability, serving checkpoints, task versions,
or evaluator versions. Seed2.0 is absent from the current source. The separately sourced historical reference is explicitly Seed2.0 Lite / 0428 / high, not a generic Seed2.0 or Pro entry.
Public cell IDs in \path{anc/figure_source/current_dailylife/leaderboard.csv}
map to exact private source cells without disclosing the source location.

\subsection{Historical solver configurations}
\label{app:model-configurations}
Table~\ref{tab:model-configurations} lists the 21 historical author-selected configurations,
with stable original IDs (B, K, and P omitted). Labels and suffixes are archived;
vendor references identify public names, not evaluated runs or serving checkpoints.
Scoring details, execution budgets, and backend identities remain unverified. These archived labels and 100-item platform aggregates must not be updated to October 2026 names or scores: the current 26-configuration cohort is Table~\ref{tab:current-dailylife}. Selection was post hoc; omission of B, K, and P does not establish invalid source measurements.

\begin{aabtable}[!t]
\centering\normalsize
\caption{\textbf{Historical solver labels and configurations (dataset tag 20260710).}
IDs preserve the archived source crosswalk. Scores are the platform-reported adjusted
100-item aggregates; adjusted per-item observations and uncertainty are unavailable.
A dash denotes no suffix in the public label. Developer references identify public names only, not evaluated runs.}
\label{tab:model-configurations}
\setlength{\tabcolsep}{4pt}
\renewcommand{\arraystretch}{1.10}
\begin{tabular}{@{}lp{0.47\textwidth}lr@{}}
\toprule
ID & Recorded model / reference & Suffix & Score (\%) \\
\midrule
A & Claude Opus 5~\citep{anthropic2026opus5} & \texttt{max} & 78.17 \\
C & GPT-5.6 Sol~\citep{official2026entryc} & \texttt{max} & 74.87 \\
D & Grok 4.5~\citep{official2026entryd} & --- & 73.40 \\
E & Claude Opus 4.8~\citep{official2026entrye} & \texttt{max} & 73.13 \\
F & GPT-5.5~\citep{official2026entryf} & \texttt{xhigh} & 71.99 \\
G & Claude Sonnet 5~\citep{official2026entryg} & \texttt{max} & 68.98 \\
H & GPT-5.6 Terra~\citep{official2026entryh} & \texttt{max} & 68.49 \\
I & Seed2.1 Pro~\citep{official2026entryi} & \texttt{0628 / high} & 66.09 \\
J & Gemini 3.6 Flash~\citep{official2026entryj} & \texttt{high} & 63.47 \\
L & GPT-5.6 Luna~\citep{official2026entryl} & \texttt{max} & 61.93 \\
M & MiniMax M3~\citep{official2026entrym} & --- & 61.57 \\
N & Kimi K2.7 Code~\citep{official2026entryn} & --- & 58.55 \\
O & GLM-5.2~\citep{official2026entryo} & \texttt{max} & 56.66 \\
Q & DeepSeek V4 Pro~\citep{official2026entryq} & \texttt{max} & 54.92 \\
R & Kimi K2.6~\citep{official2026entryr} & --- & 54.14 \\
S & Qwen3.7 Plus~\citep{official2026entrys} & \texttt{high} & 52.63 \\
T & Gemini 3.5 Flash~\citep{official2026entryt} & \texttt{high} & 51.15 \\
U & Gemini 3.1 Pro~\citep{official2026entryu} & \texttt{high} & 45.96 \\
V & Qwen3.7 Max~\citep{official2026entryv} & \texttt{high} & 45.48 \\
W & Gemini 3.5 Flash Lite~\citep{official2026entryw} & \texttt{high} & 43.16 \\
X & Seed2.0 Lite~\citep{official2026entryx} & \texttt{0428 / high} & 40.41 \\
\bottomrule
\end{tabular}
\par\smallskip\parbox{\linewidth}{\footnotesize Other names remain archived labels; only Solver A has explicit author confirmation of its public-name mapping. Cohort selection is disclosed in Methods. HIST-X is Seed2.0 Lite / 0428 / high: its source aggregate is 40.414\%, rounded here to 40.41\%; it is not a current leaderboard row.}
\end{aabtable}

\paragraph{Requested scorer configuration for the later fixed-output recording.}
The recording in Appendix~\ref{app:judge-controls-r310} requested the harness routing
alias \texttt{gpt-6-astra} for every role, including the visual role. This is the
configuration string the harness was asked for, recorded verbatim; it is deliberately
kept distinct from the public model names above and is not
evidence of a backend identity, a serving checkpoint, or a serving provider. No reference is
supplied for it, and no released model is substituted for it. The earlier recording in
Appendix~\ref{app:judge-controls} requested a different alias, so the two recordings are
not the same configuration.

\subsection{Current DailyLife configuration records and domain coverage}
\label{app:current-dailylife-details}
\label{app:current-dailylife-plot}
Table~\ref{tab:current-dailylife} reports our DailyLife contribution from the
authoritative source snapshot modified October 10, 2026 and copied October 11, 2026.
The author-selected published reporting cohort contains 26 configurations.
GPT-6-Astra-max (2026.09.04)-Codex has the largest displayed aggregate, 83.1\%;
Claude Opus 5-max (2026.07.24) records 81.4\%. GPT-5.6-Sol-max (2026.07.09)
records 77.2\% with Codex and 78.8\% with CC; Seed-2.1-Pro-260628-high records
66.3\%. These are configuration observations, with mixed or unspecified harnesses,
not model-only causal effects. The score ordering does not establish a true ranking.

Platform reports supply 11-domain breakdowns for 17 configurations, with the same
20260710 exercise versions and 100-item count vector. Associations use official
labels, authorized aliases, task configuration references, exact totals, and reuse
lineage; reused reports do not count as independent runs. Nine breakdowns remain
unavailable. These records do not independently attest backend weights or serialized
request parameters, nor establish a common budget or harness.
No paired tests, rank intervals, or domain
correlations from the historical archive apply to this table. The separately dated
Qwen runs and the two explicit GPT harness configurations remain distinct.
Seed2.0 is absent and has no reported numeric score. Preview labels are preserved;
source naming alone does not independently confirm general availability.

\paragraph{Separately sourced historical reference.}
The earlier Insight archive records \textbf{Seed-2.0-lite-0428-high} at
\textbf{40.414\%} (exported platform aggregate 0.40414) over its historical 100-item
panel, dataset version tag 20260710; the exact collection date is unavailable.
This author-requested reference retains the Lite variant and has no current-source
entry. It is excluded from the 26-row current table and from any combined
ranking or uncertainty comparison. Its public provenance ID is HIST-X.
\begin{center}\small
\begin{tabular}{@{}llr@{}}
\toprule
Historical reference & Source / dataset tag & Aggregate \\
\midrule
Seed-2.0-lite-0428-high & Insight / 20260710 & 40.414\% \\
\bottomrule
\end{tabular}
\end{center}

The complete 26-configuration table appears in the main results
(Table~\ref{tab:current-dailylife}); its original rows, scores, citations,
and provenance identifiers are unchanged.
Figure~\ref{fig:current-dailylife} shows only the 17 complete domain vectors, sorted by unrounded official overall; all 26 valid aggregate records remain in Table~\ref{tab:current-dailylife}.

\subsection{Attainable leadership under common domain weights}
\label{app:domain-linear-support}

Pairwise domain crossings do not establish that a configuration can lead a
whole comparison cohort under a common scoring rule. We therefore test a
declared family of linear domain preferences. Let $x_i\in\mathbb{Q}^{D}$ be the
reported domain vector of configuration $i$, retaining the original domain
scales and higher-is-better direction. The cohort contains exactly the
configurations with complete domain vectors in the retained analysis; missing
coordinates are not imputed. Dated model--harness identities remain separate,
including configurations with identical vectors. This analysis neither
reconstructs nor replaces the benchmark's adjusted aggregate.

\paragraph{Support and ties.}
Configuration $i$ is positively supported if there is a common weight vector
that makes it co-leading against every configuration in the cohort:
\begin{equation}
  \begin{gathered}
  w_d>0,\qquad \sum_{d=1}^{D}w_d=1,\\
  w^{\mathsf T}(x_i-x_j)\geq 0\quad\text{for every }j.
  \end{gathered}
  \label{eq:positive-domain-support}
\end{equation}
Co-leading permits ties; it does not assert a unique winner. We also test the
weaker condition obtained by allowing $w_d=0$. A configuration is
\emph{boundary-only supported} when this weaker condition holds but positive
support is excluded. Such support may discard domains and is therefore kept
distinct from a preference rule that retains every domain. The comparison
inequalities are homogeneous: positive support can equivalently be tested
with unnormalised weights $w_d\geq 1$ and the witness then normalised. This
avoids imposing an arbitrary positive lower bound on normalised weights.

\paragraph{Exact certificates.}
An accepted supporting witness has rational weights whose signs,
normalisation and comparisons against the full cohort are checked exactly.
For an exclusion certificate, nonnegative rational coefficients
$\lambda_j$ sum to unity over other configurations and define
$z=\sum_{j\ne i}\lambda_jx_j$. If $z_d\geq x_{id}$ in every domain, with strict
inequality in at least one, then every strictly positive weighted score
favours $z$ over $x_i$. Consequently at least one constituent has a greater
score, so $i$ cannot be positively supported. If the inequalities are strict
in every domain, the same argument excludes support under any nonnegative
normalised weights. The mixture is an algebraic certificate about the fixed
vectors, not an evaluated or deployable model ensemble.

For a finite cohort these alternatives are complete. Write the rows of $A$
as $x_j-x_i$ for $j\ne i$. Positive support is equivalent to feasibility of
$Av\leq-A\mathbf{1}$, $v\geq0$, after setting $w=\mathbf{1}+v$.
The finite linear alternative gives, in the infeasible case,
$\lambda\geq0$ with $A^{\mathsf T}\lambda\geq0$ and
$\mathbf{1}^{\mathsf T}A^{\mathsf T}\lambda>0$.
Normalising $\lambda$ gives the first exclusion certificate.
For nonnegative support use $Aw\leq0$, $w\geq0$ and
$\mathbf{1}^{\mathsf T}w\geq1$; its infeasibility gives
$A^{\mathsf T}\lambda\geq t\mathbf{1}$ for some $t>0$, yielding the
strict certificate. No full-dimensionality or unique-vertex assumption is
needed. The lower bound before normalisation is valid because the weight
inequalities are homogeneous.

\paragraph{Numerical and interpretive scope.}
Source decimal tokens are converted directly to exact rational coordinates.
A numerical solver may propose witnesses or mixtures, but a tolerance,
rounded score or solver infeasibility flag alone does not determine a
scientific classification. Every definite result requires the corresponding
exact certificate; a case without one remains unresolved. A boundary-only
classification requires both a nonnegative supporting witness and a
positive-support exclusion certificate. If a domain is constant throughout
the cohort, concentrating all weight there can make nonnegative support
uninformative, which further motivates the positive-weight distinction.
The conclusions concern the reported finite cohort and this linear
preference family. They are not sampling-uncertainty estimates or evidence
about newly evaluated models, adaptive construction efficacy, or unobserved
configurations.
\paragraph{Target-wise perturbation boundary.}
The additional exclusions are boundary-only supported: an all-coordinate
strict mixture gap would contradict their nonnegative support. Fix such a
target $i$ and a supporting simplex vector $q$, so that $q_d\geq 0$,
$\sum_d q_d=1$ and $q^{\mathsf T}(x_i-x_j)\geq 0$ for every $j$.
For any $\varepsilon>0$, we construct a hypothetical table with
$\max_{j,d}|x'_{jd}-x_{jd}|<\varepsilon$ on which $i$ has positive support.
The perturbation model permits independent coordinate changes in the ambient
space; it does not specify a probability distribution for evaluation error.

Let $G_i=\{j:x_j=x_i\}$ retain all identities with the target's reported vector,
and set $u=\mathbf{1}/D$. Let $M$ be the largest of zero and the values
$u^{\mathsf T}(x_j-x_i)$ for $j\notin G_i$, taking $M=0$ if there is no row
outside $G_i$.
Choose
\[
\begin{aligned}
  \eta &= \frac{\varepsilon}{4(M+\varepsilon)},\\
  w &= (1-\eta)q+\eta u .
\end{aligned}
\]
The weights are strictly positive and sum to unity. Add $\varepsilon/2$ to
every coordinate of each row in $G_i$, leaving all other rows unchanged.
The table distance is $\varepsilon/2<\varepsilon$. For $j\notin G_i$, the
nonnegative-support inequality and the definition of $M$ give
\[
\begin{aligned}
  w^{\mathsf T}(x'_i-x'_j)
    &\geq \varepsilon/2-\eta M\\
    &> \varepsilon/4>0 .
\end{aligned}
\]
Every row in $G_i$ remains tied with the target, which strictly leads all
other rows. Thus one strictly positive weighting, common to the entire
comparison cohort, makes the target co-leading without discarding duplicates.

Since the construction works for every $\varepsilon>0$, no positive ambient
radius guarantees that the target remains excluded under all allowed
perturbations. For the additional pairwise-nondominated exclusions, a joint
guarantee preserving both nondominance and exclusion is therefore also absent.
This establishes the boundary constructively, beyond the non-positive
sufficient-margin result alone. The table and weights can differ across
targets; the argument does not assert that the strict set inclusion disappears
in a common nearby table.

\paragraph{Source data and offline certificate checks.}
In \path{anc/domain-support/}, \texttt{domain-support.source-data.json} retains every qualified complete-domain configuration, the original domain scores and coordinate order, the support classifications, and the exact weight or convex-mixture certificates, together with the original exclusion accounting.
The companion \texttt{FIELD-GUIDE.md} defines the fields and preserves the distinction between strictly positive support, boundary support, and co-leading with ties.
The supplied \texttt{verify\_source\_data.py} provides offline inspection and explicit certificate verification; the commands and their scope are given in \texttt{README.md}.
Convex-mixture certificates are algebraic exclusion evidence, not evaluated ensembles.

The symbolic construction in \path{anc/boundary-support/construct_boundary_support.py} checks the target-wise
perturbation argument separately from the empirical domain certificates.

\subsection{Public model-reference scope}
\label{app:model-reference-scope}
The references beside Table~\ref{tab:current-dailylife} describe published models;
evaluation identities and scores come from the dated DailyLife source snapshot.
A publication date is distinct from a leaderboard label date. Vendor references
do not verify the evaluated checkpoint, date, effort setting, or harness.
References were checked on October 11, 2026; undated documentation is cited as such.

The four entries marked \emph{family} (DL-14, DL-15, DL-23, DL-25) cite Qwen,
Kimi, or Hunyuan documentation without establishing their exact labeled versions.
The other 22 entries cite model or provider documentation of varying specificity. DeepSeek's current
documentation describes replacement serving of legacy Flash aliases; it does not
retrospectively identify the checkpoint in a saved run.

\FloatBarrier

\section{Construction Trials and Conditional Package Assessment}
\label{app:construction-trials}
The following records separate attempted feedback comparisons from the later
assessment of one imported package. The first two trials produced no qualified
candidate pair. The imported-package records then establish a qualified weak
output, an original strong-solver failure, a supplemental strong output under
a larger turn allowance, and a later common-reference package assessment.
These stages retain their own failures and endpoints; the later observations
do not complete the earlier construction comparison.

\subsection{Two-source construction: request-size limits}
\label{app:prospective-interface-capacity}

The prospective block used June and December inputs from the same station and
year. Adaptive and independent construction each had two candidate positions.
Both conditions used a shared ceiling of 120 requests per candidate and at most
six native generation sessions: up to three hardening sessions followed by up to
three rubric sessions. Early success could return before all six sessions, and
the sessions shared the 120-request ceiling. The total request ceiling was 600:
480 for construction, 72 for probing, and 48 for final evaluation. The probing
allowance comprised eight solver runs with up to three requests each and
24 judge ballots with up to two requests each. The final-evaluation allowance
comprised four solver runs with up to three requests each, 12 judge ballots with
up to two requests each, and four independent assessments with up to three
requests each. The four source-independent final-assessment stages comprised
source reconstruction and final-package comparison for each construction
condition. HTTP starts were scheduled at least 45 seconds apart.

Two of four construction positions were attempted. No position qualified
(0/4), no judge ballots were obtained (0/36), and no source-independent
final-assessment stage was completed (0/4).
Table~\ref{tab:prospective-interface-capacity} separates request outcomes from
construction outcomes. All 26 actual HTTP starts returned status 200.

\begin{table}[!htbp]
\centering\small
\caption{\textbf{Request outcomes before construction qualification.}
C1 and C2 identify the two planned candidate positions in each condition.
HTTP counts actual starts; C and U count request results marked COMPLETE and
UNKNOWN, respectively. These request-level states are not package-quality
labels. Positions with no HTTP starts were unattempted. The origin of the
observed parsing failure remained unresolved.}
\label{tab:prospective-interface-capacity}
\setlength{\tabcolsep}{3pt}
\begin{tabular}{@{}p{0.24\linewidth}rrrp{0.40\linewidth}@{}}
\toprule
Position & HTTP & C & U & Recorded outcome \\
\midrule
Adaptive C1 & 9 & 8 & 1 & \raggedright Tool-input JSON parsing failure \tabularnewline
Adaptive C2 & 0 & 0 & 0 & \raggedright Blocked by C1 dependency \tabularnewline
Independent C1 & 17 & 17 & 0 & \raggedright Local request-size refusal \tabularnewline
Independent C2 & 0 & 0 & 0 & \raggedright Unattempted at global stop \tabularnewline
\midrule
Total & 26 & 25 & 1 & \raggedright No qualified construction \tabularnewline
\bottomrule
\end{tabular}
\end{table}

There were 27 logical admissions but only 26 actual HTTP starts. The last
admission stopped locally before commit or forwarding when the first observed
overflowing prefix reached 263,057 bytes, above the 262,144-byte limit. The
263,057-byte observation is a prefix size; the full refused-body size remains
unknown. This was a local pre-forward refusal, not a remote HTTP rejection.
The missing ballots leave semantic quality and the paired construction-policy
effect unobserved.

\subsection{Rubric-reserved construction: incomplete candidates}
\label{app:rubric-reserved-construction}
Three of four planned positions were attempted. All 60 requests completed,
but no position qualified (0/4): the three saved drafts had empty rubrics,
and adaptive c2 lacked the qualified prior feedback required to start.
Table~\ref{tab:rubric-reserved-construction} locates the stopping boundaries.

\begin{table}[!htbp]
\centering\small
\caption{\textbf{Construction failure boundaries.}}
\label{tab:rubric-reserved-construction}
\setlength{\tabcolsep}{3pt}
\begin{tabular}{@{}p{0.21\linewidth}p{0.14\linewidth}p{0.14\linewidth}p{0.40\linewidth}@{}}
\toprule
Planned position & Harden requests & Rubric requests & Last reached boundary \\
\midrule
\raggedright Adaptive c1 & 17 & 7 & \raggedright Updated draft passed original harden validation; first rubric generation returned no rubric before the local candidate-pool cap. \tabularnewline
\raggedright Independent c1 & 18 & 0 & \raggedright First native harden generation hit the local harden cap; outer validation was not reached. \tabularnewline
\raggedright Independent c2 & 18 & 0 & \raggedright Same delivery boundary; outer validation was not reached. \tabularnewline
\raggedright Adaptive c2 & 0 & 0 & \raggedright Not attempted: required qualified OwnPrior feedback was unavailable. \tabularnewline
\bottomrule
\end{tabular}
\end{table}

The four native diagnostic streams contained 105 tool invocations, including 17 invalid \texttt{Read.pages} arguments and 21 failed worktree-agent invocations in a workspace without the required Git/hooks support. These tool invocations are not provider requests. They identify interface problems, but not their causal contribution to construction failure. The observed stops were local request caps, not provider rate limiting or repeated outer rubric-validation failures.

The independent c1 draft retained six successful edits: five citation-line-range corrections and an ambiguous-reference/lever declaration. Applying those recorded edits to its initial write reproduced the saved final file byte-for-byte. The normal validation path consumes the updated artifact. Here, the independent drafts reached the native request cap or error boundary and were rejected under the no-partial-acceptance rule. All three saved drafts had empty rubrics; post hoc harden-only structural checks were narrower than the original qualification path and do not establish candidate quality.

Authored-rubric dimensions and weights are distinct from the fixed final assessment's 14 reference obligations and total weight 100. Several count/style/distribution checks are advisories, whereas illegal weights/capability IDs, tolerance types and malformed evidence are fatal. No returned rubric was observed to fail these predicates, and no general contradiction was established. Semantic adequacy remains untested.

Two prospective code corrections must not be recast as retrospective scientific gains. The stage-attribution fix labels the unattempted adaptive c2 correctly without creating an attempt. A separate latent repair-input fix supplies the last successfully returned, whole-JSON-shape-checked current-candidate rubric alongside later feedback; this is repair data, not an accepted rubric. No second rubric round occurred in the closed run.

\subsection{One imported package: execution and reference assessment}
\label{app:conditional-partial-result}

This run evaluates one authenticated imported package on the original exposed
inputs. It reuses a retained native source assessment covering all 14 common
obligations (total weight 100), without a new source conversation. The
package's native judging rubric has 10 criteria and total weight 94; it is a
different instrument and is not rescaled to the common endpoint. The
process-isolated source assessment is same-family model evidence, not an
independent ground-truth dataset. This conditional unit neither completes
the original four-candidate policy comparison nor supplies OwnPrior feedback.

The configured weak solver (\texttt{gpt-5.6-luna}) returned one qualified
output. Three ballots by \texttt{gpt-5.6-sol} each supplied valid votes for
all 10 native criteria. The original scoring routine reported a normalized
weighted score of 0.851 and data completeness of 1, with no missing-verdict,
API-error, or budget-exhaustion ballot. These are repeated judgments of one
solver output, not three independently sampled tasks or solver runs.

The strong solver (\texttt{gpt-5.6-sol}) made 24 completed physical requests
but terminated with SDK \texttt{error\_max\_turns} and return code 1. Its
metadata reports no final answer. The original qualification rule therefore
rejected the endpoint, leaving its three ballots and the final common-14
comparison unattempted. The last response requested tool use; no response
in this run ended with output-token truncation. No provider-budget or time
refusal was recorded. This establishes a solver turn-limit execution
boundary, not a zero score or scientific inferiority. SDK-reported turn
counts are not equated with physical request counts.

\begin{center}
\setlength{\tabcolsep}{3pt}
\begin{tabular}{@{}p{0.21\linewidth}p{0.16\linewidth}p{0.55\linewidth}@{}}
\hline
Stage & New requests & Observed result \\
\hline
Source assessment & 0 & Authenticated retained result reused \\
Weak solver & 19 & Qualified output \\
Weak ballots 1--3 & 13 & Three valid ballots (4/4/5 requests) \\
Strong solver & 24 & No qualified final answer \\
Strong ballots and comparison & 0 & Dependencies incomplete \\
\hline
Total & 56 & All physical responses COMPLETE \\
\hline
\end{tabular}
\end{center}

There were no new UNKNOWN responses or known-429 retries. The cumulative
acquisition and execution history is 107 physical requests: 106 COMPLETE
and one retained UNKNOWN. The 52 unused requests in the closed grant cannot
be reused. The native interval from broker initialization to durable closure
was 2,512.200 seconds. All seven owned native/converter jobs were released.

Response metadata reports 1,628,670 input tokens and 36,077 output tokens.
The five native jobs report a combined USD 9.045275; this is an SDK estimate,
not reconciled billing. Charged requested-output allowances were 229,376
tokens for this run, distinct from actual reported output usage. Historical
and cumulative invoiced costs remain unknown.

The weak measurement is conditional on this package, solver output, rubric,
and execution setting. It does not establish scoring accuracy, an independent
source-quality label, or a weak--strong difference. At closure, the common-14 comparison
was missing. Any later increase to the strong solver's turn allowance
must be reported as a changed resource condition, not as completion of this
closed equal-limit run.

\subsubsection{Supplemental strong-resource condition and missing assessment}
\label{app:conditional-strong-supplement}

The original 24/24 configured-limit observation remains unchanged: the weak
solver produced a qualified output with a normalized native score of 0.851,
whereas the strong solver exhausted its 24-turn SDK allowance without a
qualified final answer. Its failed computation and costs remain in the
accounting. There is no paired score difference for that original condition.

The supplemental run authenticated and reused the fixed source assessment
(14 obligations, total weight 100), the locked task package, the weak solver
output, and its three ballots. None was rerun. It started a fresh strong
conversation under a 48-turn SDK ceiling, without importing or replaying the
failed conversation. The strong solver returned a qualified final answer with
SDK-reported turn count 28 after 21 provider requests. SDK turns and physical
requests are distinct counters. Three valid ballots on the unchanged
10-criterion, weight-94 native rubric yielded a normalized score of 0.926;
all ten criteria had three valid votes and no missing native ballot.

The scores 0.851 and 0.926 therefore describe a retained weak output and a
fresh strong output under different configured resource conditions. They do
not establish equal-resource model superiority, a paired policy effect, or a
causal benefit of increasing the turn limit. Each score concerns one solver
output on one imported package; repeated model ballots are not independent
tasks or external ground truth. The native rubric was not rescaled to the
common-reference weight of 100.

\begin{table}[t]
\centering
\small
\setlength{\tabcolsep}{2pt}
\caption{Actual supplemental execution, including the failed comparison.
Reported SDK costs are not independently reconciled invoices.}
\label{tab:conditional-strong-supplement-cost}
\begin{tabular}{@{}p{0.49\linewidth}p{0.23\linewidth}p{0.22\linewidth}@{}}
\hline
Supplemental stage & Provider requests & SDK-reported USD \\
\hline
Strong solver & 21 & 3.342600 \\
Strong ballot 1 & 4 & 0.846905 \\
Strong ballot 2 & 4 & 0.826730 \\
Strong ballot 3 & 4 & 0.806045 \\
Common-14 comparison (no verdict) & 6 & 0.454210 \\
\hline
Total & 39 & 6.276490 \\
\hline
\end{tabular}
\end{table}

The final common-14 endpoint evaluates the fixed task package against the
locked reference, not the solver ranking. Its six requests completed and
the SDK reported success, but its recorded decision-handler counters remained zero:
zero accepted submissions, zero rejected submissions, and no qualifying
14-obligation result. The caller stopped with
\texttt{SOURCE\_ENDPOINT\_NO\_VERDICT}. This is missing package-assessment
evidence, not a score of zero, a negative scientific verdict, or an observed
weight/schema rejection. These aggregate diagnostics do not establish whether a tool call was
attempted and rejected before the handler. The retained qualified source
assessment does not substitute for this unanswered package comparison.

All 39 new requests were recorded COMPLETE, with no new UNKNOWN outcome,
no known-429 retry, and no response ending at its token limit. The responses
reported 1,136,908 input tokens and 23,678 output tokens; these totals agree
with the five SDK job records. Charged requested-output allowances totaled
233,472 tokens, which is a limit-accounting quantity rather than output
usage. Native broker time was 1,739.854 seconds. All owned jobs were released.

At the end of this supplemental run, the cumulative history contained 146 physical requests: 145 COMPLETE and
one immutable historical UNKNOWN. The original 24-turn strong failure alone
reported USD 3.690590 and is not removed by the supplemental success. The
preceding 56-request unit reported USD 9.045275; that unit plus this one
reported USD 15.321765, not the full cost of the 146-request history. Invoiced
costs for that complete history remain unknown. The 45 unused requests in
the closed supplemental grant are not available for another run.

\subsubsection{Completed common-reference package assessment}
\label{app:conditional-common14-result}

The final comparison assessed one fixed constructed package against the
independently locked source reference using the unchanged 14-obligation,
weight-100 common rubric. The source assessment, both solver outputs and all
six native ballots were authenticated and retained without rerunning them.
Solver outputs were not mounted in the comparison view. One structured
submission was accepted, with no rejected submission. The original checker
validated all 14 unique obligation IDs, their unchanged weights and required
evidence fields, together with the input, runtime, endpoint and stage bindings.
The caller's parsed labels agreed with the validated submission.

All obligations O01--O14 were labelled supported. Their fixed weights, in
that order, were 7, 7, 8, 7, 10, 8, 8, 7, 8, 5, 6, 5, 7 and 7.
Supported, contradicted and unknown weights were therefore 100, 0 and 0,
respectively, out of 100; no row or required evidence field was missing or
invalid. The resulting within-package identification interval was $[1,1]$.
This interval summarizes the submitted labels and is not a statistical
confidence interval. The 14 obligations are components of one assessment,
not independent sampled tasks. Evidence-field and provenance validation
establishes a qualifying model assessment, not independent factual validation
of every rationale or of the reference itself.

This comparison used three provider requests, all COMPLETE, with no new
UNKNOWN outcome, known-429 retry or response ending at its token limit.
The SDK reported 59,522 input tokens, 3,168 output tokens and USD~0.37681;
these token totals agreed with the three physical-response records.
Requested output allowances totalled 49,152 tokens, a ceiling-accounting
quantity rather than generated-token usage. The SDK reported 11 turns under
a 13-turn ceiling, and the broker ran for 113.876 seconds. All owned children
were reaped and the grant closed; its nine unused request slots are not
available for reuse. The SDK cost is reported usage cost, not an independently
reconciled invoice.

The cumulative history contains 153 physical requests: 152 COMPLETE and one
retained historical UNKNOWN, which was not replayed. Earlier failed attempts
and their costs remain charged. The known cost subtotal for the prior
39-request unit, the prior four-request comparison and this comparison is
USD~7.18017; it is not the total cost of the 153-request history. Missing
invoice or earlier cost information is not counted as zero.

Earlier comparisons without a qualifying structured submission remain
no-verdict observations. The old zero decision-handler count does not
establish whether a tool was attempted and rejected before that counter.
The present result comes from a new accepted native submission after
prospective interface repair, not from reconstructing an earlier verdict.
The native solver rubric still contains 10 criteria with total weight 94
and was not rescaled to the common rubric. The original matched 24/24
condition retains the strong solver's no-final outcome. The supplemental
strong output used a 48-turn ceiling whereas the retained weak output used
24; neither the supplemental scores nor the common-reference assessment
establish equal-resource model superiority or a causal effect of more turns.

\end{document}